\documentclass[accepted]{uai2025} %

\usepackage[american]{babel}

\usepackage{natbib} %
\usepackage{mathtools} %
\usepackage{booktabs} %
\usepackage{tikz} %
\usepackage{arydshln}
\usepackage{fix-cm}

\definecolor{verylightgray}{gray}{0.9}
\definecolor{lightgray}{gray}{0.8}

\definecolor{C0}{RGB}{76, 114, 176}
\colorlet{C0}{C0!17!white}
\definecolor{C1b}{RGB}{242, 173, 133}
\colorlet{C1b}{C1b!12!white}
\definecolor{C1}{RGB}{221, 132, 82}
\colorlet{C1}{C1!17!white}
\definecolor{C1d}{RGB}{204, 99, 51}
\colorlet{C1d}{C1d!21!white}
\definecolor{C2}{RGB}{85, 168, 104}
\colorlet{C2}{C2!17!white}
\definecolor{C3}{RGB}{129, 114, 178}
\colorlet{C3}{C3!17!white}
\definecolor{C4}{RGB}{196, 78, 82}
\colorlet{C4}{C4!17!white}

\usepackage{color}
\usepackage{amsmath}
\usepackage{amssymb}
\usepackage{amsthm}
\usepackage{mathtools}
\usepackage{thmtools}
\usepackage[mathscr]{eucal}
\usepackage{bm}
\usepackage{bbm}
\usepackage{relsize}
\usepackage{graphics}
\usepackage{wrapfig}
\usepackage{multirow}
\usepackage{enumitem} 
\usepackage{cancel}

\usepackage[algo2e]{algorithm2e}

\usepackage[capitalize,noabbrev]{cleveref}

\usepackage{array}

\usepackage{pgf}
\usepackage{xinttools}
\usepackage{tikz}
\usetikzlibrary{arrows.meta}
\usepackage{textcomp}
\usetikzlibrary{calc,shapes,arrows,positioning,automata,trees}
\usepackage{placeins} 
\usepackage{tabularx}
\usepackage{graphicx}
\usepackage{subcaption} 
\usepackage{listings}
\usepackage{xargs,lipsum,caption,changepage,ifthen}

\usepackage{tikz} 

\usepackage{footmisc}

\usepackage{bigdelim}
\usepackage{colortbl} 
\definecolor{mColor1}{rgb}{0.95,0.95,0.95}
\newcolumntype{a}{>{\columncolor{mColor1}}c}

\usepackage[toc,page]{appendix}

\usepackage{tcolorbox}

\definecolor{solarized@base03}{HTML}{002B36}
\definecolor{solarized@base02}{HTML}{073642}
\definecolor{solarized@base01}{HTML}{586e75}
\definecolor{solarized@base00}{HTML}{657b83}
\definecolor{solarized@base0}{HTML}{839496}
\definecolor{solarized@base1}{HTML}{93a1a1}
\definecolor{solarized@base2}{HTML}{EEE8D5}
\definecolor{solarized@base3}{HTML}{FDF6E3}
\definecolor{solarized@yellow}{HTML}{B58900}
\definecolor{solarized@orange}{HTML}{CB4B16}
\definecolor{solarized@red}{HTML}{DC322F}
\definecolor{solarized@magenta}{HTML}{D33682}
\definecolor{solarized@blue}{HTML}{268BD2}
\definecolor{solarized@cyan}{HTML}{2AA198}
\definecolor{solarized@green}{HTML}{859900}

\newtcolorbox{importantresult}{colback=solarized@yellow!5!white,
colframe=solarized@yellow,parbox, left=0.5mm, right=0.5mm,top=0.5mm,bottom=0.5mm}

\newtcolorbox{importantresult_noparbox}{colback=solarized@yellow!5!white,
colframe=solarized@yellow,parbox=false, left=0.5mm, right=0.5mm,top=0.5mm,bottom=0.5mm}

\newtheorem*{theorem*}{Theorem}
\newtheorem*{definition*}{Definition}

\newcommand\Bw{\bm{w}}
\newcommand\Bx{\bm{x}}
\newcommand\By{\bm{y}}

\newcommand\Bth{\bm{\theta}}

 \newcommand{\dR}{\mathbb{R}}

 \newcommand{\rD}{\mathrm{D}}

\newcommand{\rI}{\mathrm{I}}

 \newcommand{\cD}{\mathcal{D}}

\newcommand{\cY}{\mathcal{Y}}

\newcommand\ENT{\mathbf{\mathrm{H}}}
\newcommand\EXP{\mathbf{\mathrm{E}}}

\newcommand\argmax{\mathop{\mathrm{argmax}\,}}

\DeclarePairedDelimiterX{\kldiv}[2]{(}{)}{%
  #1\;\delimsize\|\;#2%
}
\newcommand{\KL}{\mathrm{KL}\kldiv}

\DeclarePairedDelimiterX{\mi}[2]{(}{)}{%
  #1\;\delimsize ; \;#2%
}

\DeclarePairedDelimiterX{\di}[2]{(}{)}{%
  #1\;\delimsize ; \;#2%
}

\DeclarePairedDelimiterX{\ce}[2]{(}{)}{%
  #1\;\delimsize ; \;#2%
}
\newcommand{\CE}{\mathrm{CE}\ce}

\DeclarePairedDelimiterXPP{\mii}[3]%
   {_{\mathrm{#1}}}{(}{)}{}{#2\;\delimsize ; \;#3%
}

\makeatletter
\newcommand{\dlmf}[1]{%
\citep[%
  \def\nextitem{\def\nextitem{, }}%
  \@for \el:=#1\do{\nextitem\href{http://dlmf.nist.gov/\el}{(\el)}}%
]{Olver:10}%
}
\makeatother

\usepackage{pdflscape} 

\newcolumntype{R}[1]{>{\raggedright\arraybackslash}p{#1}}
\newcolumntype{C}[1]{>{\centering\arraybackslash}p{#1}}
\newcolumntype{L}[1]{>{\raggedleft\arraybackslash}p{#1}}

\usepackage{bigdelim}
\definecolor{mColor1}{rgb}{0.95,0.95,0.95}

\hypersetup{
   colorlinks=true,
   linkcolor=[RGB]{30, 30, 180},
   citecolor=[RGB]{30, 30, 180},
   urlcolor=black,
   pdfborder=0 0 0,
   pdftitle={},
   pdfsubject={}, 
   pdfkeywords={},
   pdfauthor={},%
   pdfstartview=FitH
}

\title{On Information-Theoretic Measures of Predictive Uncertainty}

\author[1]{\href{mailto:<schweighofer@ml.jku.at>?Subject=Your UAI 2025 paper}{Kajetan Schweighofer}{}}
\author[1]{\href{mailto:<aichberger@ml.jku.at>?Subject=Your UAI 2025 paper}{Lukas Aichberger}{}}
\author[1]{\href{mailto:<ielanskyi@ml.jku.at>?Subject=Your UAI 2025 paper}{Mykyta Ielanskyi}{}}
\author[1,2]{\href{mailto:<hochreit@ml.jku.at>?Subject=Your UAI 2025 paper}{Sepp Hochreiter}{}}
\affil[1]{%
    ELLIS Unit Linz and LIT AI Lab, Institute for Machine Learning, \linebreak 
    Johannes Kepler University Linz\\
    Austria
}
\affil[2]{%
    NXAI GmbH\\
    Linz\\
    Austria
}
  
\begin{document}
\maketitle

\begin{abstract}
    Reliable estimation of predictive uncertainty is crucial for machine learning applications, particularly in high-stakes scenarios where hedging against risks is essential.
    Despite its significance, there is no universal agreement on how to best quantify predictive uncertainty.
    In this work, we revisit core concepts to propose a framework for information-theoretic measures of predictive uncertainty.
    Our proposed framework categorizes predictive uncertainty measures according to two factors: \textbf{(I)} The predicting model \textbf{(II)} The approximation of the true predictive distribution.
    Examining all possible combinations of these two factors, we derive a set of predictive uncertainty measures that includes both known and newly introduced ones.
    We extensively evaluate these measures across a broad set of tasks, identifying conditions under which certain measures excel.
    Our findings show the importance of aligning the choice of uncertainty measure with the predicting model on in-distribution (ID) data, the limitations of epistemic uncertainty measures for out-of-distribution (OOD) data, and that the disentanglement between measures varies substantially between ID and OOD data.
    Together, these insights provide a more comprehensive understanding of predictive uncertainty measures, revealing their implicit assumptions and relationships.
\end{abstract}

\section{Introduction}

Integrating machine learning models into high-stakes scenarios, such as autonomous driving or managing critical healthcare systems, introduces substantial risks.
To hedge against these risks, we need to quantify the uncertainty associated with each prediction to prevent models from making decisions that carry significant risk and uncertainty.
In such cases, it is better to defer uncertain decisions to human experts or opt for a safer, though potentially less advantageous, alternative decision.
Consequently, it is vital to employ reliable measures of predictive uncertainty and provide estimates for them when implementing machine learning models for decision making in high-stakes applications.

The entropy of the posterior predictive distribution has become the standard information-theoretic measure to assess predictive uncertainty \citep{Houlsby:11, Gal:16thesis, Depeweg:18, Smith:18, Mukhoti:23}. 
Despite its widespread use, this measure has drawn criticism \citep{Malinin:21, Wimmer:23}, prompting the proposal of alternative information-theoretic measures \citep{Malinin:21, Schweighofer:23a, Schweighofer:23b} and frameworks of uncertainty measures based on proper scoring rules that encompass information-theoretic measures \citep{kotelevskii2025risk, Hofman:24}. 
However, it remains hard to gauge in which practical settings one measure should be favored over another. 

We show that all these measures can be interpreted as approximations of the same measure, the cross-entropy between the predictive distributions of the predicting model and the true model.
The predicting model is used to predict on new data, while the true model generated the dataset.
However, since we do not know the true model, this fundamental measure is intractable to compute.
Therefore, we consider different assumptions about the predicting model and how the true predictive distribution is approximated.
This gives rise to our proposed framework to categorize information-theoretic measures of predictive uncertainty.
Our framework includes existing measures, introduces new ones, and clarifies their relationships by eliciting implicit assumptions.
In sum, our contributions are as follows:

\begin{itemize}[leftmargin=*, itemsep=0pt, partopsep=0pt, parsep=2pt]
    \item We present a unifying framework to categorize predictive uncertainty measures according to assumptions about the predicting model and how the true predictive distribution is approximated. This framework encompasses existing measures, but also suggests new ones, clarifies their relationship, and guides the selection of appropriate measures.
    \item We evaluate the measures suggested by our framework across a broad set of tasks, identifying specific conditions under which certain measures outperform others in practice. Notably, for tasks depending on which model predicts, it is best to align the uncertainty measure with the predicting model. Furthermore, we assess the level of disentanglement between measures, finding a pronounced dependency on the distribution the new data is coming from.
\end{itemize}

\section{Quantifying Predictive Uncertainty}

We consider the canonical classification setting with inputs $\Bx \in \dR^D$ and targets $y \in \cY$, where $\cY$ is the set of all $K$ possible targets.
The dataset $\cD$ is given, sampled i.i.d. according to the data generating distribution. 
We consider deep neural networks as a class of probabilistic models that map an input $\Bx$ to the $K-1$ dimensional probability simplex $\Delta^{K-1} = \{\Bth \in \dR^K \mid \theta_k \geq 0 \ \forall k, \sum_{k=1}^K \theta_k = 1\}$.
This mapping is defined as $f_{\Bw}: \dR^D \rightarrow \Delta^{K-1}$ for a model with parameters $\Bw$.
The output of this mapping defines the distribution parameters of a categorical distribution, in the following referred to as the model's predictive distribution $p(y \mid \Bx, \Bw) = \text{Cat}(y; f_{\Bw}(\Bx)) = \text{Cat}(y; \Bth)$.

The predictive distribution of a probabilistic model represents the uncertainty inherent in its predictions.
When the probability mass is uniformly distributed across all possible outcomes, it denotes complete uncertainty about the prediction, whereas concentration on a single class indicates complete certainty.
If we have access to the true data-generating model, denoted by parameters $\Bw^*$, the predictive distribution $p(y \mid \Bx, \Bw^*)$ captures the inherent and irreducible uncertainty in the prediction, often referred to as \emph{aleatoric uncertainty} (AU) \citep{Gal:16thesis, Kendall:17}.
This assumes that the chosen model class can accurately represent the true predictive distribution, thus $p(y \mid \Bx) = p(y \mid \Bx, \Bw^*)$, which is a common and often necessary assumption \citep{Huellermeier:21}.
We will discuss how to interpret AU if we do not have access to $\Bw^*$ in Sec.~\ref{sec:cross_entropy}.
The information-theoretic entropy $\ENT(\cdot)$ \citep{Shannon:48} of the true predictive distribution is a natural and universally accepted measure of AU, defined as
\begin{align} \label{eq:entropy}
    \ENT(p(y \mid \Bx, \Bw^*)) \ \coloneqq \ \EXP_{p(y \mid \Bx, \Bw^*)} \left[ - \log p(y \mid \Bx, \Bw^*) \right]
\end{align}
However, we generally don't know the true model and have to choose parameters $\Bw$ out of all possible ones.
Consequently, uncertainty arises due to the lack of knowledge about the true parameters of the model.
This is called \emph{epistemic uncertainty} (EU) \citep{Apostolakis:90, Helton:93, Helton:97, Gal:16thesis, Smith:18}.
An effective measure of predictive uncertainty should be consistent with Eq.~\eqref{eq:entropy} and capture both AU and EU, typically assumed to sum up to a \emph{total uncertainty} (TU).

\subsection{Current Standard Measure: Entropy of Posterior Predictive Distribution}

Given a dataset $\cD$ and prior $p(\Bw)$ on the model parameters, Bayes' theorem yields the posterior distribution $p(\Bw \mid \cD)$.
The posterior distribution denotes the probability that the parameters $\Bw$ match the true parameters $\Bw^*$ of the model that generated the dataset $\cD$.   
Instead of committing to a single model, the posterior distribution allows marginalizing over all possible models, which is known as Bayesian model averaging.
This gives rise to the posterior predictive distribution 
\begin{align} \label{eq:posterior_predictive}
    p(y \mid \Bx, \cD) \ = \ \EXP_{p(\Bw \mid \cD)} \left[ p(y \mid \Bx, \Bw) \right] \ .
\end{align}
The entropy of the posterior predictive distribution is the currently most widely accepted approach to measure predictive uncertainty
\citep{Houlsby:11, Gal:16thesis, Depeweg:18, Smith:18, Huellermeier:21, Mukhoti:23}.
According to a well-known result from information theory \citep{Cover:06}, this entropy can be additively decomposed into the conditional entropy and the mutual information $\rI(\cdot)$ between $y$ and $\Bw$:
\begin{align} \label{eq:original_measure}
        &\underbrace{\ENT ( p(y \mid \Bx, \cD ) )}_{\text{TU}} \\ \nonumber 
        & \quad = \ \underbrace{\EXP_{p(\Bw \mid \cD )} \left[ \ENT (p(y \mid \Bx, \Bw ) ) \right]}_{\text{AU}} \ + \ \underbrace{\rI(p(y, \Bw \mid \Bx, \cD))}_{\text{EU}} \ .
\end{align}
Eq.~\eqref{eq:original_measure} is equivalent to a decomposition of expected cross-entropy $\CE{\cdot}{\cdot}$ into conditional entropy and expected KL-divergence $\KL{\cdot}{\cdot}$ \citep{Schweighofer:23a,Schweighofer:23b}:
\begin{align} \label{eq:original_measure_2}
    &\underbrace{\EXP_{p(\Bw \mid \cD )} \left[ \CE{p(y \mid \Bx, \Bw )}{p(y \mid \Bx, \cD )} \right]}_{\text{TU}} \\ \nonumber 
    & \qquad\quad = \ \underbrace{\EXP_{p(\Bw \mid \cD )} \left[ \ENT (p(y \mid \Bx, \Bw )) \right]}_{\text{AU}} \\ \nonumber 
    & \qquad\quad + \ \underbrace{\EXP_{p(\Bw \mid \cD )} \left[ \KL{p(y \mid \Bx, \Bw )}{p(y \mid \Bx, \cD )} \right]}_{\text{EU}} \ .
\end{align}
If the parameters of the true model are known, EU vanishes and Eq.~\eqref{eq:original_measure} as well as Eq.~\eqref{eq:original_measure_2} simplify to Eq.~\eqref{eq:entropy}, thus are consistent with it.
However, the entropy of the posterior predictive distribution has been found to be inadequate for specific scenarios, such as autoregressive predictions \citep{Malinin:21} or for a given predicting model \citep{Schweighofer:23a} and was criticised on grounds of not fulfulling certain expected theoretical properties \citep{Wimmer:23}.
In response, alternative information-theoretic measures \citep{Malinin:21, Schweighofer:23a, Schweighofer:23b} and frameworks of uncertainty measures based on proper scoring rules \citep{kotelevskii2025risk, Hofman:24} have been introduced.
We seek to give a unified framework for these measures that is interpretable and can guide the selection of uncertainty measures in practice.
Therefore, we next propose a fundamental, yet generally intractable, predictive uncertainty measure, where all of these measures are special cases under specific assumptions.

\subsection{Our Proposed Measure: Cross-Entropy between Selected and True Predictive Distribution} \label{sec:cross_entropy}

An effective measure of TU should be consistent with Eq.~\eqref{eq:entropy} and should incorporate EU.
Given this, we propose to measure predictive uncertainty with the cross-entropy between the predictive distributions of a selected predicting model and the true model.
Let $p(y \mid \Bx, \cdot)$ be the predictive distribution of any selected model for some new input $\Bx$, which we will refer to as the \emph{predicting model}.
We will examine different cases for the predicting model later; for now, it suffices to consider it to be a specific model with parameters $\Bw$.
The cross-entropy between the predictive distributions of the predicting model and the true model is given by
\begin{align} \label{eq:cross_entropy}
    &\underbrace{\CE{p(y \mid \Bx, \cdot)}{p(y \mid \Bx, \Bw^*)}}_{\text{TU}} \\ \nonumber 
    & \coloneqq \ \EXP_{p(y \mid \Bx, \cdot)} \left[ - \log p(y\mid \Bx, \Bw^*) \right] \\ \rule{0pt}{2.8ex} \nonumber 
    &= \ \; \underbrace{\ENT(p(y \mid \Bx, \cdot))}_{\text{AU}} \ + \ \underbrace{\KL{p(y \mid \Bx, \cdot)}{p(y \mid \Bx, \Bw^*)}}_{\text{EU}} \ .
\end{align}
If the predictive distribution of the predicting model is equal to the predictive distribution of the true model, the EU is zero by definition of the KL-divergence and Eq.~\eqref{eq:cross_entropy} simplifies to Eq.~\eqref{eq:entropy}.
Thus, as expected, if the parameters of the true model are known, the EU vanishes.
Eq.~\eqref{eq:cross_entropy} is a fundamental, though generally intractable, measure of predictive uncertainty.
To obtain tractable measures, assumptions about the predicting model and about how to approximate the true predictive distribution are necessary.
This gives rise to our framework, which we introduce in detail in Sec.~\ref{sec:framework}.
Notably, the opposite order of the arguments in Eq.~\eqref{eq:cross_entropy} leads to the same framework but differs in the interpretation of AU and EU. 
However, we find that Eq.~\eqref{eq:cross_entropy} is more useful in practice, allowing to align the uncertainty measure to the predicting model.
For details on the alternative measure see Apx.~\ref{sec:apx:alternative_measure}.

\textbf{Interpretation of AU and EU.}
An important distinction compared to previous work is in our interpretation of AU and EU, which aligns with the understanding of \cite{Apostolakis:90, Helton:93, Helton:97, Schweighofer:23a} as follows.
The AU is not generally understood as a property of the true predictive distribution, but of the selected predicting model used to make a prediction.
Thus, it is the uncertainty that arises due to predicting with the selected probabilistic model.
The EU is defined as the additional uncertainty due to predicting with the selected predicting model instead of the true model.
Thus, it is the additional uncertainty that arises due to selecting a model from the given model class.

\section{Proposed Framework} \label{sec:framework}

Our proposed measure of predictive uncertainty (Eq.~\eqref{eq:cross_entropy}) allows for different assumptions about (I) the selected predicting model and (II) how to approximate the true predictive distribution.
We consider three different assumptions for each, denoted as \texttt{(A,B,C)} for the predicting model and \texttt{(1,2,3)} for the approximation of the true predictive distribution.
This results in nine distinct predictive uncertainty measures within our proposed framework.
An overview of all measures is given in Tab.~\ref{tab:framework_main}, summarizing possible measures of total uncertainty (TU), as well as their respective aleatoric uncertainty (AU) and epistemic uncertainty (EU).

\setlength{\tabcolsep}{13pt}
\renewcommand{\arraystretch}{1.5}
\begin{table*}
\centering
\caption{\textbf{Our proposed framework of information-theoretic measures of predictive uncertainty.} Each measure denotes a different approximation of the fundamental measure given by Eq.~\eqref{eq:cross_entropy} for different assumptions about the predicting model and how the true model is approximated. For brevity, we define $p_{\Bw} \coloneqq p(y \mid \Bx, \Bw)$, $p_{\cD} \coloneqq p(y \mid \Bx, \cD)$, and $\EXP_{\Bw} \coloneqq \EXP_{p(\Bw \mid \cD)}$ (the same for $\tilde\Bw$). Expressions with the same cell coloring are equivalent to each other. For each measure, TU additively decomposes into AU and EU by $\CE{p}{q} = \ENT(p) + \KL{p}{q}$.}
\label{tab:framework_main}
\begin{tabular}{cl|ccc}
\hline
\multirow{2}{*}{\rotatebox{90}{}} & \hfill\multirow{2}{*}{\begin{tabular}[c]{@{}c@{}}Predicting model\end{tabular}}\hfill & \multicolumn{3}{c} {Approximation of the true predictive distribution} \\ \cline{3-5}
& & \cellcolor{verylightgray} \texttt{(1)} $\; \tilde\Bw $ & \cellcolor{verylightgray} \texttt{(2)} $\; \EXP_{\tilde\Bw}$ & \cellcolor{verylightgray} \texttt{(3)} $\; \tilde\Bw \sim p(\tilde\Bw \mid \cD)$ \\
\hline
\multirow{3}{*}{\rotatebox{90}{TU}} & \cellcolor{verylightgray} \texttt{(A)} $\; \Bw$ & $\CE{p_{\Bw}}{p_{\tilde\Bw}}$ & $\CE{p_{\Bw}}{p_{\cD}} $  &  $\EXP_{\tilde\Bw} \left[ \CE{p_{\Bw}}{p_{\tilde\Bw}} \right] $ \\
& \cellcolor{verylightgray} \texttt{(B)} $\; \EXP_{\Bw}$ & \cellcolor{C1b}$\CE{p_{\cD}}{p_{\tilde\Bw}}$ & \cellcolor{C1}$\CE{p_{\cD}}{p_{\cD}} $  &  \cellcolor{C1d}$\EXP_{\tilde\Bw} \left[ \CE{p_{\cD}}{p_{\tilde\Bw}} \right] $  \\
& \cellcolor{verylightgray} \texttt{(C)} $\; \Bw \sim p(\Bw \mid \cD)$ & \cellcolor{C1b}$\EXP_{\Bw} \left[ \CE{p_{\Bw}}{p_{\tilde\Bw}} \right] $ & \cellcolor{C1}$\EXP_{\Bw} \left[ \CE{p_{\Bw}}{p_{\cD}} \right] $  &  \cellcolor{C1d}$\EXP_{\Bw} \left[ \ \EXP_{\tilde\Bw} \left[ \CE{p_{\Bw}}{p_{\tilde\Bw}} \right] \right] $ \\
\hline \hline
\multirow{3}{*}{\rotatebox{90}{AU}} & \cellcolor{verylightgray} \texttt{(A)} $\; \Bw$ & \cellcolor{C0}$\ENT (p_{\Bw})$ & \cellcolor{C0}$\ENT (p_{\Bw})$  & \cellcolor{C0}$\ENT (p_{\Bw})$ \\
&\cellcolor{verylightgray} \texttt{(B)} $\; \EXP_{\Bw}$ & \cellcolor{C1}$\ENT (p_{\cD})$ & \cellcolor{C1}$\ENT (p_{\cD})$  &  \cellcolor{C1}$\ENT (p_{\cD})$  \\
& \cellcolor{verylightgray} \texttt{(C)} $\; \Bw \sim p(\Bw \mid \cD)$ & \cellcolor{C2}$\EXP_{\Bw} \left[ \ENT (p_{\Bw}) \right] $ & \cellcolor{C2}$\EXP_{\Bw} \left[ \ENT (p_{\Bw}) \right] $  &  \cellcolor{C2}$\EXP_{\Bw} \left[ \ENT (p_{\Bw}) \right] $ \\
\hline \hline
\multirow{3}{*}{\rotatebox{90}{EU}} & \cellcolor{verylightgray} \texttt{(A)} $\; \Bw$ & $\KL{p_{\Bw}}{p_{\tilde\Bw}}$ & $\KL{p_{\Bw}}{p_{\cD}} $  &  $\EXP_{\tilde\Bw} \left[ \KL{p_{\Bw}}{p_{\tilde\Bw}} \right] $ \\
& \cellcolor{verylightgray} \texttt{(B)} $\; \EXP_{\Bw}$ & $\KL{p_{\cD}}{p_{\tilde\Bw}}$ & $\cancelto{0}{\KL{p_{\cD}}{p_{\cD}}}$  & $\EXP_{\tilde\Bw} \left[ \KL{p_{\cD}}{p_{\tilde\Bw}} \right] $  \\
& \cellcolor{verylightgray} \texttt{(C)} $\; \Bw \sim p(\Bw \mid \cD)$ & $\EXP_{\Bw} \left[ \KL{p_{\Bw}}{p_{\tilde\Bw}} \right] $ & $\EXP_{\Bw} \left[ \KL{p_{\Bw}}{p_{\cD}} \right] $  & $\EXP_{\Bw} \left[ \ \EXP_{\tilde\Bw} \left[ \KL{p_{\Bw}}{p_{\tilde\Bw}} \right] \right] $ \\
\hline
\end{tabular}
\end{table*}
\renewcommand{\arraystretch}{1}

\renewcommand{\thesubsection}{\texttt{(A,B,C)}:\hspace{-0.2cm}} %
\subsection{Predicting Model}

One can make different choices about the model used during inference for predicting the class of a new input.
The most obvious choice of a predicting model is \texttt{(A)} a pre-selected given model with parameters $\Bw$.
This is the standard case in machine learning, where model parameters are selected, e.g. by maximizing the likelihood on the training dataset or downloaded from a model hub.

Another widely used approach is \texttt{(B)} the Bayesian model average (see Eq.~\eqref{eq:posterior_predictive}).
Here, instead of predicting with a single model, the predictive distribution is marginalized over all possible models according to their posterior probability.
In practice, exact marginalization is generally intractable and therefore approximated by posterior sampling.

Finally, it is possible to \texttt{(C)} consider every possible model as the predicting model, weighted by their posterior probabilities.
This might seem counterintuitive, as it means that the predicting model is not fixed but is sampled anew for each prediction.
Nevertheless, the AU of the resulting uncertainty measures, $\EXP_{p(\Bw \mid \cD )} \left[ \ENT (p(y \mid \Bx, \Bw )) \right]$, is the best approximation of the AU under the true model for a given posterior distribution \citep{Schweighofer:23b}.
However, as pointed out by \cite{Wimmer:23}, it is neither a lower nor an upper bound on the AU under the true model and is highly dependent on the posterior distribution.

\renewcommand{\thesubsection}{\texttt{(1,2,3)}:\hspace{-0.2cm}} %
\subsection{Approximation of the True Predictive Distribution}
\renewcommand{\thesubsection}{\arabic{section}.\arabic{subsection}} %
\setcounter{subsection}{0} %

One can also make different choices about how to approximate the true predictive distribution.
The simplest but probably biased choice to approximate the true predictive distribution is \texttt{(1)} the predictive distribution under a single given model with parameters $\tilde\Bw$.
Although this might be a poor approximation, it might be the only feasible choice in specific settings.
For example, it is used in speculative decoding \citep{Stern:18, Leviathan:23}, where a small model is used to predict and whose predictive distribution is compared against a large model that serves as ground truth.
Furthermore, $\tilde\Bw$ does not necessarily approximate the true model, but instead serves as a reference model of interest, such as a previously used or competitor model, whose discrepancy from the predicting model should be captured.

Another possibility is to use \texttt{(2)} the posterior predictive distribution.
Although intuitively appealing, \cite{Schweighofer:23b} criticized this as there is no guarantee that the posterior predictive distribution approaches the true predictive distribution, even for a perfect estimate of the posterior predictive distribution.
Furthermore, there are degenerate cases where the posterior predictive distribution can't be represented by any model with non-vanishing posterior probability.
However, it is often a well-performing approximation empirically for expressive models such as neural networks.
Moreover, \texttt{(2)} is the only option that guarantees finite EU and TU (for cases \texttt{(B2)} and \texttt{(C2)}).

Finally, perhaps the most intuitive approach is to consider \texttt{(3)} all possible models according to their posterior probability.
Since any model could be the true model under the posterior, we should consider the mismatch between the predictive distribution of the predicting model and that of all possible models, weighted by their posterior probability.

\subsection{Relationships between Measures} \label{sec:relationship}

Importantly, the AU of all uncertainty measures depend only on the predicting model and does not depend on the approximation of the true predictive distribution. 
Thus, they are the same for cases \texttt{(1)}, \texttt{(2)} and \texttt{(3)}.
Furthermore, the AU of case \texttt{(B)} is an upper bound of the AU of case \texttt{(C)}, i.e. $\ENT (p(y \mid \Bx, \cD)) \geq \EXP_{p(\Bw \mid \cD )} \left[ \ENT (p(y \mid \Bx, \Bw )) \right]$, which follows from Eq.~\eqref{eq:original_measure} as the mutual information is non-negative.
Notably, \citet{kotelevskii2025risk} showed that this inequality more generally holds for any proper scoring rule.
Due to the linearity in the first argument of the cross-entropy, the TU for cases \texttt{(B)} and \texttt{(C)} are equal.
Furthermore, as already discussed, the AU for cases \texttt{(B)} and \texttt{(C)} differ by the mutual information $\EXP_{p(\Bw \mid \cD )} \left[ \KL{p(y \mid \Bx, \Bw )}{p(y \mid \Bx, \cD )} \right]$.
Therefore, the EU for cases \texttt{(B)} and \texttt{(C)} also differ by this factor.
This is trivial to see for cases \texttt{(B2)} and \texttt{(C2)}, where the EU of case \texttt{(B2)} cancels to zero and the EU of case \texttt{(C2)} is the mutual information.
For cases \texttt{(B3)} and \texttt{(C3)}, this was already mentioned by \citep{Malinin:21} and a proof was given by \citep{Schweighofer:23b}, which we include for completeness in Apx.~\ref{sec:apx:relationship_epistemic}, together with a version for cases \texttt{(B1)} and \texttt{(C1)}.

\section{Related Work}

\textbf{Information-theoretic measures.}
The standard measure (Eq.~\eqref{eq:original_measure_2}) introduced by \cite{Houlsby:11} and popularized, for instance, by \cite{Gal:16thesis, Depeweg:18, Smith:18} is the measure \texttt{(C2)}.
In the context of autoregressive predictions, %
\cite{Malinin:21} introduced measure \texttt{(B3)}, due to the feasibility of a Monte Carlo (MC) approximation compared to the standard measure \texttt{(C2)}.
\cite{Schweighofer:23a} introduced measure \texttt{(A3)} together with a posterior sampling algorithm that is explicitly taylored to this measure.
\cite{Schweighofer:23b} introduced measure \texttt{(C3)} as an improvement over the standard measure \texttt{(C2)} for certain settings and discussed \texttt{(B3)} in the appendix.
\cite{Hofmann:24} and \cite{kotelevskii2025risk} consider a broad set of proper scoring rules, including log, Brier, zero-one, and spherical score, to derive uncertainty measures.
\cite{kotelevskii2025risk} base their framework on pointwise risk and subsequent Bayesian estimation, which, for the particular case of the log score, yields measures \texttt{(B2)}, \texttt{(B3)}, \texttt{(C2)} and \texttt{(C3)}. 
\cite{Hofman:24} discuss measure \texttt{(C3)} for the particular case of the log score.
See Apx.~\ref{sec:apx:alternative_measure} for more details on how their framework relates to our framework.
Information-theoretic measures have also been considered for uncertainty estimation for large language models, i.e. autoregressive prediction.
Due to the large model sizes, recent work has focussed on AU for case \texttt{(A)} \citep{Kuhn:23, Aichberger:24}.

\textbf{Alternative measures.}
There are also other measures of predictive uncertainty, not based on information-theoretic quantities.
\citet{kotelevskii2022nonparametric} discusses the connection between pointwise risk and uncertainty and provides a nonparametric estimator thereof based on the Nadaraya-Watson kernel.
\cite{Depeweg:18} introduced variance-based measures, based on the law of total variance.
This perspective was recently developed further for specific settings \citep{Duan:24, Sale:23b}.
Furthermore, \cite{Sale:24b} introduced label-wise measures of predictive uncertainty, formulating both information-theoretic and variance-based measures.
Another idea recently proposed by \cite{Sale:24} is quantifying uncertainty through distances to reference (second-order) distributions (for TU, AU, and EU) denoting complete certainty.
Thus, the higher the distance from the reference distribution, the more uncertain the prediction.
All measures discussed so far operate on a distributional representation of uncertainty. 
Orthogonal to that, there are also set-based approaches \citep{Huellermeier:22b, Sale:23, Hofman:24b}.

\section{Experiments} \label{sec:experiments}

Next, we empirically evaluate the performance and characteristics of the uncertainty measures in our proposed framework.
Specifically, we are interested in three different aspects.
First, we investigate if there is merit in aligning the uncertainty measures with the predicting model.
Second, we investigate the performance of the different measures for detecting distributional mismatch.
Finally, we investigate the disentanglement between measures assessing their rank correlation and active learning performance.

\textbf{Datasets.}
Our experiments are performed on the CIFAR10/100 \citep{Krizhevsky:09}, SVHN \citep{Netzer:11}, Tiny-ImageNet (TIN) \citep{Le:15} and LSUN \citep{Yu:15} datasets.
For TIN, we resize the inputs to 32x32 to match the other datasets.
We train models on all datasets except LSUN, which is used solely as an OOD dataset.

\textbf{Models and training.}
We used Deep Ensembles \citep{Lakshminarayanan:17} to approximate posterior expectations through samples (10), as they are the de facto gold standard \citep{Ovadia:19, Izlmailov:21, kotelevskii2025risk}.
For example, the posterior predictive distribution given by Eq.~\eqref{eq:posterior_predictive} is approximated with $N$ samples:
\begin{align} \label{eq:posterior_predictive_mc}
    p(y \mid \Bx, \cD) \ \approx \ \frac{1}{N} \sum_{n=1}^N p(y \mid \Bx, \Bw_n) \ , 
\end{align}
where $\Bw_n \sim p(\Bw \mid \cD)$.
We used three different model architectures for our experiments: ResNet-18 \citep{He:16}, DenseNet-169 \citep{Huang:17} and RegNet-Y 800MF \citep{Radosavovic:20}.
Individual models were trained for 100 epochs using SGD with momentum of 0.9 with a batch size of 256 and an initial learning rate of 1e-2. 
Furthermore, a standard combination of linear (from factor 1 to 0.1) and cosine annealing schedulers was used.
The main paper's results are for ResNet-18. 
Results for the two other architectures, found in Apx.~\ref{apx:sec:networks}, are consistent to those.

\textbf{Predictive uncertainty measures.}
We consider all measures proposed by our framework, see Tab.~\ref{tab:framework_main}. 
For example, the (total) measure \texttt{(A1)} is referred to as TU \texttt{(A1)} with its aleatoric component as AU \texttt{(A)} and its epistemic component as EU \texttt{(A1)}.
Here, AU \texttt{(A)} is used instead of AU \texttt{(A1)} to emphasize the independence from the approximation of the true predictive distribution.
Also, we group equivalent TU measures.
For example TU \texttt{(B1)} and TU \texttt{(C1)} are merged to TU \texttt{(B/C1)}.

\setlength{\tabcolsep}{1.2pt}
\renewcommand{\arraystretch}{1.2}
\begin{table*} \smaller
\centering
\caption{\textbf{Selective prediction under different predicting models.}  AUARC for different predicting models, i.e., the single model, the average model and a model according to the posterior, using different predictive uncertainty measures as score. We highlight the two best measures for each setting; TU \texttt{(B/C2)} and AU \texttt{(B)} are equivalent, thus both get highlighted. For the single predicting model, TU \texttt{(A3)} and EU \texttt{(A3)} perform best, in the other two settings, TU \texttt{(B/C3)} and TU \texttt{(B/C2)} (AU \texttt{(B)}). Results are averaged over all datasets with statistics over five runs.}
\label{tab:selective_prediction}
\begin{tabular}{lcccccccccccccccccc}
\hline
Measures: &  \multicolumn{6}{c}{TU}  & \multicolumn{3}{c}{AU} & \multicolumn{8}{c}{EU} & Random \\
\cdashline{2-7} \cdashline{11-18} \cline{1-1}
\multicolumn{1}{l}{{Prediction}}  & \texttt{A1} & \texttt{A2} & \texttt{A3} & \texttt{B/C1} & \textbf{\texttt{B/C2}} & \texttt{B/C3} & \texttt{A} & \textbf{\texttt{B}} & \texttt{C} & \texttt{A1} & \texttt{A2} & \texttt{A3} & \texttt{B1} & \texttt{B3} & \texttt{C1} & \texttt{C2} & \texttt{C3} & Baseline \\
\hline
\textit{Single} & $87.96$ & $88.10$ & \cellcolor{verylightgray}$88.24$ & $88.07$ & $88.12$ & $88.22$ & $87.50$ & $88.12$ & $87.95$ & $87.30$ & $88.11$ & \cellcolor{verylightgray}$88.23$ & $86.58$ & $87.60$ & $87.31$ & $88.12$ & $87.85$ & $79.67$ \\
\textit{Model} & $\pm0.10$ & $\pm0.10$ & \cellcolor{verylightgray}$\pm0.11$ & $\pm0.09$ & $\pm0.09$ & $\pm0.09$ & $\pm0.11$ & $\pm0.09$ & $\pm0.08$ & $\pm0.08$ & $\pm0.07$ & \cellcolor{verylightgray}$\pm0.11$ & $\pm0.10$ & $\pm0.11$ & $\pm0.10$ & $\pm0.11$ & $\pm0.11$ & $\pm0.10$ \\
\cdashline{1-19}[1pt/2pt]
\textit{Average}  & $89.28$ & $89.52$ & $89.46$ & $90.17$ & \cellcolor{verylightgray}$90.23$ & \cellcolor{verylightgray}$90.30$ & $89.20$ & \cellcolor{verylightgray}$90.23$ & $90.11$ & $88.63$ & $89.15$ & $89.17$ & $88.73$ & $89.52$ & $89.36$ & $90.07$ & $89.78$ & $83.02$ \\
\textit{Model} & $\pm0.06$ & $\pm0.08$ & $\pm0.08$ & $\pm0.05$ & \cellcolor{verylightgray}$\pm0.05$ & \cellcolor{verylightgray}$\pm0.05$ & $\pm0.08$ & \cellcolor{verylightgray}$\pm0.05$ & $\pm0.06$ & $\pm0.04$ & $\pm0.05$ & $\pm0.09$ & $\pm0.03$ & $\pm0.05$ & $\pm0.04$ & $\pm0.06$ & $\pm0.06$ & $\pm0.08$ \\
\cdashline{1-19}[1pt/2pt]
\textit{Acc. to} & $87.74$ & $88.00$ & $87.96$ & $88.74$ & \cellcolor{verylightgray}$88.79$ & \cellcolor{verylightgray}$88.90$ & $87.61$ & \cellcolor{verylightgray}$88.79$ & $88.63$ & $87.06$ & $87.67$ & $87.73$ & $87.20$ & $88.22$ & $87.94$ & $88.75$ & $88.48$ & $80.53$ \\
\textit{Posterior} & $\pm0.08$ & $\pm0.06$ & $\pm0.07$ & $\pm0.09$ & \cellcolor{verylightgray}$\pm0.08$ & \cellcolor{verylightgray}$\pm0.07$ & $\pm0.06$ & \cellcolor{verylightgray}$\pm0.08$ & $\pm0.08$ & $\pm0.07$ & $\pm0.07$ & $\pm0.06$ & $\pm0.12$ & $\pm0.08$ & $\pm0.11$ & $\pm0.06$ & $\pm0.07$ & $\pm0.06$ \\
\hline
\end{tabular}
\end{table*}
\renewcommand{\arraystretch}{1}

\subsection{Aligning The Uncertainty Measure With The Predicting Model} \label{sec:aligning}

The widely regarded selective prediction and misclassification tasks essentially evaluate the correlation of an uncertainty measure with the correctness of the prediction.
However, it is not a priori clear which model is used to predict and if the uncertainty measure needs to be aligned to this choice, which we investigate in the following experiments.

\textbf{Selective prediction.}
In this task, the model's predictions are limited to a specific subset, and its performance is evaluated on that subset.
Therefore, we sampled models on the CIFAR10/100, SVHN and TIN datasets obtain uncertainty estimates and, importantly, predictions on their respective test datasets.
We evaluated the accuracy for a subset of predictions of \textit{(i) the single model}, \textit{(ii) the average model} and \textit{(iii) some model according to the posterior distribution} to investigate the impact of aligning the measure of uncertainty with the predicting model - \texttt{(A)} for setting (i), \texttt{(B)} for setting (ii) and \texttt{(C)} for setting (iii).
The single model for (i) is the first of the sampled models.
The average model for (ii) is defined by Eq.~\eqref{eq:posterior_predictive_mc}, averaging over all sampled models.
For (iii), one model from the sampled models was randomly selected for each prediction.
Note that (iii), also called the fully Bayesian setting, is rather unrealistic in practice.
Selecting a new model according to the posterior for every prediction does not improve over a single fixed model (i) in expectation, if the performance of individual sampled models is comparable.
Generally, it is also more convenient to stick to a single model.

\begin{figure}[t!]
    \centering
    \captionsetup[subfigure]{labelformat=empty}
    \captionsetup[subfigure]{aboveskip=2pt, belowskip=-1pt}
    \captionsetup{aboveskip=2pt, belowskip=1pt}
    {\small Prediction: \textit{Single Model}}
    \includegraphics[width=\linewidth, trim = 0.3cm 1.5cm 0.3cm 1cm, clip]{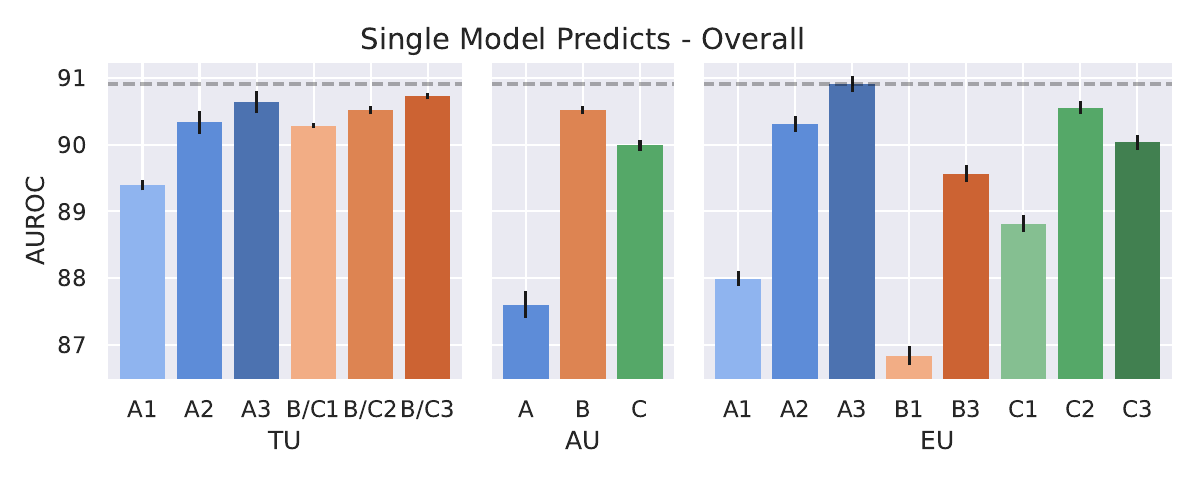}
    {\small Prediction: \textit{Average Model}}
    \includegraphics[width=\linewidth, trim = 0.3cm 1.5cm 0.3cm 1cm, clip]{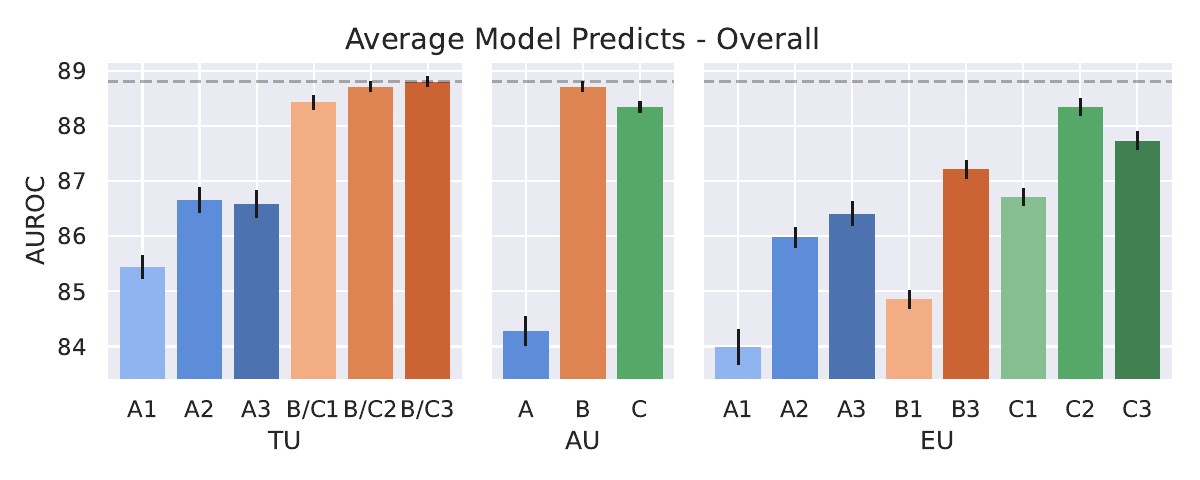}
    {\small Prediction: \textit{According to Posterior}}
    \includegraphics[width=\linewidth, trim = 0.3cm 0.3cm 0.3cm 1cm, clip]{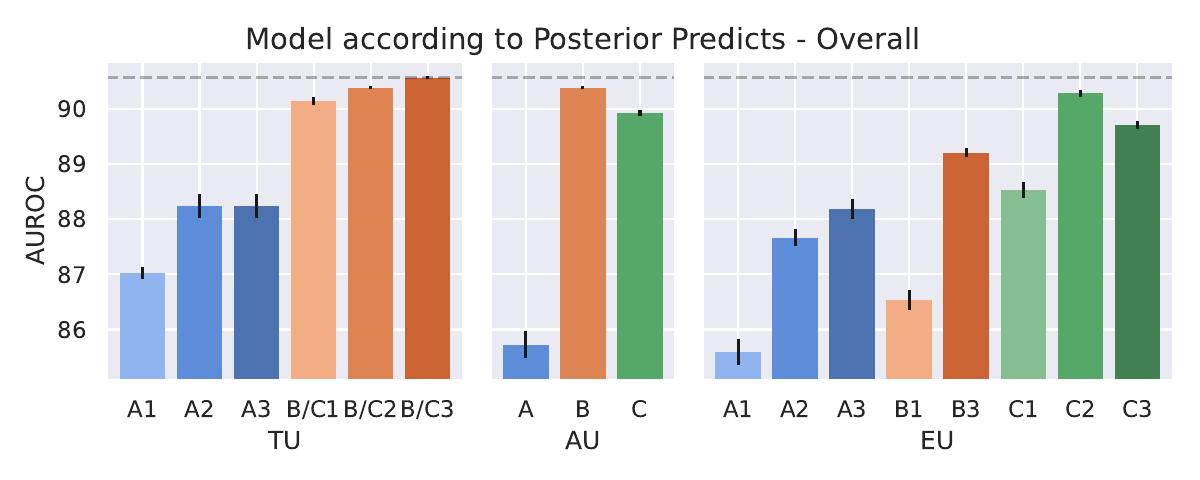} 
    \caption{\textbf{Misclassification detection under different predicting models.} AUROC for distinguishing correct from incorrect predictions under different predicting models, using the different proposed measures of uncertainty as score. EU \texttt{(A3)} performs best when predicting with the single model. For the other cases, TU \texttt{(B/C3)} performs best. AUROCs are averages over all datasets. Statistics over five runs.}
    \label{fig:misc}
\end{figure}

We evaluated subsets ranging from the most certain 50\% of datapoints to the entire dataset. 
To compare how well uncertainty measures rank data for selecting those subsets, we used the area under the accuracy rejection curve (AUARC) as the performance metric.
Results are provided in Tab.~\ref{tab:selective_prediction}, showing that the optimal measure depends on the model used for prediction.
For (i), TU \texttt{(A3)} performs best, followed by EU \texttt{(A3)}, while for (ii) and (iii), TU \texttt{(B/C3)} performs best.
Notably, uncertainty measures perform essentially the same for settings (ii) and (iii).
As we will show in Sec.~\ref{sec:disentanglement}, uncertainty measures for \texttt{(B)} and \texttt{(C)} are highly correlated on ID data, which explains their similar performance in settings (ii) and (iii).
Additional details and plots of accuracy rejection curves are provided in Apx.~\ref{apx:sec:detailed_results}.

\textbf{Misclassification detection.}
This task evaluates the ability of an uncertainty measure to distinguish between the set of correct and incorrectly predicted samples.
The setup in this experiment is identical to the selective prediction setup.
We consider the AUROC for distinguishing between correctly and incorrectly predicted datapoints for the different proposed measures of predictive uncertainty as scoring functions.
Alternative metrics commonly used to evaluate misclassification detection, such as AUPR or FPR@TPR95, were also considered in our experiments.
Those induced the same ordering of uncertainty measures, thus we report the AUROC for experiments of this type.

The results are provided in Fig.~\ref{fig:misc}.
We average over the four considered datasets and report means and standard deviations over five independent runs.
The results for individual datasets are reported in Fig.~\ref{fig:misc:det:single}~-~Fig.~\ref{fig:misc:det:posterior} in the appendix.
For detecting the misclassification of (i) the single model, EU \texttt{(A3)} performs best.
However, when predicting with (ii) the average or (iii) a model according to the posterior, TU \texttt{(B/C3)} performs best.
This mirrors the results of the selective prediction task, confirming the importance of aligning the uncertainty measure to the predicting model.

\textbf{Adversarial example detection.}
Additionally, we explored the efficacy of uncertainty measures for detecting adversarial examples on a specific model under FGSM \citep{Goodfellow:15} and PGD \citep{Madry:18} attacks.
Importantly, we do not intend to claim any level of adversarial robustness to these attacks, but use them as a tool to understand the behavior of the uncertainty measures resulting from our framework.
Results and detailed discussion are provided in Apx.~\ref{sec:apx:adversarial} due to space limitations.

\begin{figure*}[t!]
    \centering
    \captionsetup{aboveskip=2pt, belowskip=2pt}
    \begin{subfigure}[b]{\textwidth}
        \includegraphics[width=\textwidth, trim = 0.3cm 0.3cm 0.3cm 1cm, clip]{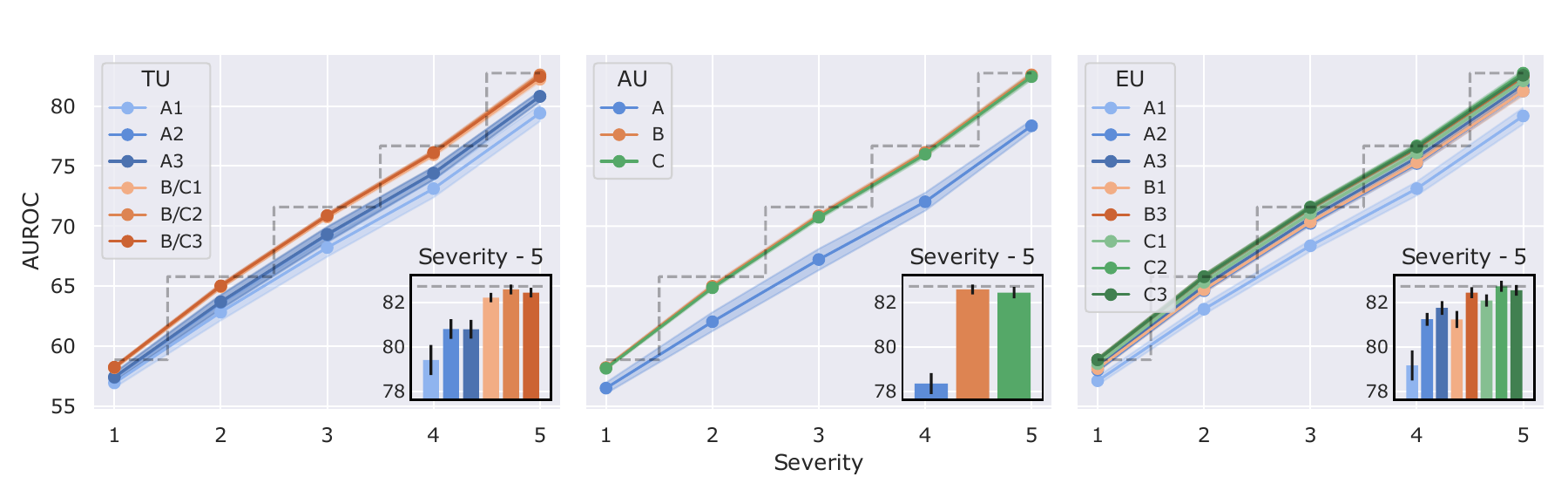}
    \end{subfigure}
    \caption{\textbf{Distribution shift detection on CIFAR10-C.} AUROC for distinguishing between clean and corrupted test datapoints, using the different proposed measures of uncertainty as score. Black dashed line shows the maximum AUROC over all measures per severity. Insets shows detailed results for the highest severity. Statistics over five runs.}
    \label{fig:cifar10c}
\end{figure*}

\subsection{Detecting Distributional Mismatch} \label{sec:ood}

The first set of experiments aimed to assess the correlation between an uncertainty measure and prediction correctness.
This is typically evaluated on i.i.d. test data drawn from the same distribution as the training dataset $\cD$.
However, there is no guarantee that the model will perform reliably on test data from a different distribution. 
Therefore, an effective uncertainty measure should ideally assign high uncertainty to such out-of-distribution (OOD) samples.

\textbf{Out-of-distribution detection.}
We considered standard OOD detection dataset pairings from the literature \citep{Mukhoti:23, Hofmann:24} for our experiments.
Again, we sampled models on CIFAR10/100, SVHN and TIN. 
Then, we use the respective test datasets as ID dataset and the test datasets of the remaining datasets, as well as LSUN, as OOD datasets.
Thus, we consider ID/OOD dataset pairings CIFAR10/CIFAR100, CIFAR10/SVHN, and so on; a total of 16 pairings.
We compare the AUROC for distinguishing between ID and OOD datapoints for each measure within our framework as a scoring function.
Alternative commonly used metrics such as the AUPR and the FPR@TPR95 were also considered.
However, since they induced the same ranking of uncertainty measures, we report the AUROC for the OOD detection experiments.

\begin{figure}[t!]
    \centering
    \includegraphics[width=\linewidth, trim = 0.3cm 0.3cm 0.3cm 1cm, clip]{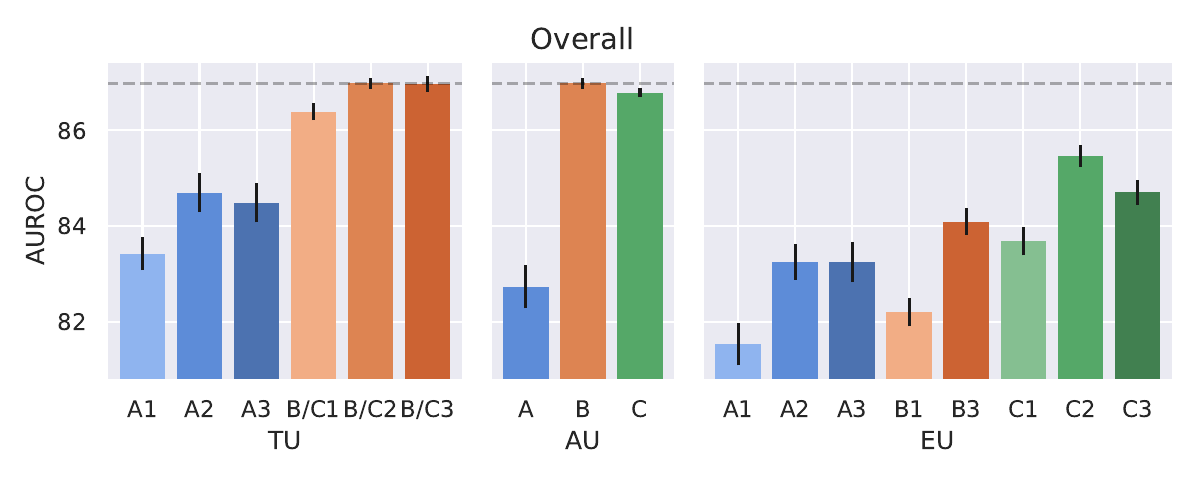}
    \caption{\textbf{OOD detection.} AUROC for distinguishing between ID and OOD datapoints using the different proposed measures of uncertainty as score. TU \texttt{(B/C2)} and TU \texttt{(B/C3)} perform best. AUROCs are averaged over all ID / OOD combinations. Statistics over five runs.}
    \label{fig:ood}
\end{figure}

Results are provided in Fig.~\ref{fig:ood}, showing that TU \texttt{(B/C2)} and TU \texttt{(B/C3)} perform best in this task.
We observe, that across all measures, TU and AU perform better than EU, which is contrary to assumptions commonly formulated in the literature \citep[e.g.,][]{Mukhoti:23, Mucsanyi:24}.
We find, that only on a single ID / OOD dataset combination (TIN/SVHN, see Fig.~\ref{fig:ood:det:tin} in the appendix) out of the 16 pairings we considered, an EU measure performs best.
One possible explanation for the strong performance of AU \texttt{(B)} and \texttt{(C)} could be low noise for these datasets.

\textbf{Distribution shift detection.}
The considered dataset pairings represent comparatively strong OOD-ness.
Therefore, we additionally investigated the performance of the uncertainty measures for detecting distribution shifts on CIFAR10 using the CIFAR10-C \citep{Hendrycks:19} dataset.
The different levels of corruption provided by this dataset can be considered as increasing levels of OOD-ness.
We utilized the 15 main corruptions and excluded the four additional corruptions designated for hyperparameter tuning. 
We report averages over all 15 corruptions in Fig.~\ref{fig:cifar10c} for the five severity levels of corruption provided, showing the AUROC of distinguishing between clean and corrupted versions of the CIFAR10 test dataset using the different uncertainty measures.
The results show that EU measures are more effective than AU or TU measures at intermediate severities but become equally effective for the highest severity.

\subsection{Disentanglement of Measures} \label{sec:disentanglement}

There has been a recent surge of interest in the disentanglement of AU and EU \citep{Valdenegro-toro:22, Mukhoti:23, Mucsanyi:24}.
Furthermore, concerns about the additive decomposition TU = AU + EU has been raised \citep{Wimmer:23}.
Therefore, we investigate the correlation of our measures on both ID and OOD data and conclude with active learning experiments, one of the main tasks requiring disentangled EU measures.

\textbf{Rank correlation of measures.}
Following \citet{Mucsanyi:24}, we investigate the rank correlation of the uncertainty measures in our proposed framework.
We consider the rank correlation (Kendall's $\tau_b$ to account for ties) of measures on both ID and OOD data.
Models are sampled on CIFAR10, thus the ID dataset is the CIFAR10 test dataset, and we consider LSUN as the OOD dataset.

\begin{figure}[t!]
    \centering
    \captionsetup{aboveskip=2pt, belowskip=2pt}
    \includegraphics[width=\linewidth, trim = 0.4cm 0.5cm 1.5cm 1cm, clip]{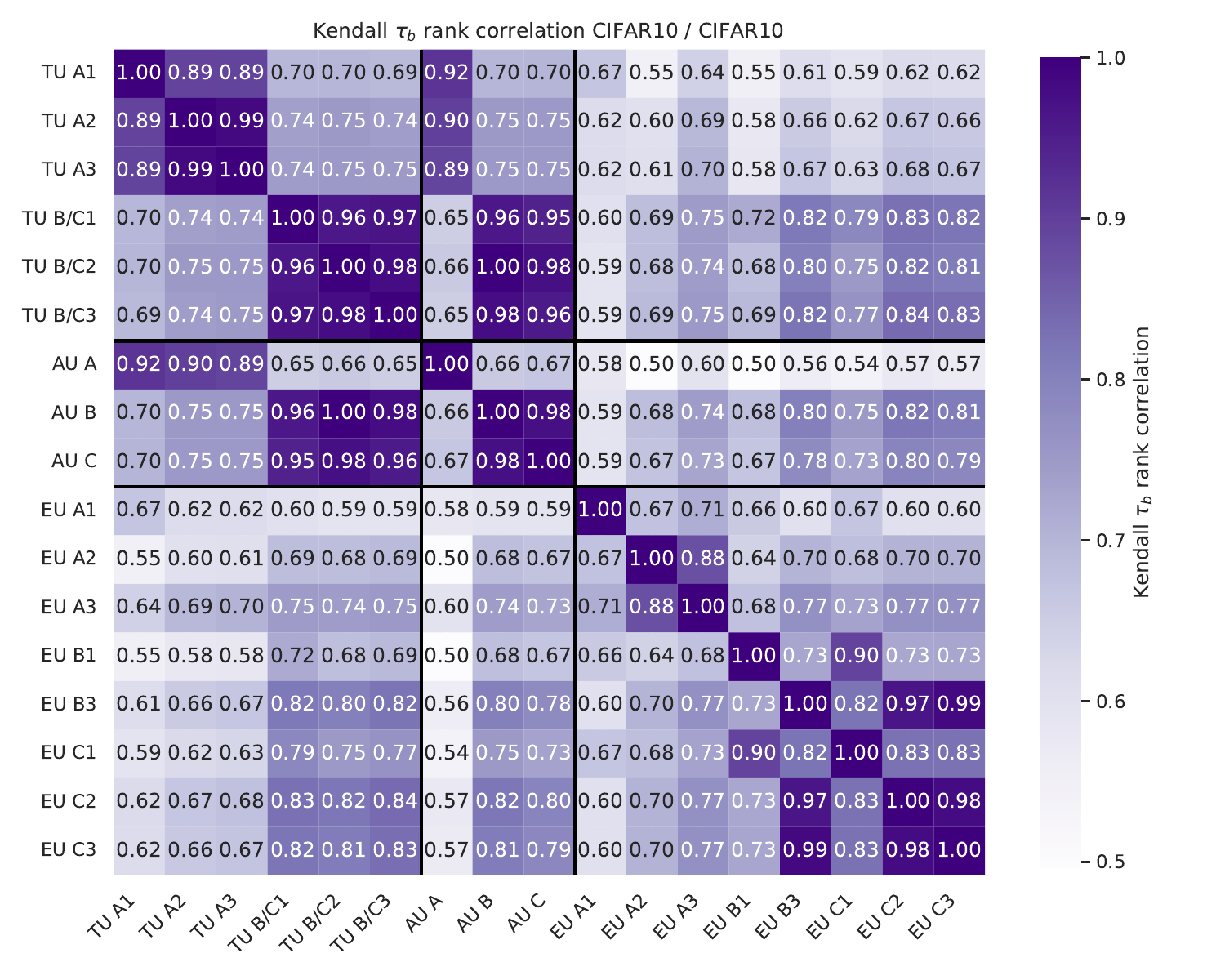}
    \caption{\textbf{Correlation of uncertainty measures on ID dataset (CIFAR10).} High rank correlation blocks exist.}
    \label{fig:correlation_id}
\end{figure}

The results for the ID dataset are shown in Fig.~\ref{fig:correlation_id}.
We observe, that for TU and AU measures, there are two highly correlated blocks.
Those measures that depend on a single predicting model and the rest.
While there is a stark contrast between considering the single predicting model and the rest, there are no real differences between considering the average model as predicting model and each model according to the posterior distribution.
Furthermore, the approximation of the true predictive distribution \texttt{(1,2,3)} does not seem to play a major role for TU and AU on the ID dataset.
For EU, there is a block of highly correlated measures, \texttt{(B3)}, \texttt{(C2)} and \texttt{(C3)}.
This can be explained due to their additive relationship (see Apx.~\ref{sec:apx:relationship_epistemic}).
The same holds for the strong correlation between EU \texttt{(B1)} and \texttt{(C1)}, which share a similar relationship (again, see Apx.~\ref{sec:apx:relationship_epistemic}).
Finally, EU \texttt{(A2)} and \texttt{(A3)} have a high correlation. 
Although they do not share a similar relationship as the other blocks of high correlation for EU measures, EU \texttt{(A3)} is an upper bound of \texttt{(A2)}.

\begin{figure}[t!]
    \centering
    \captionsetup{aboveskip=2pt, belowskip=2pt}
    \includegraphics[width=\linewidth, trim = 0.4cm 0.5cm 1.5cm 1cm, clip]{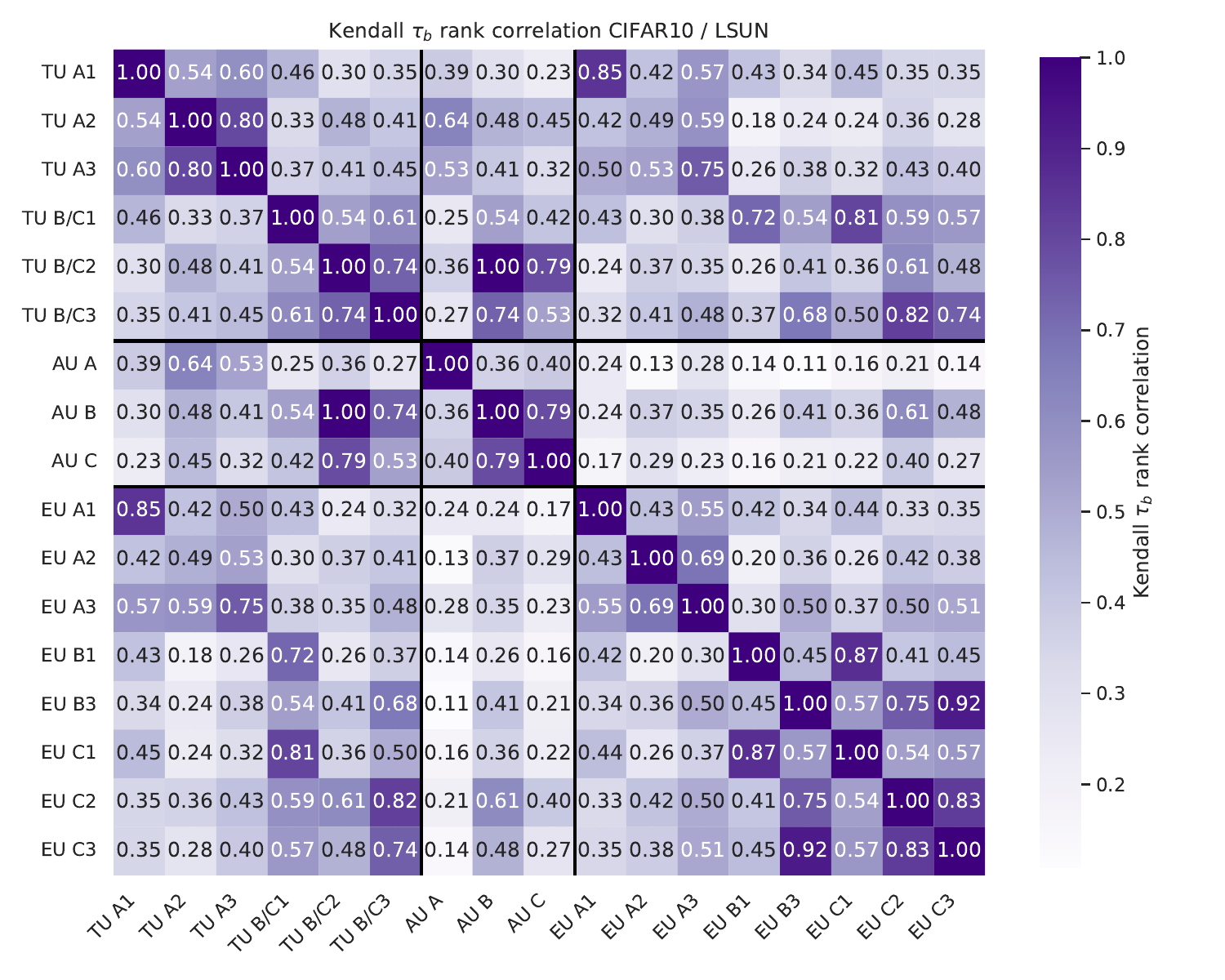}
    \caption{\textbf{Correlation of uncertainty measures on OOD dataset (LSUN).} Rank correlations are very low overall.}
    \label{fig:correlation_ood}
\end{figure}

The results for the OOD dataset are shown in Fig.~\ref{fig:correlation_ood}.
Here, we do not observe similarly high rank correlations as for the ID dataset.
The strongest rank correlation exhibited is those for EU \texttt{(B3)}, \texttt{(C2)} and \texttt{(C3)}, as well as EU \texttt{(B1)} and \texttt{(C1)}.
Note that TU \texttt{(B/C3)} and AU \texttt{(B)} are equivalent, thus perfectly correlated.
In general, we observe a very distinct disentanglement of uncertainty measures in our proposed framework.
Although there are blocks of high correlation for ID data, we mostly do not observe such strong correlations for OOD data.

\textbf{Active Learning.}
Finally, we investigated the proposed framework of uncertainty measures on active learning tasks.
Given the computational complexity of the active learning setting, we utilized different datasets and models as for the rest of our experiments.
Specifically, we used the MNIST \citep{LeCun:98} and FMNIST \citep{Xiao:17} datasets and a small CNN. 
Details on the network architecture and training procedure are provided in Apx.~\ref{apx:active_learning}.
We obtained five posterior samples (ensemble members) in each acquisition round.
The average over the predictive distributions of those sampled models, the approximated posterior predictive, was used to calculate the accuracies for each acquisition step. 
The same sampled models were used to approximate the respective uncertainty measures as acquisition functions to select the next datapoints from the pool dataset to transfer to the training dataset. 

For MNIST, we started with 20 datapoints in the training dataset and the remaining 49,980 datapoints in the pool dataset.
Those 20 datapoints were balanced so that two datapoints from each class were contained.
In each iteration, the five samples with the highest uncertainty were selected from the pool dataset and added to the training dataset.
We considered TU, AU and EU for measures \texttt{(B2)}, \texttt{(B3)}, \texttt{(C2)} and \texttt{(C3)} as acquisition functions, as well as random selection as a baseline.
We did not investigate measures \texttt{(A1)}, \texttt{(A2)}, \texttt{(A3)}, \texttt{(B1)} and \texttt{(C1)} due to the long runtimes of the experiments, but would expect them to perform worse than those considered in light of the other experiments we conducted.
However, an interesting situation could be EU \texttt{(A1)} when training a single model on the dataset in the current iteration and comparing the model from the previous iteration.
Future work should investigate this setting, e.g. in transfer learning scenarios.

\begin{figure}
    \centering
    \captionsetup{aboveskip=4pt, belowskip=2pt}
    \includegraphics[width=\linewidth, trim = 0.3cm 0.3cm 0.3cm 1cm, clip]{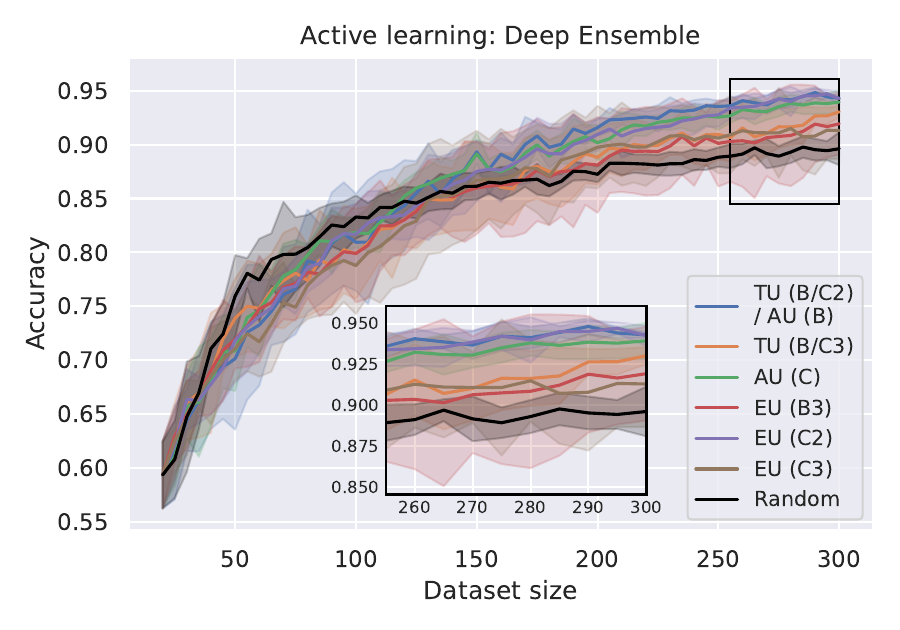}
    \caption{\textbf{Active learning on MNIST.} EU \texttt{(C3)} and EU \texttt{(B3)} perform worst overall. The random baseline performs strong for small dataset sizes. The accuracy is those of the posterior predictive. Statistics over five runs.}
    \label{fig:active_learning_mnist}
\end{figure}

The results are given in Fig.~\ref{fig:active_learning_mnist}.
We observe that EU \texttt{(C2)}, the mutual information, leads to the best performance for any of the EU measures.
Interestingly, we find TU \texttt{(B/C2)} which is identical to AU \texttt{(B)} to be equally well performing.
The same is found for TU \texttt{(B/C3)}.
EU \texttt{(B3)} and EU \texttt{(C3)} are found to be the worst performing measures as acquisition functions, contrary to the sentiment that EU estimators should perform best in this task \citep[e.g.][]{Gal:17, Mukhoti:23}.
In contrast, AU \texttt{(C)}, performs very well as an acquisition function.
The random sampling baseline is also very effective as an acquisition function until the training dataset size reaches around 100 samples, more effective than any of the considered uncertainty measures.
We hypothesize that until a certain dataset size, models sampled from the posterior are not specified enough and provide no good signal of what datapoints to add next, in fact even the contrary as they are performing worse than random.
This behavior would be interesting to investigate in more details in future experiments.

For FMNIST, we started with 1000 datapoints in the training dataset and the remaining 49,000 datapoints in the pool dataset.
The initial training dataset was balanced such that 100 datapoints from each class were included.
In each iteration, the 15 samples with the highest uncertainty are selected from the pool dataset and added to the training dataset.
As for the MNIST experiment, we considered TU, AU and EU for measures \texttt{(B2)}, \texttt{(B3)}, \texttt{(C2)} and \texttt{(C3)} as acquisition functions, as well as random selection as baseline.

\begin{figure}
    \centering
    \captionsetup{aboveskip=4pt, belowskip=2pt}
    \includegraphics[width=\linewidth, trim = 0.3cm 0.3cm 0.3cm 1cm, clip]{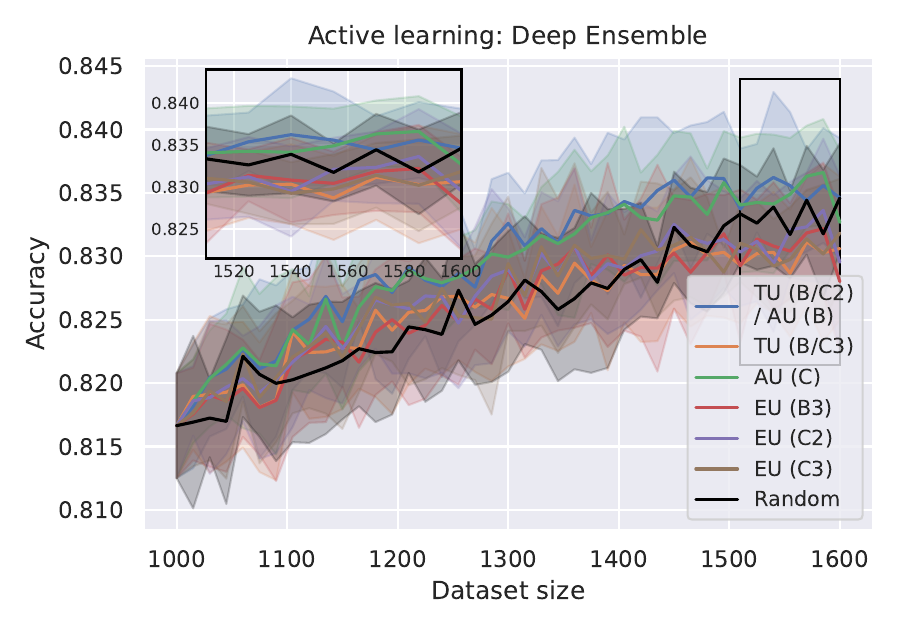}
    \caption{\textbf{Active learning on FMNIST.} All three EU measures perform worse than TU \texttt{(B/C2)} / AU \texttt{(B)} and AU \texttt{(C)}. The accuracy is those of the posterior predictive. Statistics over ten runs.}
    \label{fig:active_learning_fmnist}
\end{figure}

The results are provided in Fig.~\ref{fig:active_learning_fmnist}.
We find that all EU measures including the mutual information EU \texttt{(C2)}, are outperformed by TU \texttt{(B/C2)} which is identical to AU \texttt{(B)}, and AU \texttt{(C)}.
Until around 1500 samples, the random baseline performs roughly on par with the EU measures, afterwards it even performs slightly better.
Although both datasets are relatively simple, FMNIST proved notably more challenging for the active learning pipeline and has higher variance between runs.
This is likely due to its increased complexity and less clear-cut class boundaries.

\section{Conclusion}

In this work, we introduced a unifying framework to categorize information-theoretic measures of predictive uncertainty.
By systematically considering the choice of the predicting model and the approximation of the true predictive distribution, we derived a family of uncertainty measures that encompasses existing measures and introduces new ones.
This framework clarifies the relationships between different uncertainty measures and their implicit assumptions, offering a more structured way of understanding them and applying them in practice.

Our empirical evaluations demonstrated the importance of aligning the uncertainty measure with the predicting model.
We found that different measures perform optimally, depending on the predicting model, emphasizing the need for careful selection based on the intended application. 
Furthermore, our results challenge prevailing assumptions in the literature, such as the effectiveness of EU measures for OOD detection, showing that AU often plays a dominant role.
We also investigated the degree of disentanglement between the measures of our proposed framework. 
Our findings indicate that, while there are strong correlations between specific uncertainty measures for ID data, correlations weaken significantly in OOD scenarios.
This suggests that the disentanglement of measures may be highly context-dependent, which should be investigated in depth in the future.
Future work should also explore extensions of our framework to autoregressive models. Here, uncertainty estimation plays a crucial role in addressing hallucinations, particularly a subset known as confabulations, and in ensuring reliable text generation \citep{Xiao:21, Yadkori:24, Aichberger:24, Farquhar:24}.

Overall, our work provides a clear conceptual foundation for understanding predictive uncertainty measures and offers actionable insights for their selection for a given task.

\begin{acknowledgements} 

We thank Andreas Kirsch for constructive feedback on an earlier version of this work.

The ELLIS Unit Linz, the LIT AI Lab, the Institute for Machine Learning, are supported by the Federal State Upper Austria. We thank the projects FWF AIRI FG 9-N (10.55776/FG9), AI4GreenHeatingGrids (FFG- 899943), Stars4Waters (HORIZON-CL6-2021-CLIMATE-01-01), FWF Bilateral Artificial Intelligence (10.55776/COE12). We thank NXAI GmbH, Audi AG, Silicon Austria Labs (SAL), Merck Healthcare KGaA, GLS (Univ. Waterloo), T\"{U}V Holding GmbH, Software Competence Center Hagenberg GmbH, dSPACE GmbH, TRUMPF SE + Co. KG.

\end{acknowledgements}

\bibliography{literature}

\newpage

\onecolumn

\title{On Information-Theoretic Measures of Predictive Uncertainty\\(Supplementary Material)}
\maketitle
\appendix
\renewcommand{\thesubsection}{\thesection.\arabic{subsection}}

\section{Technical Details}

In this section, we provide additional details on the relationship between epistemic components as well as the MC approximations for all measures.
Furthermore, we discuss potential generalizations of our framework to the Rényi cross-entropy and other proper scoring rules.
Finally, we discuss the alternative order of arguments for Eq.~\eqref{eq:cross_entropy} and its implications for interpreting the resulting uncertainty measures as well as the regression setting.

\subsection{Relationships between Epistemic Components} \label{sec:apx:relationship_epistemic}

\cite{Schweighofer:23b} proved the relationship that the sum of the EU of C2 and B3 is equivalent to the EU of C3.
For completeness, we provide a version of the proof as follows:

\begin{align} \label{eq:derivation_c3}
    &\overbrace{\EXP_{p(\Bw \mid \cD)} \left[\KL{p(y \mid \Bx, \Bw)}{p(y \mid \Bx, \cD)} \right]}^{\text{EU \texttt{(C2)} - Mutual Information}} \ + \ \overbrace{\EXP_{p(\tilde{\Bw} \mid \cD )} \left[ \KL{p(y \mid \Bx, \cD)}{p(y \mid \Bx, \tilde\Bw)} \right]}^{\text{EU \texttt{(B3)}}} \\
    \label{eq:derivation_c3:step_2}
    &\quad = \ \EXP_{p(\Bw \mid \cD)} \left[ \EXP_{p(y \mid \Bx, \Bw)} \left[ \log \frac{p(y \mid \Bx, \Bw)}{p(y \mid \Bx, \cD)} \right] \right] \ + \ \EXP_{p(\tilde\Bw \mid \cD)} \left[ \EXP_{p(y \mid \Bx, \cD)} \left[ \log \frac{p(y \mid \Bx, \cD)}{p(y \mid \Bx, \tilde\Bw)} \right] \right] \\
    \label{eq:derivation_c3:step_3}
    &\quad = \ \EXP_{p(\Bw \mid \cD)} \left[ \EXP_{p(y \mid \Bx, \Bw)} \left[ \log p(y \mid \Bx, \Bw) \right] \ - \ \EXP_{p(y \mid \Bx, \Bw)} \left[ \log {p(y \mid \Bx, \cD)} \right] \right] \ + \\ \nonumber
    &\qquad\;\; \EXP_{p(\tilde\Bw \mid \cD)} \left[ \EXP_{p(y \mid \Bx, \cD)} \left[ \log {p(y \mid \Bx, \cD)} \right] \ - \ \EXP_{p(y \mid \Bx, \cD)} \left[ \log p(y \mid \Bx, \tilde\Bw) \right] \right] \\
    \label{eq:derivation_c3:step_4}
    &\quad = \ \EXP_{p(\Bw \mid \cD)} \left[ \EXP_{p(y \mid \Bx, \Bw)} \left[ \log p(y \mid \Bx, \Bw) \right] \right] \ - \ \cancel{\EXP_{p(y \mid \Bx, \cD)} \left[ \log {p(y \mid \Bx, \cD)} \right]} \ + \\ \nonumber
    &\qquad\;\; \cancel{\EXP_{p(y \mid \Bx, \cD)} \left[ \log {p(y \mid \Bx, \cD)} \right]} \ - \ \EXP_{p(\tilde\Bw \mid \cD)} \left[ \EXP_{p(y \mid \Bx, \cD)} \left[ \log p(y \mid \Bx, \tilde\Bw) \right] \right] \\
    \label{eq:derivation_c3:step_5}
    &\quad = \ \EXP_{p(\Bw \mid \cD)} \left[ \EXP_{p(y \mid \Bx, \Bw)} \left[ \log p(y \mid \Bx, \Bw) \right] \right] \ - \\ \nonumber 
    &\qquad\;\; \EXP_{p(\tilde\Bw \mid \cD)} \left[ \EXP_{p(\Bw \mid \cD)} \left[ \EXP_{p(y \mid \Bx, \Bw)} \left[ \log p(y \mid \Bx, \tilde\Bw) \right] \right] \right] \\
    \label{eq:derivation_c3:step_6}
    &\quad = \ \EXP_{p(\Bw \mid \cD)} \left[ \EXP_{p(\tilde\Bw \mid \cD)} \left[ \EXP_{p(y \mid \Bx, \Bw)} \left[ \log p(y \mid \Bx, \Bw) \right] \right] \right] \ - \\ \nonumber 
    &\qquad\;\; \EXP_{p(\Bw \mid \cD)} \left[ \EXP_{p(\tilde\Bw \mid \cD)} \left[ \EXP_{p(y \mid \Bx, \Bw)} \left[ \log p(y \mid \Bx, \tilde\Bw) \right] \right] \right] \\
    \label{eq:derivation_c3:step_7}
    &\quad = \ \EXP_{p(\Bw \mid \cD)} \left[ \EXP_{p(\tilde\Bw \mid \cD)} \left[ \EXP_{p(y \mid \Bx, \Bw)} \left[ \log \frac{p(y \mid \Bx, \Bw)}{p(y \mid \Bx, \tilde\Bw)} \right] \right] \right] \\
    \label{eq:derivation_c3:step_8}
    &\quad = \ \underbrace{\EXP_{p(\Bw \mid \cD)} \left[\EXP_{p(\tilde\Bw \mid \cD)} \left[ \KL{p(y \mid \Bx, \Bw)}{p(y \mid \Bx, \tilde\Bw)} \right] \right]}_{\text{EU \texttt{(C3)}}} \ ,
\end{align}

which is what we wanted to show.
The step from \eqref{eq:derivation_c3:step_3} to \eqref{eq:derivation_c3:step_4} is due to additivity and linearity of expectations.
The step from \eqref{eq:derivation_c3:step_5} to \eqref{eq:derivation_c3:step_6} is due to the fact that we can insert the expectation $\EXP_{p(\tilde\Bw \mid \cD)}$ in the first term as it does not depend on $\tilde\Bw$ and due to the fact that $p(\tilde\Bw \mid \cD) = p(\Bw \mid \cD)$.\\
\qed

Furthermore, a similar proof can be constructed for EU \texttt{(C1)} = EU \texttt{(C2)} + EU \texttt{(B1)} as follows:

\begin{align} \label{eq:derivation_c1}
    &\overbrace{\EXP_{p(\Bw \mid \cD)} \left[\KL{p(y \mid \Bx, \Bw)}{p(y \mid \Bx, \cD)} \right]}^{\text{EU \texttt{(C2)} - Mutual Information}} \ + \ \overbrace{\KL{p(y \mid \Bx, \cD)}{p(y \mid \Bx, \tilde\Bw)}}^{\text{EU \texttt{(B1)}}} \\
    \label{eq:derivation_c1:step_2}
    &\quad = \ \EXP_{p(\Bw \mid \cD)} \left[ \EXP_{p(y \mid \Bx, \Bw)} \left[ \log \frac{p(y \mid \Bx, \Bw)}{p(y \mid \Bx, \cD)} \right] \right] \ + \ \EXP_{p(y \mid \Bx, \cD)} \left[ \log \frac{p(y \mid \Bx, \cD)}{p(y \mid \Bx, \tilde\Bw)} \right] \\
    \label{eq:derivation_c1:step_3}
    &\quad = \ \EXP_{p(\Bw \mid \cD)} \left[ \EXP_{p(y \mid \Bx, \Bw)} \left[ \log p(y \mid \Bx, \Bw) \right] \ - \ \EXP_{p(y \mid \Bx, \Bw)} \left[ \log {p(y \mid \Bx, \cD)} \right] \right] \ + \\ \nonumber
    &\qquad\;\; \EXP_{p(y \mid \Bx, \cD)} \left[ \log {p(y \mid \Bx, \cD)} \right] \ - \ \EXP_{p(y \mid \Bx, \cD)} \left[ \log p(y \mid \Bx, \tilde\Bw) \right]\\
    \label{eq:derivation_c1:step_4}
    &\quad = \ \EXP_{p(\Bw \mid \cD)} \left[ \EXP_{p(y \mid \Bx, \Bw)} \left[ \log p(y \mid \Bx, \Bw) \right] \right] \ - \ \cancel{\EXP_{p(y \mid \Bx, \cD)} \left[ \log {p(y \mid \Bx, \cD)} \right]} \ + \\ \nonumber
    &\qquad\;\; \cancel{\EXP_{p(y \mid \Bx, \cD)} \left[ \log {p(y \mid \Bx, \cD)} \right]} \ - \ \EXP_{p(y \mid \Bx, \cD)} \left[ \log p(y \mid \Bx, \tilde\Bw) \right] \\
    \label{eq:derivation_c1:step_5}
    &\quad = \ \EXP_{p(\Bw \mid \cD)} \left[ \EXP_{p(y \mid \Bx, \Bw)} \left[ \log p(y \mid \Bx, \Bw) \right] \right] \ - \ \EXP_{p(\Bw \mid \cD)} \left[ \EXP_{p(y \mid \Bx, \Bw)} \left[ \log p(y \mid \Bx, \tilde\Bw) \right] \right] \\
    \label{eq:derivation_c1:step_6}
    &\quad = \ \EXP_{p(\Bw \mid \cD)} \left[ \EXP_{p(y \mid \Bx, \Bw)} \left[ \log \frac{p(y \mid \Bx, \Bw)}{p(y \mid \Bx, \tilde\Bw)} \right] \right]\\
    \label{eq:derivation_c1:step_7}
    &\quad = \ \underbrace{\EXP_{p(\Bw \mid \cD)} \left[ \KL{p(y \mid \Bx, \Bw)}{p(y \mid \Bx, \tilde\Bw)} \right]}_{\text{EU \texttt{(C1)}}} \ ,
\end{align}

which is what we wanted to show. 
Again, the step from \eqref{eq:derivation_c1:step_3} to \eqref{eq:derivation_c1:step_4} is due to additivity and linearity of expectations.
The linearity property is used to get to \eqref{eq:derivation_c1:step_5}, after which elementary algebra leads to the result.\\
\qed

In the same fashion, it is possible to construct a proof for EU \texttt{(C2)} = EU \texttt{(C2)} + EU \texttt{(B2)}.
However, as we know that EU \texttt{(B2)} = 0, this is trivial.

\subsection{Monte Carlo Approximations} \label{apx:sec:mc_approx}

The measures we proposed through our framework, except for measure \texttt{(A1)}, incorporate posterior expectations $\EXP_{p(\Bw \mid \cD)} \left[ \cdot \right]$.
These are generally intractable to calculate exactly and are thus approximated through samples drawn from the distribution - a Monte Carlo approximation of the expectation.
In this section we provide those approximations explicitly and discuss efficient ways to implement them, utilizing relationships between individual measures.

We assume that the posterior $p(\Bw \mid \cD)$ models to predict are drawn from and the posterior $p(\tilde{\Bw} \mid \cD)$ approximations of the true model are drawn from are the same.
However, in practice it is generally the case that models for averaging are selected based on their accuracy on a validation set, or more generally that they are selected in a way optimal for predicting well.
When sampling potential true models that are likely under the data, the functional diversity of samples is often of concern, e.g. as done with the sampling algorithm in \cite{Schweighofer:23a}.
This can be seen as either having different posteriors due to different priors or having different algorithms to obtain samples from the same posterior.
However, if the samples are not the same, many of the relationships in Sec.~\ref{sec:relationship} are no longer valid, thus also the coloring in Tab.~\ref{tab:framework_main}.
The provided implementation is able to handle both the case where different samples are used for the MC approximation as well as the case where the same samples are used and thus more efficient computation of TU \texttt{(C3)} and EU \texttt{(C3)} is possible.
The MC approximations for a single set of samples $\{\Bw_n\}_{n=1}^N$ from $p(\Bw \mid \cD) = p(\tilde\Bw \mid \cD)$ are given by

\textbf{TU \texttt{(A2)}:}
\begin{align} \label{eq:tu_mc_a2}
    &\CE{p(\By \mid \Bx, \Bw)}{p(\By \mid \Bx, \cD)} \ = \ \CE{p(\By \mid \Bx, \Bw)}{ \EXP_{\tilde\Bw} \left[ p(\By \mid \Bx, \tilde\Bw) \right]} \\ \nonumber
    &\qquad \approx \ \CE{p(\By \mid \Bx, \Bw)}{\frac{1}{N} \sum_{n=1}^N p(\By \mid \Bx, \Bw_n )}, \qquad \Bw_n \sim p(\Bw \mid \cD)
\end{align}

\textbf{TU \texttt{(A3)}:}
\begin{align} \label{eq:tu_mc_a3}
    &\EXP_{\tilde\Bw} \left[ \CE{p(\By \mid \Bx, \Bw)}{p(\By \mid \Bx, \tilde\Bw)} \right] \\ \nonumber
    &\qquad \approx \ \frac{1}{N} \sum_{n=1}^N  \CE{ p(\By \mid \Bx, \Bw)}{p(\By \mid \Bx, \Bw_n)}, \qquad \Bw_n \sim p(\Bw \mid \cD)
\end{align}

\textbf{TU \texttt{(B/C1)}:}
\begin{align} \label{eq:tu_mc_bc1}
    &\CE{p(\By \mid \Bx, \cD)}{p(\By \mid \Bx, \tilde\Bw)} \ = \ \EXP_{\Bw} \left[ \CE{p(\By \mid \Bx, \Bw)}{p(\By \mid \Bx, \tilde\Bw)} \right] \\ \nonumber
    &\qquad \approx \ \frac{1}{N} \sum_{n=1}^N  \CE{ p(\By \mid \Bx, \Bw_n)}{p(\By \mid \Bx, \tilde\Bw)}, \qquad \Bw_n \sim p(\Bw \mid \cD)
\end{align}

\textbf{TU \texttt{(B/C2)}:}
\begin{align} \label{eq:tu_mc_bc2}
    &\CE{p(\By \mid \Bx, \cD)}{p(\By \mid \Bx, \cD)} \ = \ \EXP_{\Bw} \left[ \CE{p(\By \mid \Bx, \Bw)}{ \EXP_{\tilde\Bw} \left[p(\By \mid \Bx, \tilde\Bw)\right]} \right] \\ \nonumber
    &\qquad \approx \ \frac{1}{N} \sum_{n=1}^N \CE{p(\By \mid \Bx, \Bw_n)}{ \frac{1}{N} \sum_{n=1}^N p(\By \mid \Bx, \Bw_n)}, \qquad \Bw_n \sim p(\Bw \mid \cD)
\end{align}

\textbf{TU \texttt{(B/C3)}:}
\begin{align} \label{eq:tu_mc_bc3}
    &\EXP_{\tilde\Bw} \left[ \CE{(p(\By \mid \Bx, \cD )}{p(\By \mid \Bx, \tilde\Bw )} \right] \ = \ \EXP_{\Bw} \left[ \EXP_{\tilde\Bw} \left[ \CE{p(\By \mid \Bx, \Bw)}{p(\By \mid \Bx, \tilde\Bw)} \right] \right] \\ \nonumber
    & \qquad \approx \ \frac{1}{N (N-1)} \sum_{n=1}^N \sum_{n'=1}^N  \CE{p(\By \mid \Bx, \Bw_n )}{p(\By \mid \Bx, \Bw_{n'} )}, \qquad \Bw_n \sim p(\Bw \mid \cD)
\end{align}

\textbf{AU \texttt{(B)}:}
\begin{align} \label{eq:au_mc_b}
    &\ENT(p(\By \mid \Bx, \cD )) \ = \ \ENT( \EXP_{\Bw} \left[ p(\By \mid \Bx, \Bw ) \right] ) \\ \nonumber
    &\qquad \approx \ \ENT (\frac{1}{N} \sum_{n=1}^N p(\By \mid \Bx, \Bw_n )), \qquad \Bw_n \sim p(\Bw \mid \cD)
\end{align}

\textbf{AU \texttt{(C)}:}
\begin{align} \label{eq:au_mc_c}
    &\EXP_{\Bw} \left[\ENT(p(\By \mid \Bx, \Bw )) \right] \ \approx \ \frac{1}{N} \sum_{n=1}^N \ENT(p(\By \mid \Bx, \Bw_n ) ), \qquad \Bw_n \sim p(\Bw \mid \cD)
\end{align}

\textbf{EU \texttt{(A2)}:}
\begin{align} \label{eq:eu_mc_a2}
    &\KL{p(\By \mid \Bx, \Bw )}{p(\By \mid \Bx, \cD)} \ = \ \KL{p(\By \mid \Bx, \Bw )}{\EXP_{\tilde\Bw} \left[p(\By \mid \Bx, \tilde\Bw )\right]}  \\ \nonumber 
    &\qquad \approx \ \KL{p(\By \mid \Bx, \Bw)}{ \frac{1}{N} \sum_{n=1}^N p(\By \mid \Bx, \Bw_n)}, \qquad \Bw_n \sim p(\Bw \mid \cD)
\end{align}

\textbf{EU \texttt{(A3)}:}
\begin{align} \label{eq:eu_mc_a3}
    &\EXP_{\tilde\Bw} \left[ \KL{p(\By \mid \Bx, \Bw )}{p(\By \mid \Bx, \tilde\Bw )} \right] \\ \nonumber 
    &\qquad \approx \ \frac{1}{N} \sum_{n=1}^N  \KL{p(\By \mid \Bx, \Bw)}{p(\By \mid \Bx, \Bw_n)}, \qquad \Bw_n \sim p(\Bw \mid \cD)
\end{align}

\textbf{EU \texttt{(B1)}:}
\begin{align} \label{eq:eu_mc_b1}
    &\KL{p(\By \mid \Bx, \cD)}{p(\By \mid \Bx, \tilde\Bw)} \ = \ \KL{\EXP_{\Bw} \left[p(\By \mid \Bx, \Bw )\right]}{p(\By \mid \Bx, \tilde\Bw )} \\ \nonumber 
    &\qquad \approx \ \KL{\frac{1}{N} \sum_{n=1}^N p(\By \mid \Bx, \Bw_n)}{  p(\By \mid \Bx, \tilde\Bw)}, \qquad \Bw_n \sim p(\Bw \mid \cD)
\end{align}

\textbf{EU \texttt{(B3)}:}
\begin{align} \label{eq:eu_mc_b3}
    &\EXP_{\tilde\Bw} \left[ \KL{p(\By \mid \Bx, \cD)}{p(\By \mid \Bx, \tilde\Bw)} \right] \ = \ \EXP_{\tilde\Bw} \left[ \KL{\EXP_{\Bw} \left[p(\By \mid \Bx, \Bw ) \right]}{p(\By \mid \Bx, \tilde\Bw)} \right] \\ \nonumber 
    &\qquad \approx \ \frac{1}{N} \sum_{n=1}^N  \KL{\frac{1}{N} \sum_{n=1}^N p(\By \mid \Bx, \Bw_n )}{p(\By \mid \Bx, \Bw_n )}, \qquad \Bw_n \sim p(\Bw \mid \cD)
\end{align}

\textbf{EU \texttt{(C1)}:}
\begin{align} \label{eq:eu_mc_c1}
    &\EXP_{\Bw} \left[ \KL{p(\By \mid \Bx, \Bw )}{p(\By \mid \Bx, \tilde\Bw )} \right] \\ \nonumber 
    &\qquad \approx \ \frac{1}{N} \sum_{n=1}^N  \KL{p(\By \mid \Bx, \Bw_n )}{p(\By \mid \Bx, \Bw)}, \qquad \Bw_n \sim p(\Bw \mid \cD)
\end{align}

\textbf{EU \texttt{(C2)}:}
\begin{align} \label{eq:eu_mc_c2}
    &\EXP_{\Bw} \left[ \KL{p(\By \mid \Bx, \Bw )}{p(\By \mid \Bx, \cD )} \right] \ = \ \EXP_{\Bw} \left[ \KL{p(\By \mid \Bx, \Bw )}{\EXP_{\tilde\Bw} \left[p(\By \mid \Bx, \tilde\Bw)\right]} \right] \\ \nonumber 
    &\qquad \approx \ \frac{1}{N} \sum_{n=1}^N \KL{p(\By \mid \Bx, \Bw_n )}{\frac{1}{N} \sum_{n=1}^N p(\By \mid \Bx, \Bw_n )}, \qquad \Bw_n \sim p(\Bw \mid \cD)
\end{align}

\textbf{EU \texttt{(C3)}:}
\begin{align} \label{eq:eu_mc_c3}
    &\EXP_{\Bw} \left[ \EXP_{\tilde\Bw} \left[ \KL{p(\By \mid \Bx, \Bw)}{p(\By \mid \Bx, \tilde\Bw)} \right] \right] \\ \nonumber 
    &\qquad \approx \ \frac{1}{N (N-1)} \sum_{n=1}^N \sum_{n'=1}^N  \KL{p(\By \mid \Bx, \Bw_n )}{p(\By \mid \Bx, \Bw_{n'} )}, \qquad \Bw_n \sim p(\Bw \mid \cD)
\end{align}

\subsection{Generalization to Renyi Cross-Entropy} \label{apx:sec:renyi}

In this section we review the Rényi cross-entropy which is a generalization of the cross-entropy discussed in the main paper.
This allows to directly transfer our proposed measure of predictive uncertainty in Eq.~\eqref{eq:cross_entropy} and the framework we introduced based on it (overview in Tab.~\ref{tab:framework_main}) to other instances of Rényi cross-entropies.

Let us start with the Rényi entropy, which was proposed as a generalization of the Shannon entropy, in that for the limit of the Rényi parameter $\alpha \rightarrow 1$ the Rényi entropy becomes the Shannon entropy.
For two discrete distributions $p$ and $q$ on the same support $\mathcal{Y}$ it is defined as
\begin{align}
    \ENT_\alpha(p) \ = \ \frac{1}{1 - \alpha} \log \sum_i p_i^\alpha
\end{align}
Similarly, the Rényi divergence is a generalization of the Kullback-Leibler (KL) divergence, in that for the limit of the Rényi parameter $\alpha \rightarrow 1$ the Rényi divergence becomes the KL divergence.
It is defined as
\begin{align}
    \rD_\alpha(p \mid\mid q) \ = \ \frac{1}{\alpha - 1} \log \sum_i p_i^\alpha \ q_i^{1 - \alpha}
\end{align}
Note that there are also versions of both for continuous distributions, basically exchanging the sum with an integral.
However, the resulting Rényi differential entropy shares the same deficiencies as the Shannon differential entropy.

What is left is defining the Rényi cross-entropy.
Motivated by the additive decomposition of Shannon cross-entropy into the entropy and KL divergence, \cite{Sarraf:21} proposed to define the Rényi cross-entropy as
\begin{align} \label{eq:renyi_cross_entropy}
    \mathrm{CE}_\alpha(p \ ; \ q) \ := \ \ENT_\alpha(p) \ + \ \rD_\alpha(p \mid\mid q)
\end{align}
Multiple closed form solutions for different values of $\alpha$ are already known for the Rényi entropy and divergence, making this a very simple solution.
Furthermore, \cite{Valverde:19} introduced a closed form solution, which has been simplified to the following form by \cite{Thierrin:22}:
\begin{align}
    \mathrm{CE}_\alpha(p \ ; \ q) \ := \frac{1}{1 - \alpha} \log \sum_i p_i \ q_i^{\alpha - 1}
\end{align}
Furthermore, \cite{Thierrin:22} proposes closed form solutions for this form of the Rényi differential cross-entropy for various continuous distributions.

In the following we stick to the definition of the Rényi cross-entropy by \cite{Sarraf:21} (Eq.~\eqref{eq:renyi_cross_entropy}) and state the respective entropy and divergence for special cases of $\alpha$.
By defining the arbitrary discrete distributions as $p := p(y \mid \Bx, \cdot)$ and $q := p(y \mid \Bx, \Bw^*)$ each value of $\alpha$ yields a variant our proposed measure of predictive uncertainty (Eq.~\eqref{eq:cross_entropy}), giving rise to variants of our proposed framework.

$\alpha = 0$: The measure of entropy is called the Hartley or max-entropy, which is the cardinality of possible events $\cY$.
It is given by
\begin{align}
    \ENT_0(p) := \log |\cY| \ .
\end{align}
The divergence is called max-divergence and is given by
\begin{align}
    \mathrm{D}_0(p \mid\mid q) := - \log Q(\{i : p_i > 0\}) \ .
\end{align}

$\alpha = \frac{1}{2}$: The measure of entropy is referred to as Bhattacharyya-entropy.
It is given by
\begin{align}
    \ENT_{\frac{1}{2}}(p) := 2 \log \sum_i \sqrt{p_i} \ .
\end{align}
The divergence is called Bhattacharyya-divergence (minus twice the logarithm of the Bhattacharyya coefficient) and is given by
\begin{align}
    \mathrm{D}_{\frac{1}{2}}(p \mid\mid q) :=  - 2 \log \sum_i \sqrt{p_i q_i} \ .
\end{align}

$\alpha = 1$: This case is the well known Shannon-entropy, given by
\begin{align}
    \ENT_1(p) = \ENT(p) := - \sum_i p_i \log p_i \ .
\end{align}
The divergence is known as Kullback-Leibler divergence, given by
\begin{align}
    \mathrm{D}_1(p \mid\mid q) = \KL{p}{q} :=  \sum_i p_i \log \frac{p_i}{q_i} \ .
\end{align}

$\alpha = 2$: This case is called the collision entropy, which is closely related to the index of coincidence. 
It is given by
\begin{align}
    \ENT_2(p) := - \log \sum_i p_i^2 \ .
\end{align}
The corresponding divergence is based upon the chi-square divergence
\begin{align}
    \mathrm{D}_2(p \mid\mid q) := \log \left(\sum_{i=1}^N \frac{p_i^2}{q_i} \right) = \log \left( 1 + \sum_{i=1}^N \frac{(p_i - q_i)^2}{q_i} \right) \ .
\end{align}

$\alpha = \infty$: The entropy is known as the min-entropy.
It is given by
\begin{align}
    \ENT_\infty(p) := - \log \max_i p_i \ .
\end{align}
The divergence 
\begin{align}
    \mathrm{D}_\infty(p \mid\mid q) := \log \sup_i \frac{p_i}{q_i} \ .
\end{align}

\textbf{Notes.}
Realizations of Renyi entropy satisfy the inequalities
\begin{align}
    \ENT_0(p) \geq \ENT_1(p) \geq \ENT_2(p) \geq \ENT_\infty(p)
\end{align}
Also Theorem 7 in \cite{vanErven:14} states that Renyi divergences are continuous in the order of $\alpha$.

\subsection{Generalization to other Proper Scoring Rules} \label{apx:sec:scoring_rules}

Another perspective on our measure of uncertainty (Eq.~\eqref{eq:cross_entropy}) is the interpretation of \cite{kotelevskii2025risk} and \cite{Hofman:24}, which consider Bayesian approximations of multiple proper scoring rules as measures of uncertainty. 
They consider the zero-one, Brier, and Spherical score in addition to the log-score, which is the cross-entropy upon which the information-theoretic measures we discussed in the main paper are based (see Eq.~\eqref{eq:cross_entropy}).
Note however, that they consider the opposite order of arguments (see Apx.~\ref{sec:apx:alternative_measure}), which leads to similar measures due to symmetry, yet has a different interpretation.
For the zero-one score, the resulting framework of measures according to our interpretation is given in Tab.~\ref{tab:framework_zero_one}, for the Brier score in Tab.~\ref{tab:framework_brier} and for the spherical score it is given in Tab.~\ref{tab:framework_spherical}.

\setlength{\tabcolsep}{7pt}
\renewcommand{\arraystretch}{1.8}
\begin{table}
\centering
\caption{\textbf{Our proposed framework applied under the zero-one score.} Each measure denotes a different instantiation of our proposed measure given by Eq.~\eqref{eq:cross_entropy}, but using the zero-one score instead of the cross-entropy (log score) for different assumptions about the predicting model and how the true model is approximated. For brevity, we define $p_{\Bw} \coloneqq p(y \mid \Bx, \Bw)$, $p_{\cD} \coloneqq p(y \mid \Bx, \cD)$, $\EXP_{\Bw} \coloneqq \EXP_{p(\Bw \mid \cD)}$ (the same for $\tilde\Bw$) and $p_{\bullet}(\widehat{p_{\circ}}) \coloneqq p(y = \argmax p(y \mid \Bx, \circ) \mid \Bx, \bullet)$. Expressions with the same cell coloring are equivalent to each other. Each measure of TU additively decomposes into AU and EU.}%
\label{tab:framework_zero_one}
{
\begin{tabular}{cl|ccc}
\hline
\multirow{2}{*}{\rotatebox{90}{}} & \hfill\multirow{2}{*}{\begin{tabular}[c]{@{}c@{}}Predicting model\end{tabular}}\hfill & \multicolumn{3}{c} {Approximation of true predictive distribution} \\ \cline{3-5}
& & \cellcolor{verylightgray} \texttt{(1)} $\; \tilde\Bw $ & \cellcolor{verylightgray} \texttt{(2)} $\; \EXP_{\tilde\Bw}$ & \cellcolor{verylightgray} \texttt{(3)} $\; \tilde\Bw \sim p(\tilde\Bw \mid \cD)$ \\
\hline
\multirow{3}{*}{\rotatebox{90}{TU}} & \cellcolor{verylightgray} \texttt{(A)} $\; \Bw$ & $1 - p_{\Bw}(\widehat{p_{\tilde\Bw}})$ & $1 - p_{\Bw}(\widehat{p_{\cD}}) $  &  $\EXP_{\tilde\Bw} \left[ 1 - p_{\Bw}(\widehat{p_{\tilde\Bw}}) \right] $ \\
& \cellcolor{verylightgray} \texttt{(B)} $\; \EXP_{\Bw}$ & \cellcolor{C1b} $1 - p_{\cD}(\widehat{p_{\tilde\Bw}})$ & \cellcolor{C1} $1 - p_{\cD}(\widehat{p_{\cD}})$  & \cellcolor{C1d} $\EXP_{\tilde\Bw} \left[ 1 - p_{\cD}(\widehat{p_{\tilde\Bw}}) \right] $  \\
& \cellcolor{verylightgray} \texttt{(C)} $\; \Bw \sim p(\Bw \mid \cD)$  & \cellcolor{C1b} $\EXP_{\Bw} \left[ 1 - p_{\Bw}(\widehat{p_{\tilde\Bw}}) \right] $ & \cellcolor{C1} $\EXP_{\Bw} \left[ 1 - p_{\Bw}(\widehat{p_{\cD}}) \right] $  & \cellcolor{C1d} $\EXP_{\Bw} \left[ \ \EXP_{\tilde\Bw} \left[ 1 - p_{\Bw}(\widehat{p_{\tilde\Bw}}) \right] \right] $ \\
\hline \hline
\multirow{3}{*}{\rotatebox{90}{AU}} & \cellcolor{verylightgray} \texttt{(A)} $\; \Bw$ & \cellcolor{C0}$1 - \max p_{\Bw}$ & \cellcolor{C0}$1 - \max p_{\Bw}$  & \cellcolor{C0}$1 - \max p_{\Bw}$ \\
&\cellcolor{verylightgray} \texttt{(B)} $\; \EXP_{\Bw}$ & \cellcolor{C1}$1 - \max p_{\cD}$ & \cellcolor{C1}$1 - \max p_{\cD}$  &  \cellcolor{C1}$1 - \max p_{\cD}$  \\
& \cellcolor{verylightgray} \texttt{(C)} $\; \Bw \sim p(\Bw \mid \cD)$ & \cellcolor{C2}$1 - \EXP_{\Bw} \left[ \max p_{\Bw} \right] $ & \cellcolor{C2}$1 - \EXP_{\Bw} \left[ \max p_{\Bw} \right] $  &  \cellcolor{C2}$1 - \EXP_{\Bw} \left[ \max p_{\Bw} \right] $ \\
\hline \hline
\multirow{3}{*}{\rotatebox{90}{EU}} & \cellcolor{verylightgray} \texttt{(A)} $\; \Bw$ & $\max p_{\Bw} - p_{\Bw}(\widehat{p_{\tilde\Bw}})$ & $\max p_{\Bw} - p_{\Bw}(\widehat{p_{\cD}}) $  &  $\EXP_{\tilde\Bw} \left[ \max p_{\Bw} - p_{\Bw}(\widehat{p_{\tilde\Bw}}) \right] $ \\
& \cellcolor{verylightgray} \texttt{(B)} $\; \EXP_{\Bw}$ & $\max p_{\cD} - p_{\cD}(\widehat{p_{\tilde\Bw}})$ & $\cancelto{0}{\max p_{\cD} - p_{\cD}(\widehat{p_{\cD}})}$  & $\EXP_{\tilde\Bw} \left[ \max p_{\cD} - p_{\cD}(\widehat{p_{\tilde\Bw}}) \right] $  \\
& \cellcolor{verylightgray} \texttt{(C)} $\; \Bw \sim p(\Bw \mid \cD)$ & $\EXP_{\Bw} \left[ \max p_{\Bw} - p_{\Bw}(\widehat{p_{\tilde\Bw}}) \right] $ & $\EXP_{\Bw} \left[ \max p_{\Bw} - p_{\Bw}(\widehat{p_{\cD}}) \right] $  & $\EXP_{\Bw} \left[ \ \EXP_{\tilde\Bw} \left[ \max p_{\Bw} - p_{\Bw}(\widehat{p_{\tilde\Bw}}) \right] \right] $ \\
\hline
\end{tabular}
}
\end{table}
\renewcommand{\arraystretch}{1}

\setlength{\tabcolsep}{4pt}
\renewcommand{\arraystretch}{2}
\begin{table}
\centering
\caption{\textbf{Our proposed framework applied under the Brier score.} Each measure denotes a different instantiation of our proposed measure given by Eq.~\eqref{eq:cross_entropy}, but using the Brier score instead of the cross-entropy (log score) for different assumptions about the predicting model and how the true model is approximated. For brevity, we define $p_{\Bw} \coloneqq p(y \mid \Bx, \Bw)$, $p_{\cD} \coloneqq p(y \mid \Bx, \cD)$, and $\EXP_{\Bw} \coloneqq \EXP_{p(\Bw \mid \cD)}$ (the same for $\tilde\Bw$). The 2-norm is defined as $\|p(y \mid \Bx, \bullet)\|_2 \coloneqq \sqrt{\sum_{k=1}^K p(y = k \mid \Bx, \bullet)^2}$. Expressions with the same cell coloring are equivalent to each other. Each measure of TU additively decomposes into AU and EU.}%
\label{tab:framework_brier}
{\smaller
\begin{tabular}{cl|ccc}
\hline
\multirow{2}{*}{\rotatebox{90}{}} & \hfill\multirow{2}{*}{\begin{tabular}[c]{@{}c@{}}{\scriptsize Predicting model}\end{tabular}}\hfill & \multicolumn{3}{c} {\scriptsize Approximation of true predictive distribution} \\ \cline{3-5}
& & \cellcolor{verylightgray} \texttt{(1)} $\; \tilde\Bw $ & \cellcolor{verylightgray} \texttt{(2)} $\; \EXP_{\tilde\Bw}$ & \cellcolor{verylightgray} \texttt{(3)} $\; \tilde\Bw \sim p(\tilde\Bw \mid \cD)$ \\
\hline
\multirow{3}{*}{\rotatebox{90}{TU}} & \cellcolor{verylightgray} \texttt{(A)} $\; \Bw$ & $1 - \|p_{\Bw}\|_2^2 + \|p_{\Bw} - p_{\tilde\Bw}\|_2^2$ & $1 - \|p_{\Bw}\|_2^2 + \|p_{\Bw} - p_{\cD}\|_2^2$  &  $\EXP_{\tilde\Bw} \left[ 1 - \|p_{\Bw}\|_2^2 + \|p_{\Bw} - p_{\tilde\Bw}\|_2^2 \right] $ \\
& \cellcolor{verylightgray} \texttt{(B)} $\; \EXP_{\Bw}$ & $1 - \|p_{\cD}\|_2^2 + \|p_{\cD} - p_{\tilde\Bw}\|_2^2$ & \cellcolor{C1} $1 - \|p_{\cD}\|_2^2 + \cancelto{0}{\|p_{\cD} - p_{\cD}\|_2^2} $  &  $\EXP_{\tilde\Bw} \left[ 1 - \|p_{\cD}\|_2^2 + \|p_{\cD} - p_{\tilde\Bw}\|_2^2 \right] $  \\
& \cellcolor{verylightgray} \texttt{(C)} $\; \Bw \sim p(\Bw \mid \cD)$ & $\EXP_{\Bw} \left[ 1 - \|p_{\Bw}\|_2^2 + \|p_{\Bw} - p_{\tilde\Bw}\|_2^2 \right] $ & $\EXP_{\Bw} \left[ 1 - \|p_{\Bw}\|_2^2 + \|p_{\Bw} - p_{\cD}\|_2^2 \right] $  &  $\EXP_{\Bw} \left[ \ \EXP_{\tilde\Bw} \left[ 1 - \|p_{\Bw}\|_2^2 + \|p_{\Bw} - p_{\tilde\Bw}\|_2^2 \right] \right] $ \\
\hline \hline
\multirow{3}{*}{\rotatebox{90}{AU}} & \cellcolor{verylightgray} \texttt{(A)} $\; \Bw$ & \cellcolor{C0}$1 - \|p_{\Bw}\|_2^2$ & \cellcolor{C0}$1 - \|p_{\Bw}\|_2^2$  & \cellcolor{C0}$1 - \|p_{\Bw}\|_2^2$ \\
&\cellcolor{verylightgray} \texttt{(B)} $\; \EXP_{\Bw}$ & \cellcolor{C1}$1 - \|p_{\cD}\|_2^2$ & \cellcolor{C1}$1 - \|p_{\cD}\|_2^2$  &  \cellcolor{C1}$1 - \|p_{\cD}\|_2^2$  \\
& \cellcolor{verylightgray} \texttt{(C)} $\; \Bw \sim p(\Bw \mid \cD)$ & \cellcolor{C2}$\EXP_{\Bw} \left[ 1 - \|p_{\Bw}\|_2^2 \right] $ & \cellcolor{C2}$\EXP_{\Bw} \left[ 1 - \|p_{\Bw}\|_2^2 \right] $  &  \cellcolor{C2}$\EXP_{\Bw} \left[ 1 - \|p_{\Bw}\|_2^2 \right] $ \\
\hline \hline
\multirow{3}{*}{\rotatebox{90}{EU}} & \cellcolor{verylightgray} \texttt{(A)} $\; \Bw$ & $\|p_{\Bw} - p_{\tilde\Bw}\|_2^2$ & $\|p_{\Bw} - p_{\cD}\|_2^2$  &  $\EXP_{\tilde\Bw} \left[ \|p_{\Bw} - p_{\tilde\Bw}\|_2^2 \right] $ \\
& \cellcolor{verylightgray} \texttt{(B)} $\; \EXP_{\Bw}$ & $\|p_{\cD} - p_{\tilde\Bw}\|_2^2$ & $\cancelto{0}{\|p_{\cD} - p_{\cD}\|_2^2}$  & $\EXP_{\tilde\Bw} \left[ \|p_{\cD} - p_{\tilde\Bw}\|_2^2 \right] $  \\
& \cellcolor{verylightgray} \texttt{(C)} $\; \Bw \sim p(\Bw \mid \cD)$ & $\EXP_{\Bw} \left[ \|p_{\Bw} - p_{\tilde\Bw}\|_2^2 \right] $ & $\EXP_{\Bw} \left[ \|p_{\Bw} - p_{\cD}\|_2^2 \right] $  & $\EXP_{\Bw} \left[ \ \EXP_{\tilde\Bw} \left[ \|p_{\Bw} - p_{\tilde\Bw}\|_2^2 \right] \right] $ \\
\hline
\end{tabular}
}
\end{table}
\renewcommand{\arraystretch}{1}

\setlength{\tabcolsep}{7pt}
\renewcommand{\arraystretch}{2.2}
\begin{table}
\centering
\caption{\textbf{Our proposed framework applied under the spherical score.} Each measure denotes a different instantiation of our proposed measure given by Eq.~\eqref{eq:cross_entropy}, but using the spherical score instead of the cross-entropy (log score) for different assumptions about the predicting model and how the true model is approximated. For brevity, we define $p_{\Bw} \coloneqq p(y \mid \Bx, \Bw)$, $p_{\cD} \coloneqq p(y \mid \Bx, \cD)$, and $\EXP_{\Bw} \coloneqq \EXP_{p(\Bw \mid \cD)}$ (the same for $\tilde\Bw$). The 2-norm is defined as $\|p_\bullet\|_2 \coloneqq \sqrt{\sum_{k=1}^K p(y = k \mid \Bx, \bullet)^2}$. Furthermore, the scalar product is defined as $\langle p_\bullet, p_\circ \rangle \coloneqq \sum_{k=1}^K p(y = k \mid \Bx, \bullet) \cdot p(y = k \mid \Bx, \circ)$. Expressions with the same cell coloring are equivalent to each other. Each measure of TU additively decomposes into AU and EU. \\}%
\label{tab:framework_spherical}
{
\begin{tabular}{cl|ccc}
\hline
\multirow{2}{*}{\rotatebox{90}{}} & \hfill\multirow{2}{*}{\begin{tabular}[c]{@{}c@{}}{Predicting model}\end{tabular}}\hfill & \multicolumn{3}{c} {Approximation of true predictive distribution} \\ \cline{3-5}
& & \cellcolor{verylightgray} \texttt{(1)} $\; \tilde\Bw $ & \cellcolor{verylightgray} \texttt{(2)} $\; \EXP_{\tilde\Bw}$ & \cellcolor{verylightgray} \texttt{(3)} $\; \tilde\Bw \sim p(\tilde\Bw \mid \cD)$ \\
\hline
\multirow{3}{*}{\rotatebox{90}{TU}} & \cellcolor{verylightgray} \texttt{(A)} $\; \Bw$ & $1 - \frac{\langle p_{\Bw}, p_{\tilde\Bw} \rangle}{\|p_{\tilde\Bw}\|_2}$ & $1 - \frac{\langle p_{\Bw}, p_{\cD} \rangle}{\|p_{\cD}\|_2}$  &  $\EXP_{\tilde\Bw} \left[ 1 - \frac{\langle p_{\Bw}, p_{\tilde\Bw} \rangle}{\|p_{\tilde\Bw}\|_2} \right] $ \\
& \cellcolor{verylightgray} \texttt{(B)} $\; \EXP_{\Bw}$ & \cellcolor{C1b} $1 - \frac{\langle p_{\cD}, p_{\tilde\Bw} \rangle}{\|p_{\tilde\Bw}\|_2}$ & \cellcolor{C1} $1 - \frac{\langle p_{\cD}, p_{\cD} \rangle}{\|p_{\cD}\|_2}$ &  \cellcolor{C1d} $\EXP_{\tilde\Bw} \left[ 1 - \frac{\langle p_{\cD}, p_{\tilde\Bw} \rangle}{\|p_{\tilde\Bw}\|_2} \right] $  \\
& \cellcolor{verylightgray} \texttt{(C)} $\; \Bw \sim p(\Bw \mid \cD)$ & \cellcolor{C1b} $\EXP_{\Bw} \left[ 1 - \frac{\langle p_{\Bw}, p_{\tilde\Bw} \rangle}{\|p_{\tilde\Bw}\|_2} \right] $ & \cellcolor{C1} $\EXP_{\Bw} \left[ 1 - \frac{\langle p_{\Bw}, p_{\cD} \rangle}{\|p_{\cD}\|_2} \right] $  & \cellcolor{C1d} $\EXP_{\Bw} \left[ \ \EXP_{\tilde\Bw} \left[ 1 - \frac{\langle p_{\Bw}, p_{\tilde\Bw} \rangle}{\|p_{\tilde\Bw}\|_2} \right] \right] $ \\
\hline \hline
\multirow{3}{*}{\rotatebox{90}{AU}} & \cellcolor{verylightgray} \texttt{(A)} $\; \Bw$ & \cellcolor{C0}$1 - \|p_{\Bw}\|_2$ & \cellcolor{C0}$1 - \|p_{\Bw}\|_2$  & \cellcolor{C0}$1 - \|p_{\Bw}\|_2$ \\
&\cellcolor{verylightgray} \texttt{(B)} $\; \EXP_{\Bw}$ & \cellcolor{C1}$1 - \|p_{\cD}\|_2$ & \cellcolor{C1}$1 - \|p_{\cD}\|_2$  &  \cellcolor{C1}$1 - \|p_{\cD}\|_2$  \\
& \cellcolor{verylightgray} \texttt{(C)} $\; \Bw \sim p(\Bw \mid \cD)$ & \cellcolor{C2}$\EXP_{\Bw} \left[ 1 - \|p_{\Bw}\|_2 \right] $ & \cellcolor{C2}$\EXP_{\Bw} \left[ 1 - \|p_{\Bw}\|_2 \right] $  &  \cellcolor{C2}$\EXP_{\Bw} \left[ 1 - \|p_{\Bw}\|_2 \right] $ \\
\hline \hline
\multirow{3}{*}{\rotatebox{90}{EU}} & \cellcolor{verylightgray} \texttt{(A)} $\; \Bw$ & $\|p_{\Bw}\|_2 - \frac{\langle p_{\Bw}, p_{\tilde\Bw} \rangle}{\|p_{\tilde\Bw}\|_2}$ & $\|p_{\Bw}\|_2 - \frac{\langle p_{\Bw}, p_{\cD} \rangle}{\|p_{\cD}\|_2}$  &  $\EXP_{\tilde\Bw} \left[ \|p_{\Bw}\|_2 - \frac{\langle p_{\Bw}, p_{\tilde\Bw} \rangle}{\|p_{\tilde\Bw}\|_2} \right] $ \\
& \cellcolor{verylightgray} \texttt{(B)} $\; \EXP_{\Bw}$ & $\|p_{\cD}\|_2 - \frac{\langle p_{\cD}, p_{\tilde\Bw} \rangle}{\|p_{\tilde\Bw}\|_2}$ & $\cancelto{0}{\|p_{\cD}\|_2 - \frac{\langle p_{\cD}, p_{\cD} \rangle}{\|p_{\cD}\|_2}}$  & $\EXP_{\tilde\Bw} \left[ \|p_{\cD}\|_2 - \frac{\langle p_{\cD}, p_{\tilde\Bw} \rangle}{\|p_{\tilde\Bw}\|_2} \right] $  \\
& \cellcolor{verylightgray} \texttt{(C)} $\; \Bw \sim p(\Bw \mid \cD)$ & $\EXP_{\Bw} \left[ \|p_{\Bw}\|_2 - \frac{\langle p_{\Bw}, p_{\tilde\Bw} \rangle}{\|p_{\tilde\Bw}\|_2} \right] $ & $\EXP_{\Bw} \left[ \|p_{\Bw}\|_2 - \frac{\langle p_{\Bw}, p_{\cD} \rangle}{\|p_{\cD}\|_2} \right] $  & $\EXP_{\Bw} \left[ \ \EXP_{\tilde\Bw} \left[ \|p_{\Bw}\|_2 - \frac{\langle p_{\Bw}, p_{\tilde\Bw} \rangle}{\|p_{\tilde\Bw}\|_2} \right] \right] $ \\
\hline
\end{tabular}
}
\end{table}
\renewcommand{\arraystretch}{1}

\subsection{Alternative Measure} \label{sec:apx:alternative_measure}
The reverse order of the arguments for the cross-entropy in Eq.~\eqref{eq:cross_entropy}, that is, $\CE{p(y \mid \Bx, \Bw^*)}{p(y \mid \Bx, \cdot)}$, gives rise to an alternative measure that is consistent with Eq.~\eqref{eq:entropy}.
This measure, also known as ``pointwise risk'' under the log score at an input (point) $\Bx$, has been considered as a measure of predictive uncertainty \citep{Gruber:23, Lahlou:23, kotelevskii2025risk, Hofman:24}.
\citet{kotelevskii2025risk} introduced a framework based on similar ideas for Bayesian approximation as we consider in our framework (i.e. settings \texttt{(B)}, \texttt{(C)}, \texttt{(2)} and \texttt{(3)}) and is the work closest to our work.
They introduce a framework based on decompositions of proper scoring rules (pointwise risk measures), where the logarithmic proper scoring rule is a special case, and subsequent Bayesian approximation.
However, they do not cover settings \texttt{(A)} and \texttt{(1)} in our framework, introduce their framework based on the notion of pointwise risk and focus on different empirical aspects of their resulting measures.

We argue that our proposed measure (Eq.~\eqref{eq:cross_entropy}) has a more meaningful interpretation than $\CE{p(y \mid \Bx, \Bw^*)}{p(y \mid \Bx, \cdot)}$.
Our measure considers the uncertainty inherent to predicting with the selected model, plus the uncertainty due to any potential mismatch with the true model.
The alternative measure considers the uncertainty inherent to predicting with the true model, plus the uncertainty due to any potential mismatch with the selected model.
However, we generally don't know the true model, thus can't actually use it to predict and have to resort to an approximation of the true model anyways.

\subsection{Regression} \label{apx:sec:regression}

For a probabilistic regression model, e.g. under a Gaussian assumption, the distribution parameters are estimated, i.e. mean and variance for the Gaussian predictive distribution.
The model is then trained by minimizing the negative log-likelihood under the training dataset.

Many works follow \cite{Depeweg:18} and utilize a variance decomposition for uncertainty quantification, where the AU is the expected variance and the EU is the variance of means, where expectation and variance are over the model posterior.
However, \cite{Depeweg:18} also consider the uncertainty measure given by Eq.~\eqref{eq:original_measure}, using differential entropies for the continuous predictive distributions.
The same can be done in order to adapt our framework in Tab.~\ref{tab:framework_main} for continuous predictive distributions.

Nevertheless, there are two important drawbacks one need to consider when doing this.
First, differential entropy can be unbounded, depending on the nature of the predictive distribution.
For the example of a Gaussian, it can be between $-\infty$ and $\infty$.
In addition, it is not invariant to a change of variables, making it a relative rather than an absolute measure.
Second, the posterior predictive distribution as defined in Eq.~\eqref{eq:posterior_predictive} is generally a mixture of individual distributions, unlike in the discrete case.
This makes MC approximations of the resulting measures more involved.

\clearpage
\section{Experimental Details and Additional Results}

In this section, we provide additional empirical results of our evaluation of the proposed framework of uncertainty measures. \\
The code to reproduce our experiments %
is available at \url{https:\\github.com/ml-jku/uncertainty-measures}.

\subsection{Illustrative Example}

Here, we provide an illustrative synthetic example often discussed in the literature \citep{Wimmer:23, Schweighofer:23b, Sale:23b}.
We consider a predictor defined as a Bernoulli distribution leading to the predictive distribution $p(y \mid \theta)$.
Thus, there is no model involved for mapping from the input space to the Bernoulli parameter.
The only free parameter is the Bernoulli parameter.
Therefore, the posterior distribution is defined as $p(\theta \mid \cD) = p(\cD \mid \theta) p(\theta) / p(\cD)$.
To examplify our framework, we consider a Beta posterior distribution $Beta(\theta; 2, 3)$.
The true Bernoulli parameter $\theta^*$ is not known.

\begin{figure}[h!]
    \centering
    \includegraphics[width=\textwidth, trim=3mm 3mm 3mm 3mm, clip]{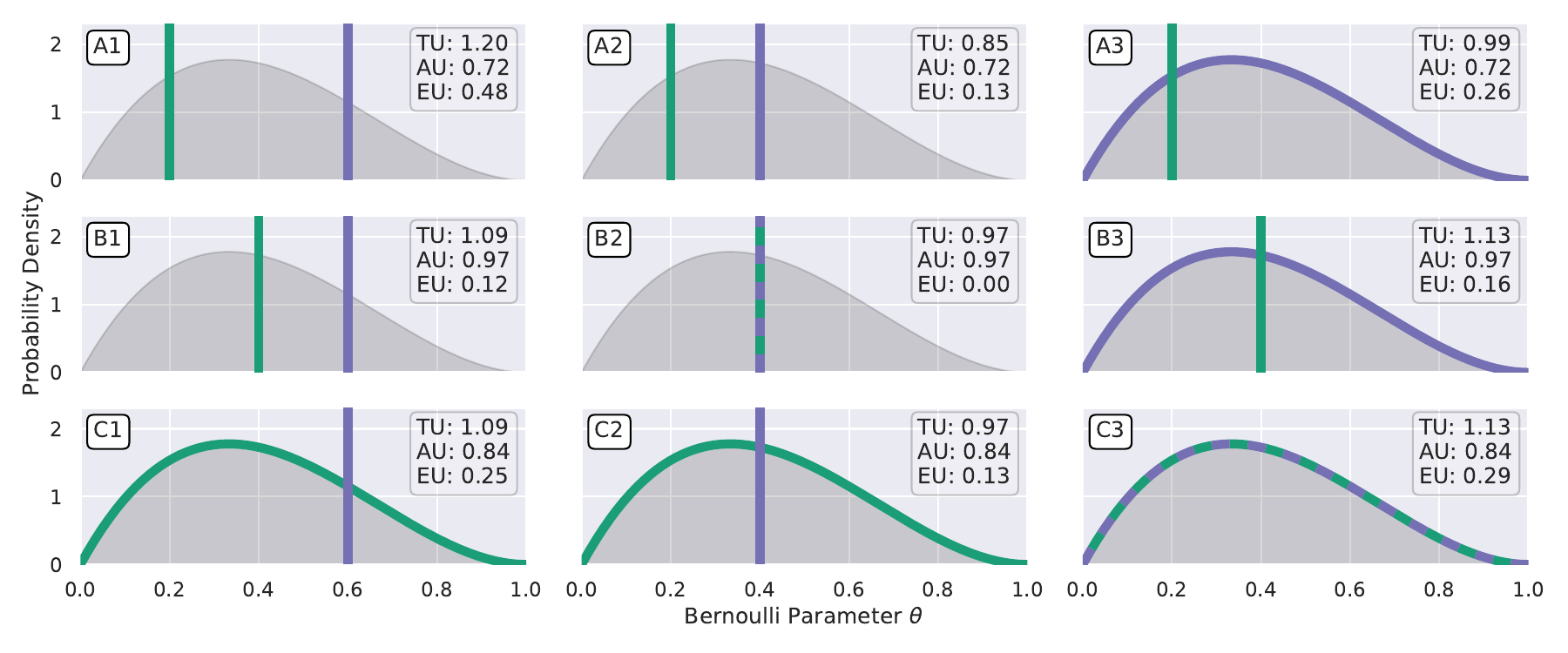}
    \caption{Uncertainty measures given by \textcolor[HTML]{1B9E77}{the predicting model} and \textcolor[HTML]{7570B3}{the approximation of the true model}. We consider the posterior distribution $Beta(\theta; 2, 3)$, shaded in gray.}
    \label{fig:illustrative example}
\end{figure}

Results are shown in Fig.~\ref{fig:illustrative example}, depicting what is considered as predicting model (green) and what is compared to as approximation of the true model.
The green line for measures \texttt{(A1/2/3)} and the violet line for measures \texttt{(A/B/C1)} were chosen arbitrarily, but different to the expected Bernoulli parameter to exemplify the differences between measures.

\subsection{Detailed Results on Main Experiments} \label{apx:sec:detailed_results}

The results for selective prediction in the main paper only provide the AUARC, but not individual accuracy-rejection curves, which we provide in this section.
Furthermore, results for misclassification detection and OOD detection in the main paper show aggregate performances over multiple datasets to provide more robust conclusions about the performance of individual measures of uncertainty.
In this section, we provide individual results for completeness.

\begin{figure}
    \centering
    \begin{subfigure}[b]{\textwidth}
        \includegraphics[width=\textwidth, trim = 0cm 1cm 0cm 1cm, clip]{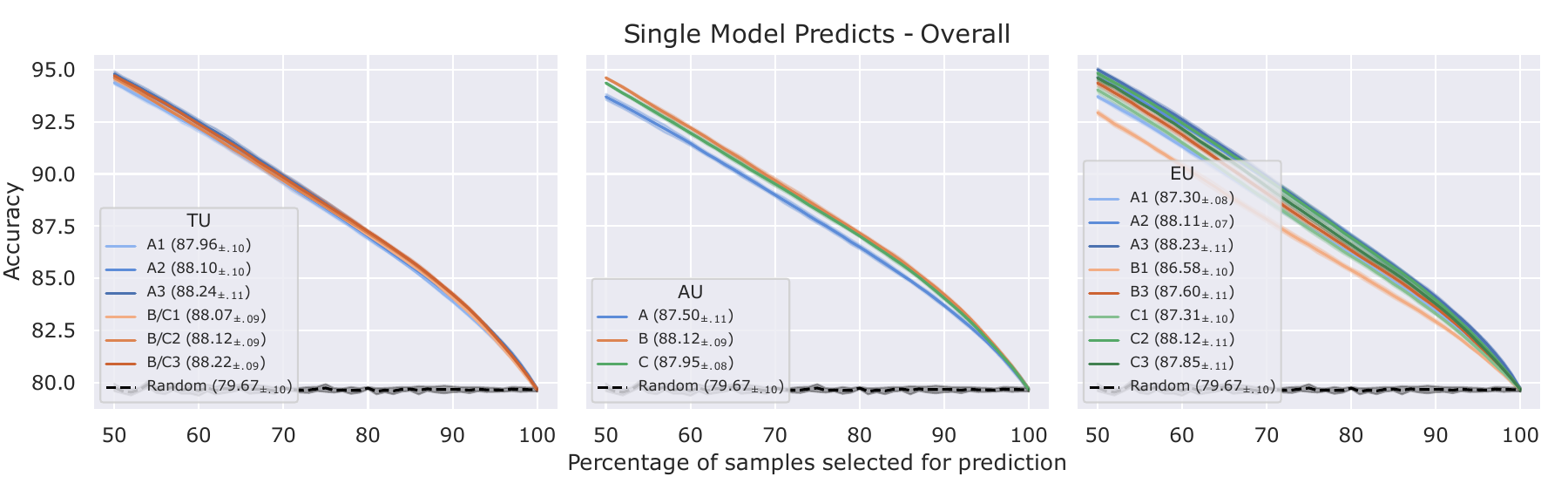}
        \subcaption{Prediction: single model}
    \end{subfigure}
    \begin{subfigure}[b]{\textwidth}
        \includegraphics[width=\textwidth, trim = 0cm 1cm 0cm 1cm, clip]{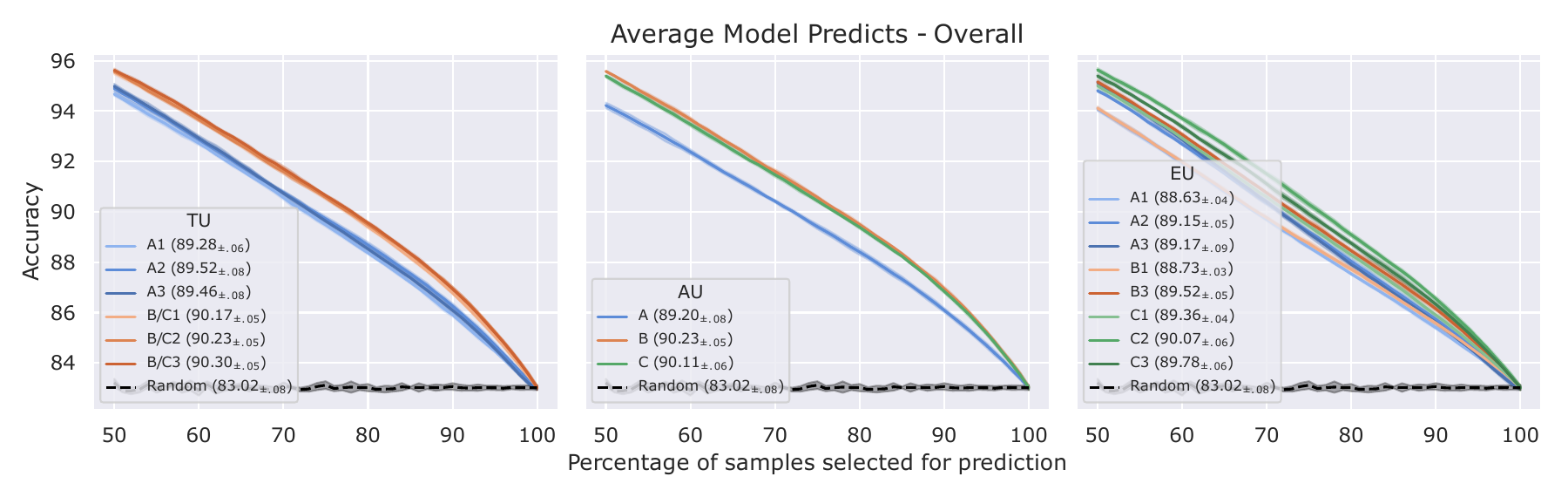}
        \subcaption{Prediction: average model}
    \end{subfigure}
    \begin{subfigure}[b]{\textwidth}
        \includegraphics[width=\textwidth, trim = 0cm 1cm 0cm 1cm, clip]{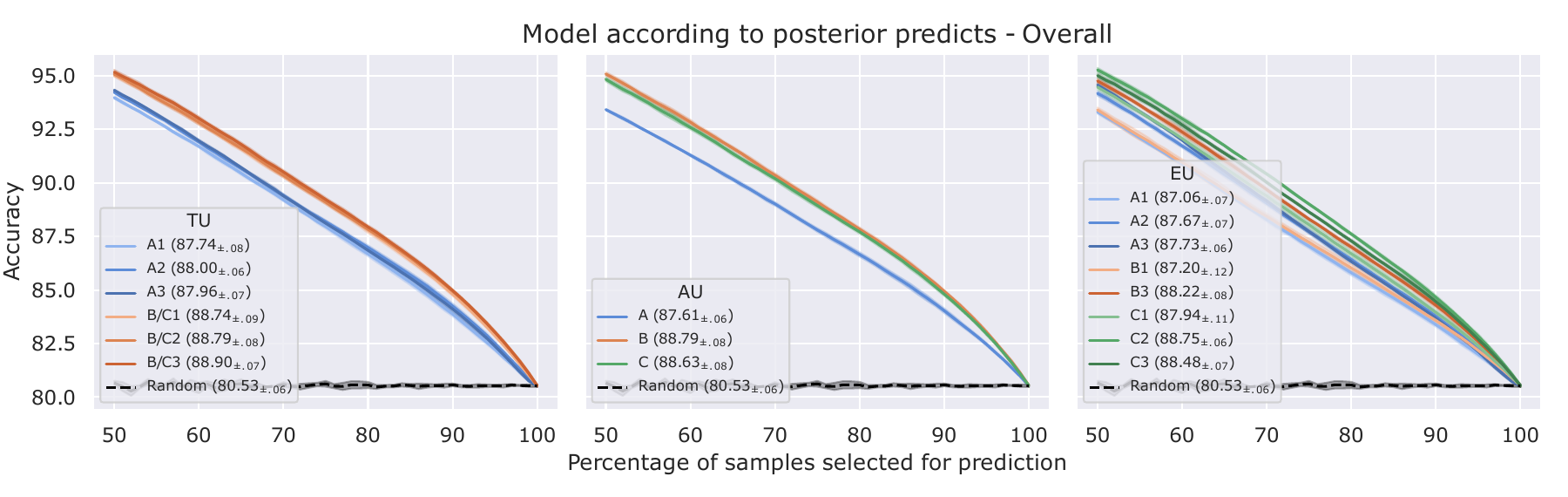}
        \subcaption{Prediction: according to posterior}
    \end{subfigure}
    \caption{\textbf{Selective prediction results.} Accuracies per fraction of datapoints the respecitive predictive model model predicts on, as well as AUARC (tabulated in legend) using different proposed measures of uncertainty as score. Accuracies are averaged over all datasets. Means and standard deviations are calculated using five runs.}
    \label{fig:selpred}
\end{figure}

\paragraph{Selective prediction.} 
We provide detailed additional results for selective prediction as discussed in the main paper.
The accuracy rejection curves and their AUARCs are shown in Fig.~\ref{fig:selpred}.
For predicting using a single model, the best measure is TU \texttt{(A3)}.
Overall, measures that consider the single model as predicting model perform well overall for measures of TU, AU and EU.
For predicting using the average model, TU \texttt{(B/C3)} performs best. 
Finally, the results for predicting under a model according to the posterior are very similar to the results under the average model.

\begin{figure}[h]
    \centering
    \captionsetup[subfigure]{labelformat=empty}
    \captionsetup[subfigure]{aboveskip=2pt, belowskip=3pt}
    \captionsetup{aboveskip=2pt, belowskip=3pt}
    \begin{subfigure}[b]{0.49\textwidth}
        \includegraphics[width=\textwidth, trim = 0.3cm 0.3cm 0.3cm 1cm, clip]{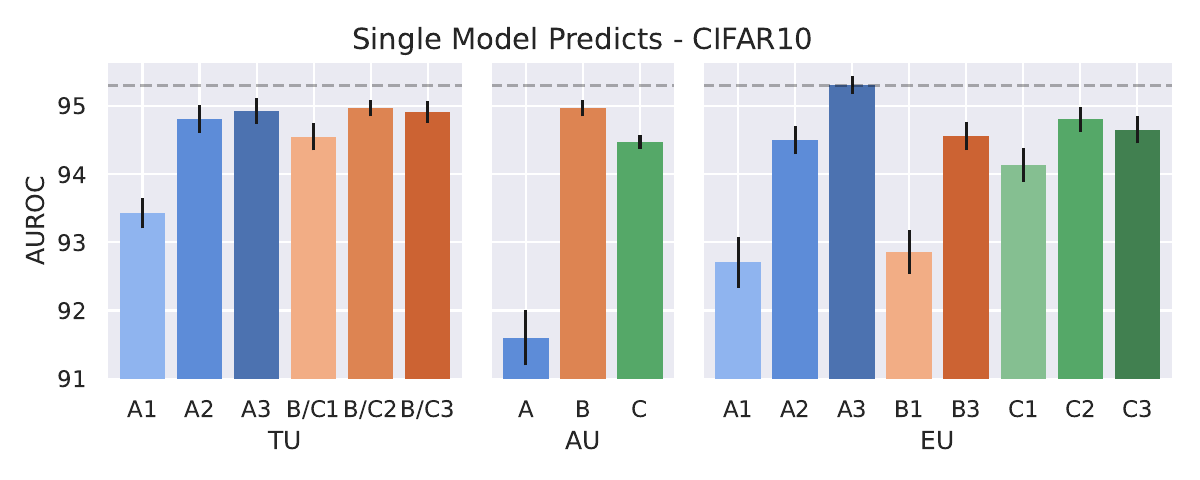}
        \subcaption{CIFAR10}
    \end{subfigure}
    \hfill
    \begin{subfigure}[b]{0.49\textwidth}
        \includegraphics[width=\textwidth, trim = 0.3cm 0.3cm 0.3cm 1cm, clip]{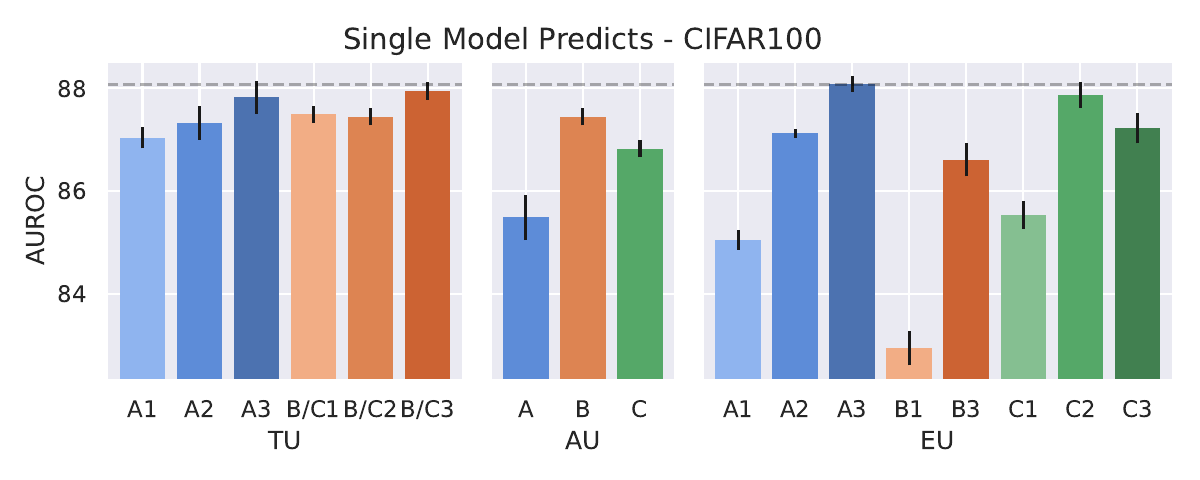}
        \subcaption{CIFAR100}
    \end{subfigure}
    \hfill
    \begin{subfigure}[b]{0.49\textwidth}
        \includegraphics[width=\textwidth, trim = 0.3cm 0.3cm 0.3cm 1cm, clip]{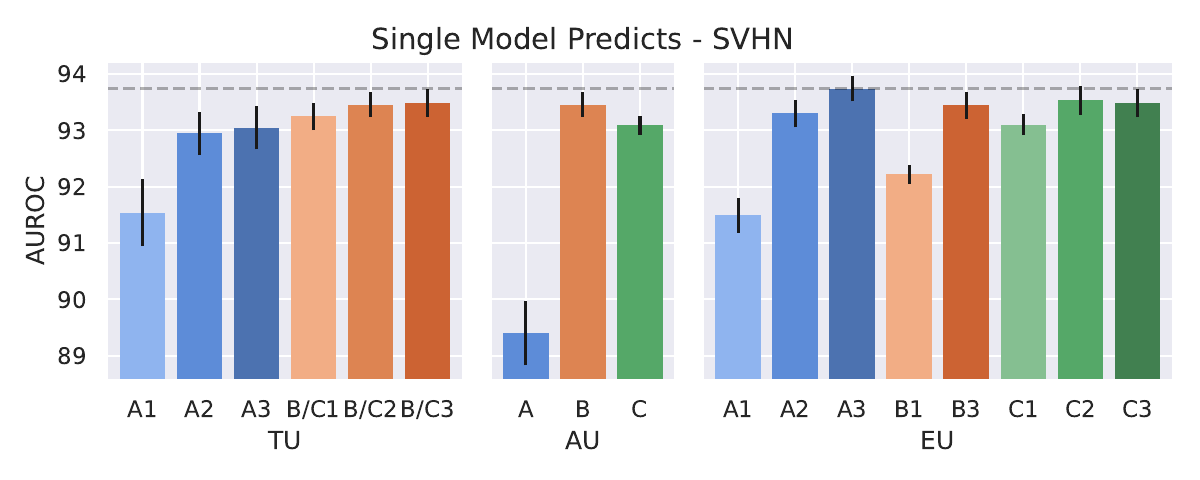}
        \subcaption{SVHN}
    \end{subfigure}
    \hfill
    \begin{subfigure}[b]{0.49\textwidth}
        \includegraphics[width=\textwidth, trim = 0.3cm 0.3cm 0.3cm 1cm, clip]{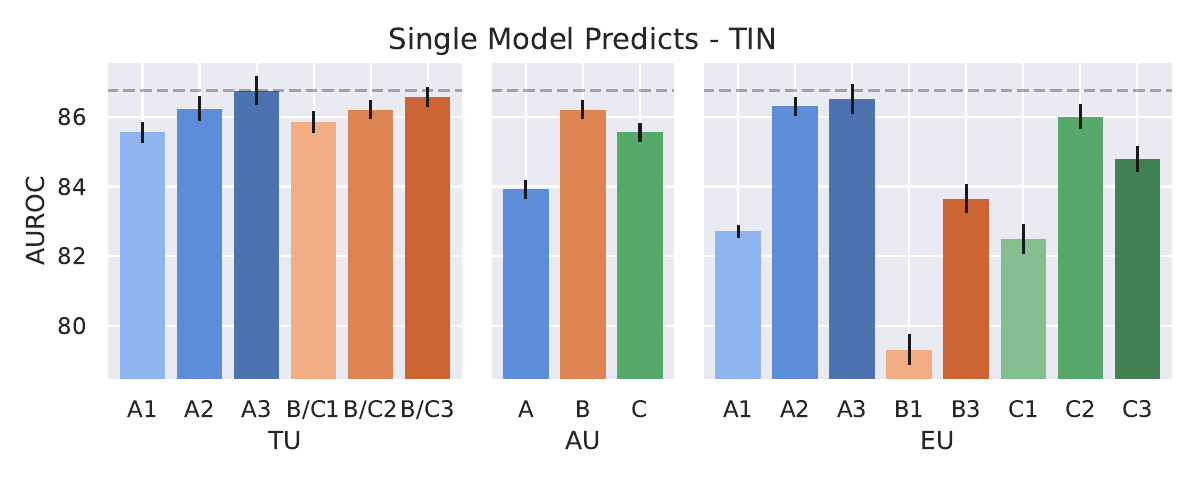}
        \subcaption{TIN}
    \end{subfigure}
    \caption{\textbf{Misclassification detection results under (i) single predicting model.} AUROC for distinguishing between correctly and incorrectly predicted datapoints under a single predicting model, using the different proposed measures of uncertainty as score. Means and standard deviations are calculated using five runs.}
    \label{fig:misc:det:single}
\end{figure}

\begin{figure}
    \centering
    \captionsetup[subfigure]{labelformat=empty}
    \captionsetup[subfigure]{aboveskip=2pt, belowskip=3pt}
    \captionsetup{aboveskip=2pt, belowskip=3pt}
    \begin{subfigure}[b]{0.49\textwidth}
        \includegraphics[width=\textwidth, trim = 0.3cm 0.3cm 0.3cm 1cm, clip]{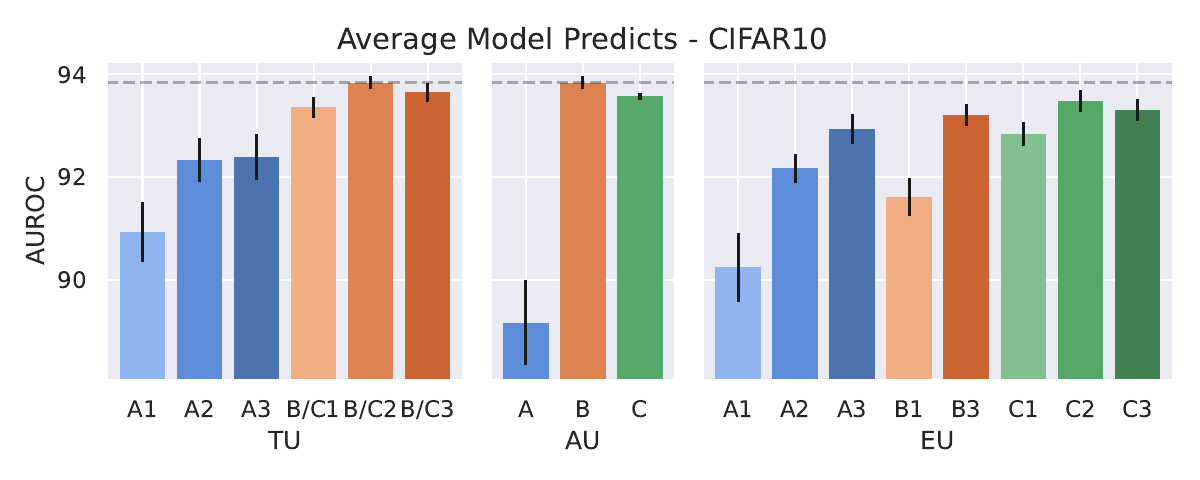}
        \subcaption{CIFAR10}
    \end{subfigure}
    \hfill
    \begin{subfigure}[b]{0.49\textwidth}
        \includegraphics[width=\textwidth, trim = 0.3cm 0.3cm 0.3cm 1cm, clip]{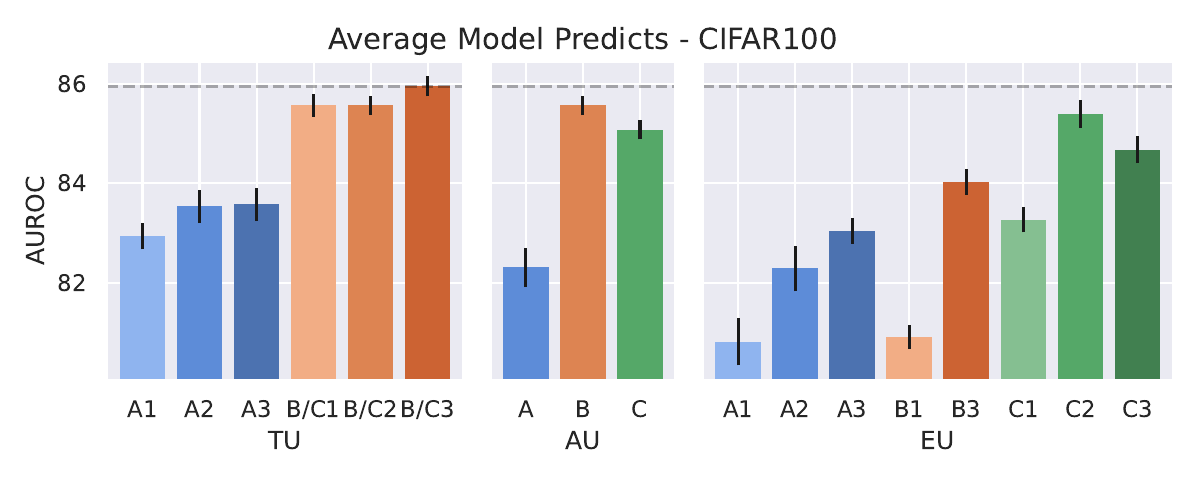}
        \subcaption{CIFAR100}
    \end{subfigure}
    \hfill
    \begin{subfigure}[b]{0.49\textwidth}
        \includegraphics[width=\textwidth, trim = 0.3cm 0.3cm 0.3cm 1cm, clip]{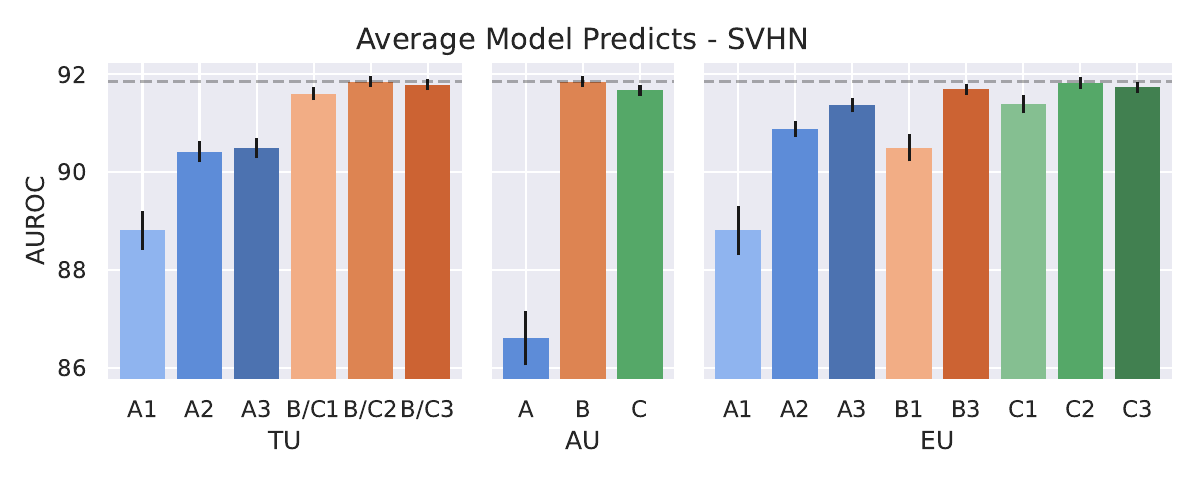}
        \subcaption{SVHN}
    \end{subfigure}
    \hfill
    \begin{subfigure}[b]{0.49\textwidth}
        \includegraphics[width=\textwidth, trim = 0.3cm 0.3cm 0.3cm 1cm, clip]{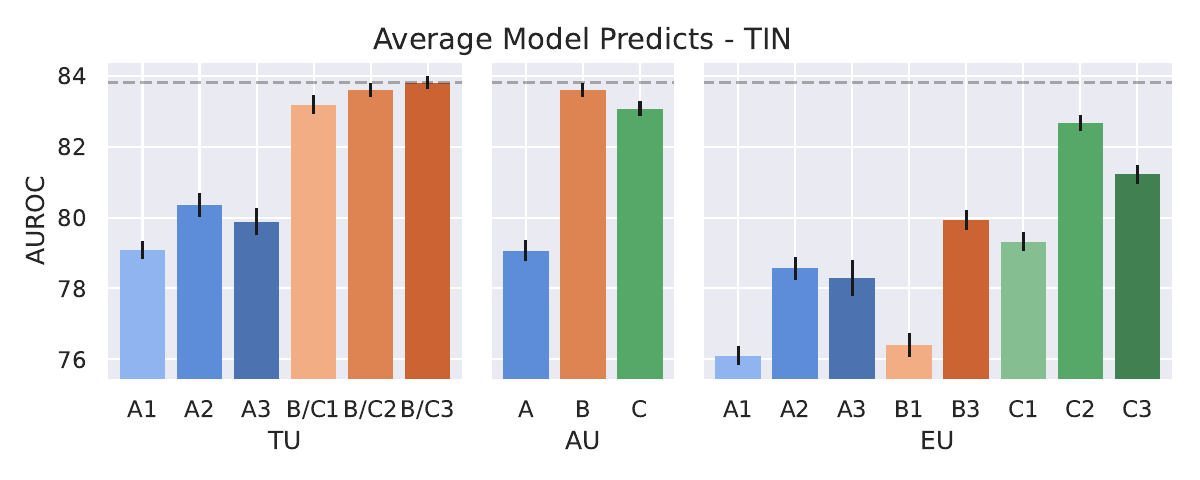}
        \subcaption{TIN}
    \end{subfigure}
    \caption{\textbf{Misclassification detection results under (ii) average predicting model.} AUROC for distinguishing between correctly and incorrectly predicted datapoints under the average predicting model, using the different proposed measures of uncertainty as score. Means and standard deviations are calculated using five runs.}
    \label{fig:misc:det:average}
\end{figure}

\begin{figure}
    \centering
    \captionsetup[subfigure]{labelformat=empty}
    \captionsetup[subfigure]{aboveskip=2pt, belowskip=3pt}
    \captionsetup{aboveskip=2pt, belowskip=3pt}
    \begin{subfigure}[b]{0.49\textwidth}
        \includegraphics[width=\textwidth, trim = 0.3cm 0.3cm 0.3cm 1cm, clip]{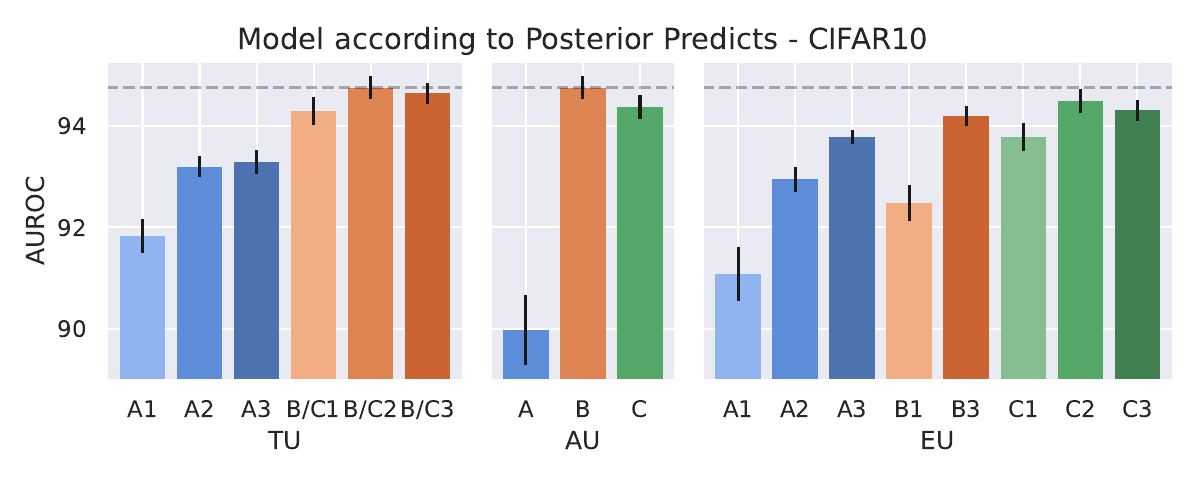}
        \subcaption{CIFAR10}
    \end{subfigure}
    \hfill
    \begin{subfigure}[b]{0.49\textwidth}
        \includegraphics[width=\textwidth, trim = 0.3cm 0.3cm 0.3cm 1cm, clip]{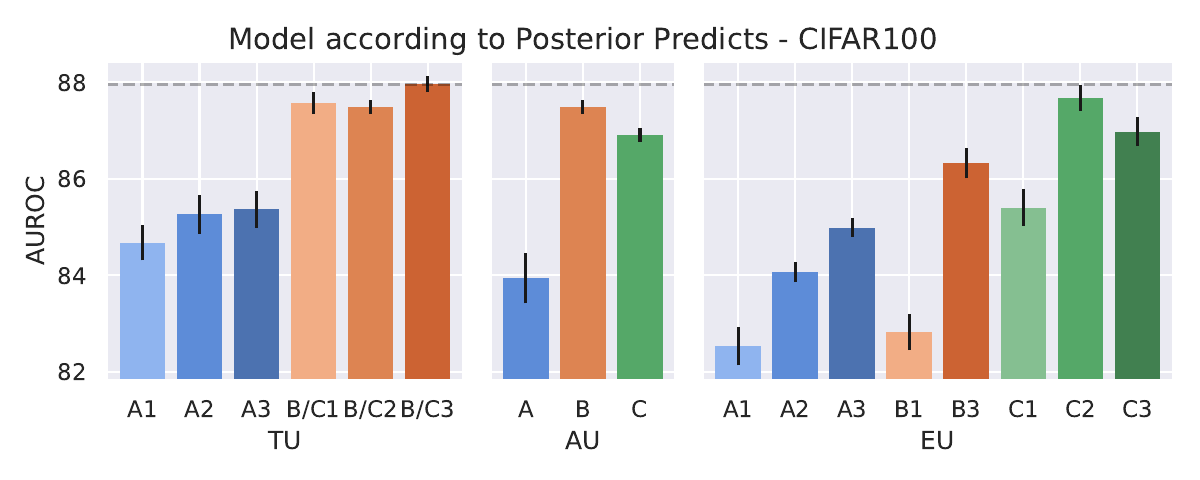}
        \subcaption{CIFAR100}
    \end{subfigure}
    \hfill
    \begin{subfigure}[b]{0.49\textwidth}
        \includegraphics[width=\textwidth, trim = 0.3cm 0.3cm 0.3cm 1cm, clip]{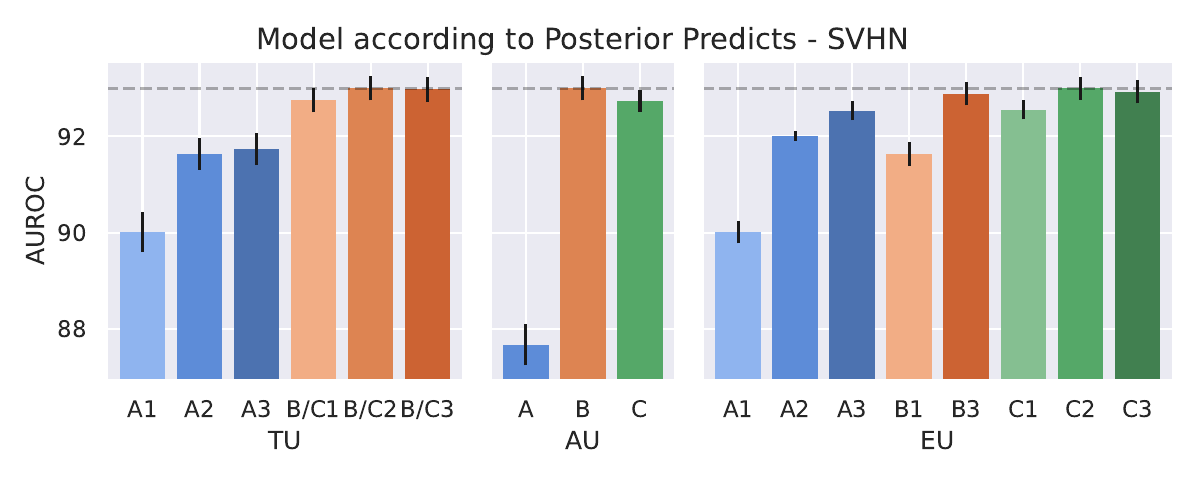}
        \subcaption{SVHN}
    \end{subfigure}
    \hfill
    \begin{subfigure}[b]{0.49\textwidth}
        \includegraphics[width=\textwidth, trim = 0.3cm 0.3cm 0.3cm 1cm, clip]{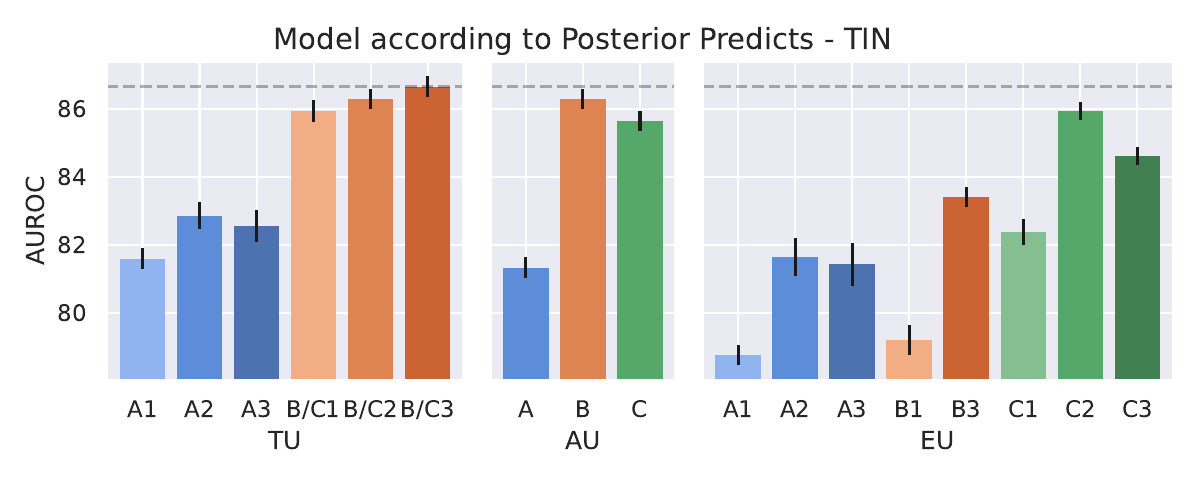}
        \subcaption{TIN}
    \end{subfigure}
    \caption{\textbf{Misclassification detection results under (iii) a model according to the posterior.} AUROC for distinguishing between correctly and incorrectly predicted datapoints under a model according to the posterior, using the different proposed measures of uncertainty as score. Means and standard deviations are calculated using five runs.}
    \label{fig:misc:det:posterior}
\end{figure}

\paragraph{Misclassification detection.}
The detailed results for misclassification detection are given in Fig.~\ref{fig:misc:det:single} for a single predicting model, in Fig.~\ref{fig:misc:det:average} for the average predicting model as well as in Fig.~\ref{fig:misc:det:posterior} for predicting with a model according to the posterior.
Although there are nuanced differences between datasets, conclusions translate very well between them.

\begin{figure}
    \centering
    \captionsetup[subfigure]{labelformat=empty}
    \captionsetup[subfigure]{aboveskip=2pt, belowskip=3pt}
    \captionsetup{aboveskip=2pt, belowskip=3pt}
    \begin{subfigure}[b]{0.49\textwidth}
        \includegraphics[width=\textwidth, trim = 0.3cm 0.3cm 0.3cm 1cm, clip]{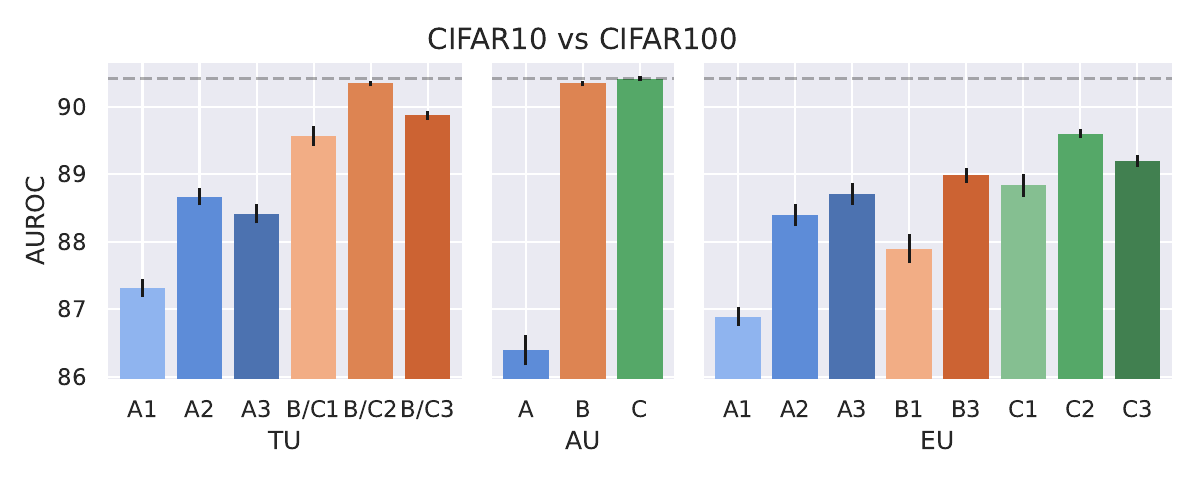}
        \subcaption{CIFAR10 / CIFAR100}
    \end{subfigure}
    \hfill
    \begin{subfigure}[b]{0.49\textwidth}
        \includegraphics[width=\textwidth, trim = 0.3cm 0.3cm 0.3cm 1cm, clip]{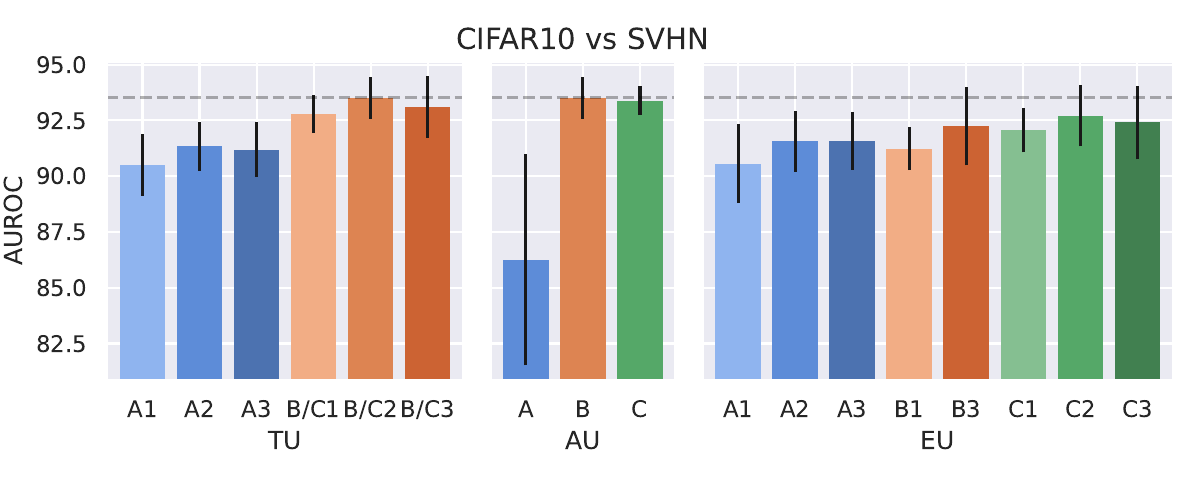}
        \subcaption{CIFAR10 / SVHN}
    \end{subfigure}
    \hfill
    \begin{subfigure}[b]{0.49\textwidth}
        \includegraphics[width=\textwidth, trim = 0.3cm 0.3cm 0.3cm 1cm, clip]{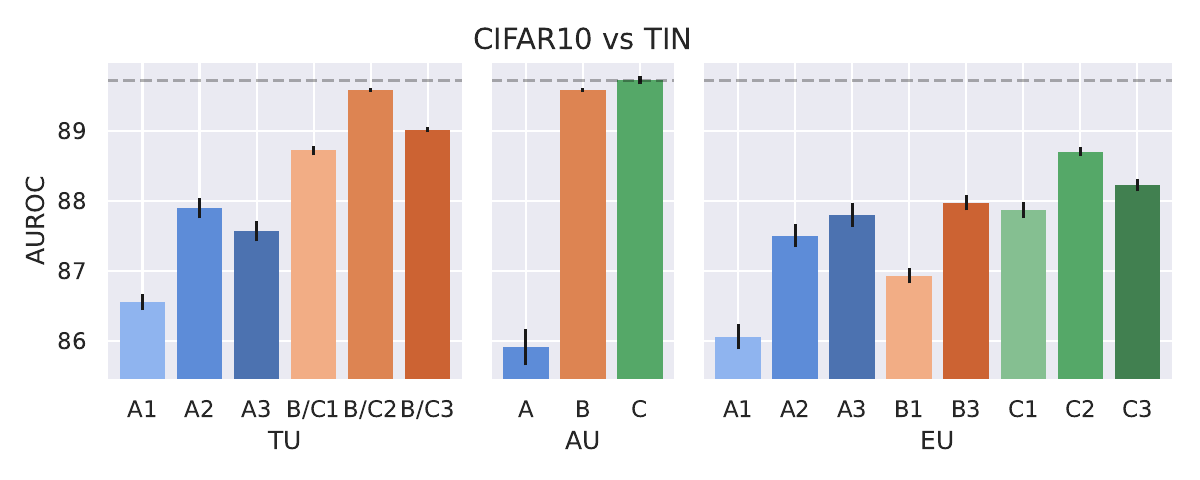}
        \subcaption{CIFAR10 / TIN}
    \end{subfigure}
    \hfill
    \begin{subfigure}[b]{0.49\textwidth}
        \includegraphics[width=\textwidth, trim = 0.3cm 0.3cm 0.3cm 1cm, clip]{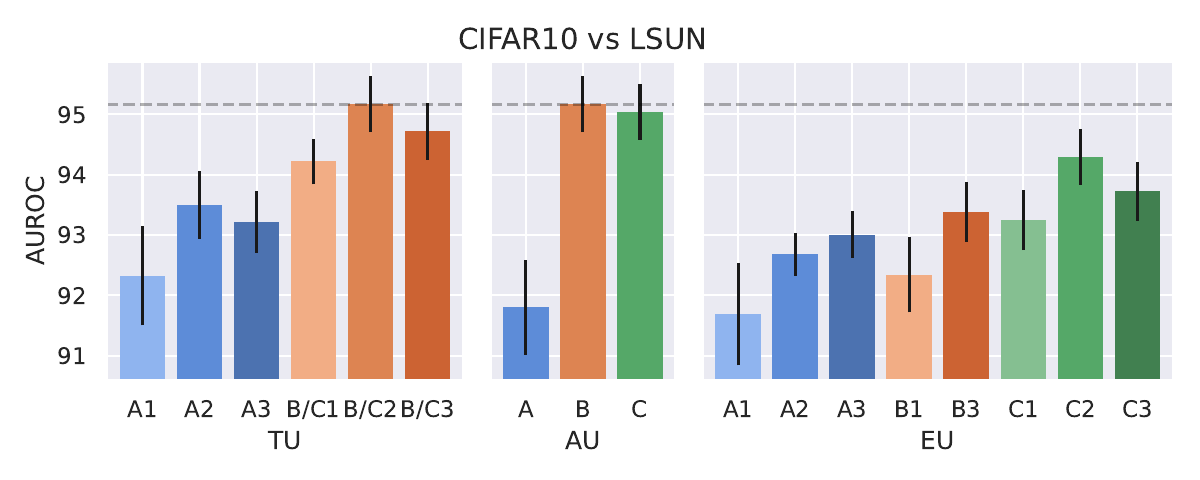}
        \subcaption{CIFAR10 / LSUN}
    \end{subfigure}
    \caption{\textbf{OOD detection results for CIFAR10.} AUROC for distinguishing between ID and OOD datapoints using the different proposed measures of uncertainty as score. Means and standard deviations are calculated using five runs.}
    \label{fig:ood:det:cifar10}
\end{figure}

\begin{figure}
    \centering
    \captionsetup[subfigure]{labelformat=empty}
    \captionsetup[subfigure]{aboveskip=2pt, belowskip=3pt}
    \captionsetup{aboveskip=2pt, belowskip=3pt}
    \begin{subfigure}[b]{0.49\textwidth}
        \includegraphics[width=\textwidth, trim = 0.3cm 0.3cm 0.3cm 1cm, clip]{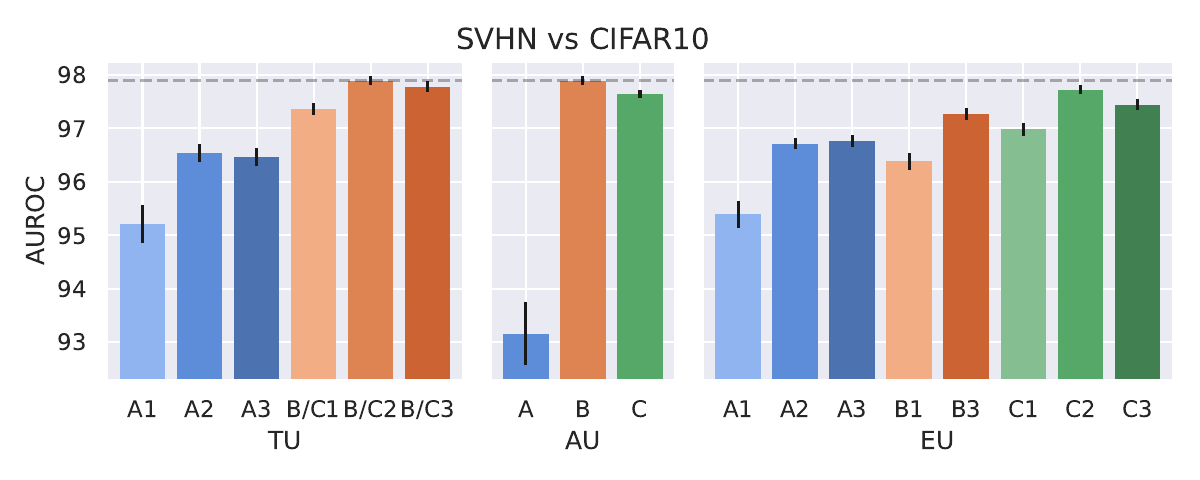}
        \subcaption{SVHN / CIFAR10}
    \end{subfigure}
    \hfill
    \begin{subfigure}[b]{0.49\textwidth}
        \includegraphics[width=\textwidth, trim = 0.3cm 0.3cm 0.3cm 1cm, clip]{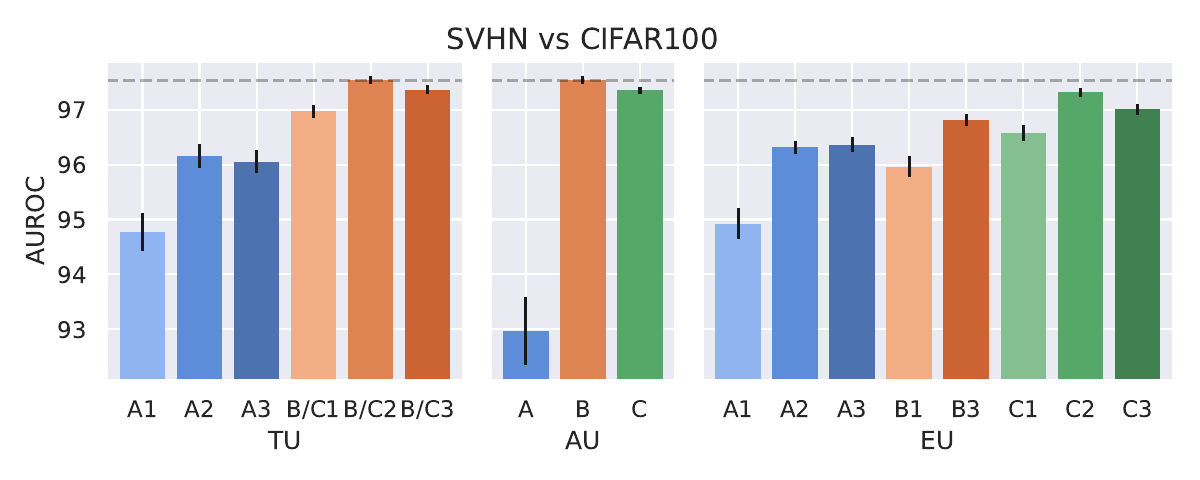}
        \subcaption{SVHN / CIFAR100}
    \end{subfigure}
    \hfill
    \begin{subfigure}[b]{0.49\textwidth}
        \includegraphics[width=\textwidth, trim = 0.3cm 0.3cm 0.3cm 1cm, clip]{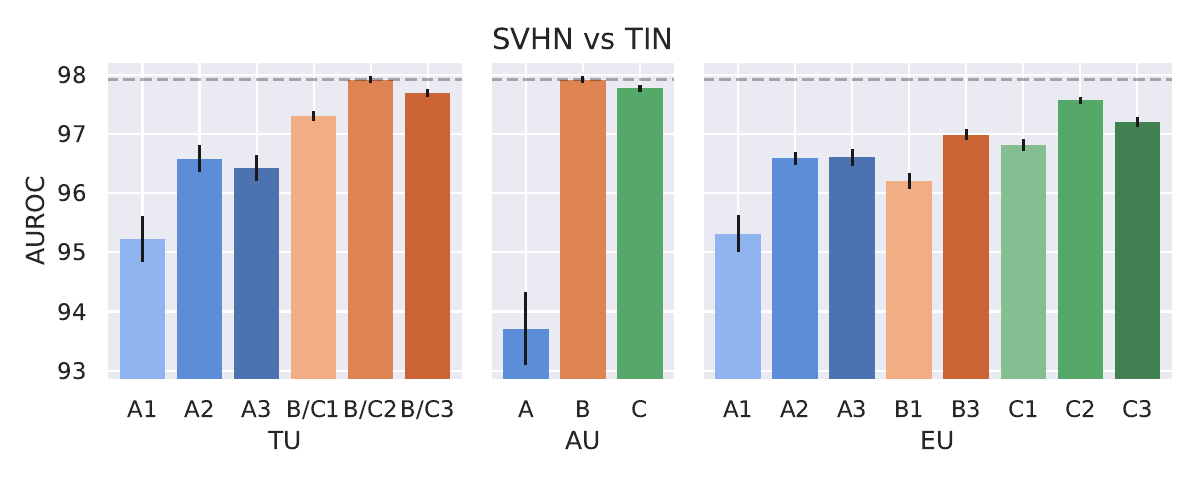}
        \subcaption{SVHN / TIN}
    \end{subfigure}
    \hfill
    \begin{subfigure}[b]{0.49\textwidth}
        \includegraphics[width=\textwidth, trim = 0.3cm 0.3cm 0.3cm 1cm, clip]{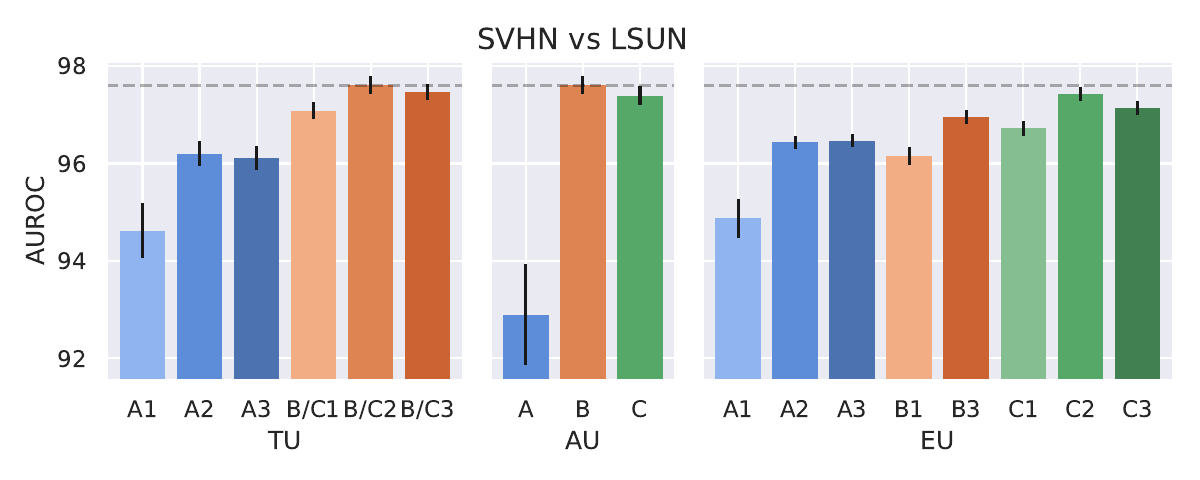}
        \subcaption{SVHN / LSUN}
    \end{subfigure}
    \caption{\textbf{OOD detection results for SVHN.} AUROC for distinguishing between ID and OOD datapoints using the different proposed measures of uncertainty as score. Means and standard deviations are calculated using five runs.}
    \label{fig:ood:det:svhn}
\end{figure}

\begin{figure}
    \centering
    \captionsetup[subfigure]{labelformat=empty}
    \captionsetup[subfigure]{aboveskip=2pt, belowskip=3pt}
    \captionsetup{aboveskip=2pt, belowskip=3pt}
    \begin{subfigure}[b]{0.49\textwidth}
        \includegraphics[width=\textwidth, trim = 0.3cm 0.3cm 0.3cm 1cm, clip]{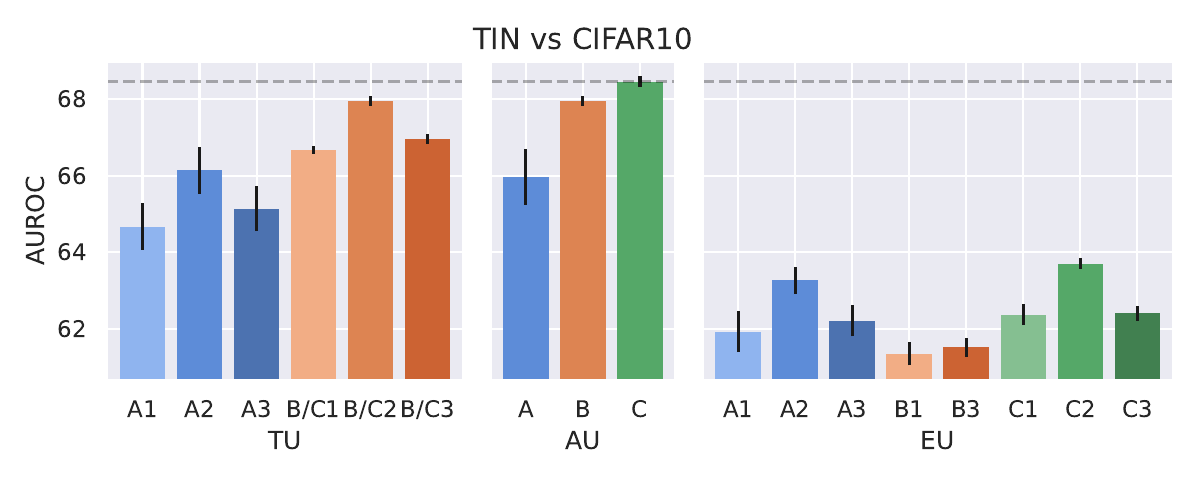}
        \subcaption{TIN / CIFAR10}
    \end{subfigure}
    \hfill
    \begin{subfigure}[b]{0.49\textwidth}
        \includegraphics[width=\textwidth, trim = 0.3cm 0.3cm 0.3cm 1cm, clip]{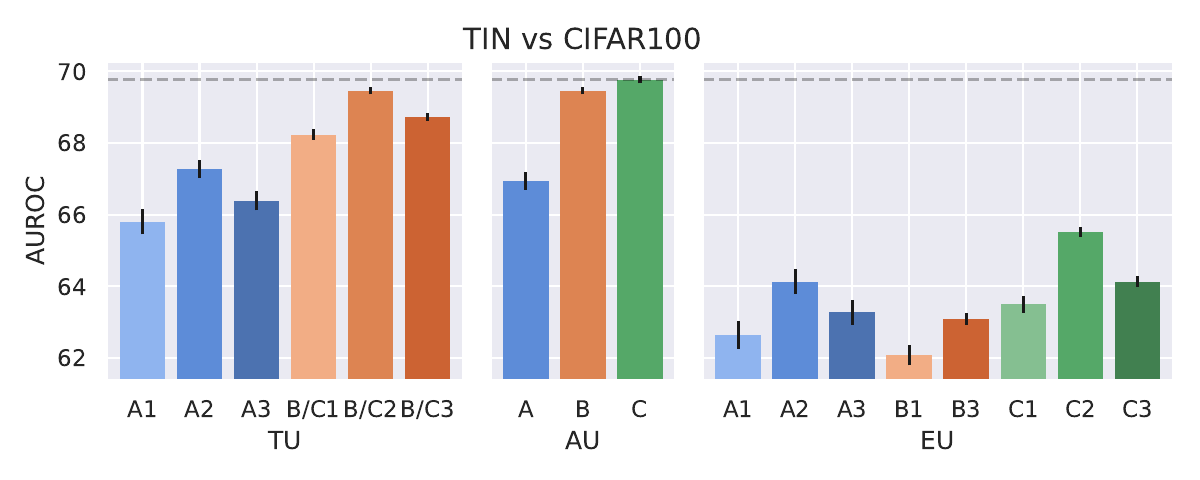}
        \subcaption{TIN / CIFAR100}
    \end{subfigure}
    \hfill
    \begin{subfigure}[b]{0.49\textwidth}
        \includegraphics[width=\textwidth, trim = 0.3cm 0.3cm 0.3cm 1cm, clip]{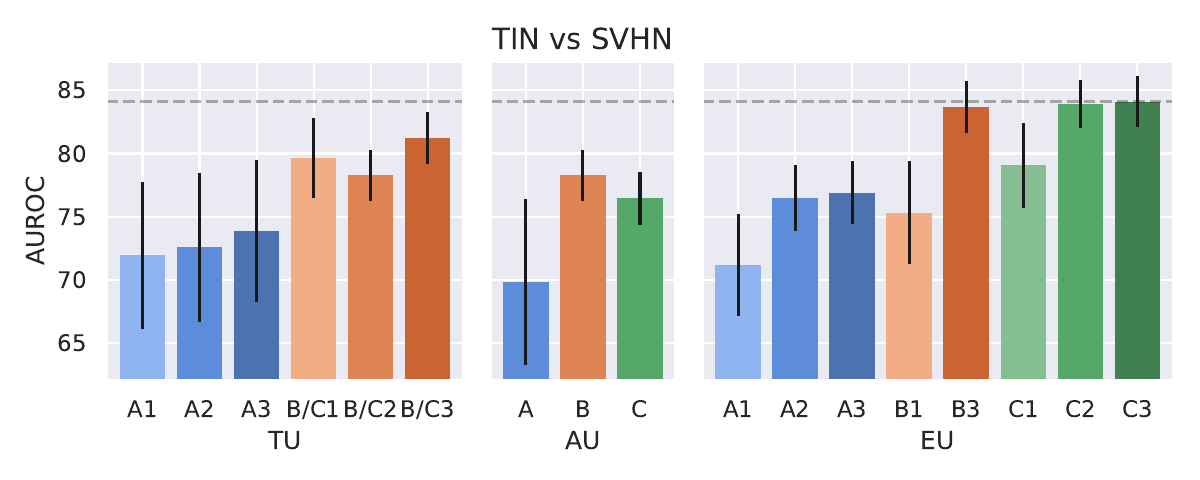}
        \subcaption{TIN / SVHN}
    \end{subfigure}
    \hfill
    \begin{subfigure}[b]{0.49\textwidth}
        \includegraphics[width=\textwidth, trim = 0.3cm 0.3cm 0.3cm 1cm, clip]{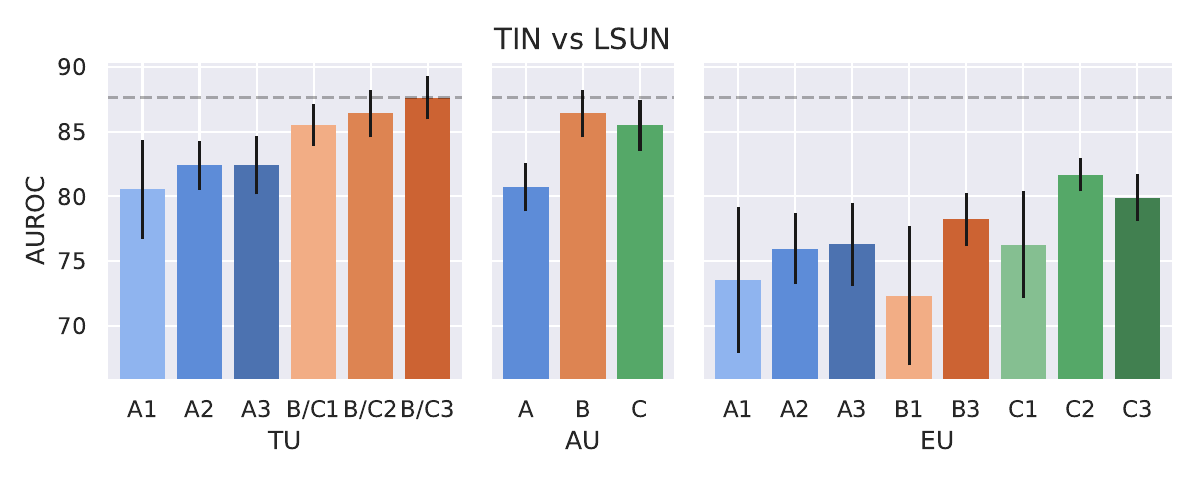}
        \subcaption{TIN / LSUN}
    \end{subfigure}
    \caption{\textbf{OOD detection results for TIN.} AUROC for distinguishing between ID and OOD datapoints using the different proposed measures of uncertainty as score. Means and standard deviations are calculated using five runs.}
    \label{fig:ood:det:tin}
\end{figure}

\paragraph{OOD detection.}
The detailed results for OOD detection for CIFAR10 as ID dataset are given in Fig.~\ref{fig:ood:det:cifar10}, for CIFAR100 as ID dataset in Fig.~\ref{fig:ood:det:cifar100}, for SVHN as ID dataset in Fig.~\ref{fig:ood:det:svhn} and for TIN as ID dataset in Fig.~\ref{fig:ood:det:tin}.
We observe the highest variability of experiments for TIN as ID dataset, where there is high variability depending on the OOD dataset.
For the other ID datasets, the different OOD datasets lead to very similar results.

\clearpage

\begin{figure}
    \centering
    \captionsetup[subfigure]{labelformat=empty}
    \captionsetup[subfigure]{aboveskip=2pt, belowskip=3pt}
    \captionsetup{aboveskip=2pt, belowskip=3pt}
    \begin{subfigure}[b]{0.49\textwidth}
        \includegraphics[width=\textwidth, trim = 0.3cm 0.3cm 0.3cm 1cm, clip]{figures/ood/detail/ood_det_deep_ensemble_resnet18_auroc_cifar10_cifar100.pdf}
        \subcaption{CIFAR100 / CIFAR10}
    \end{subfigure}
    \hfill
    \begin{subfigure}[b]{0.49\textwidth}
        \includegraphics[width=\textwidth, trim = 0.3cm 0.3cm 0.3cm 1cm, clip]{figures/ood/detail/ood_det_deep_ensemble_resnet18_auroc_cifar10_svhn.pdf}
        \subcaption{CIFAR100 / SVHN}
    \end{subfigure}
    \hfill
    \begin{subfigure}[b]{0.49\textwidth}
        \includegraphics[width=\textwidth, trim = 0.3cm 0.3cm 0.3cm 1cm, clip]{figures/ood/detail/ood_det_deep_ensemble_resnet18_auroc_cifar10_tin.pdf}
        \subcaption{CIFAR100 / TIN}
    \end{subfigure}
    \hfill
    \begin{subfigure}[b]{0.49\textwidth}
        \includegraphics[width=\textwidth, trim = 0.3cm 0.3cm 0.3cm 1cm, clip]{figures/ood/detail/ood_det_deep_ensemble_resnet18_auroc_cifar10_lsun.pdf}
        \subcaption{CIFAR100 / LSUN}
    \end{subfigure}
    \caption{\textbf{OOD detection results for CIFAR100.} AUROC for distinguishing between ID and OOD datapoints using the different proposed measures of uncertainty as score. Means and standard deviations are calculated using five runs.}
    \label{fig:ood:det:cifar100}
\end{figure}

\clearpage

\subsection{Different Network Architecture} \label{apx:sec:networks}

\begin{figure}[h!]
    \centering
    \captionsetup[subfigure]{labelformat=empty}
    \captionsetup[subfigure]{aboveskip=2pt, belowskip=3pt}
    \captionsetup{aboveskip=2pt, belowskip=3pt}
    \begin{subfigure}[b]{0.495\textwidth}
        \includegraphics[width=\textwidth, trim = 0.3cm 0.3cm 0.3cm 0.3cm, clip]{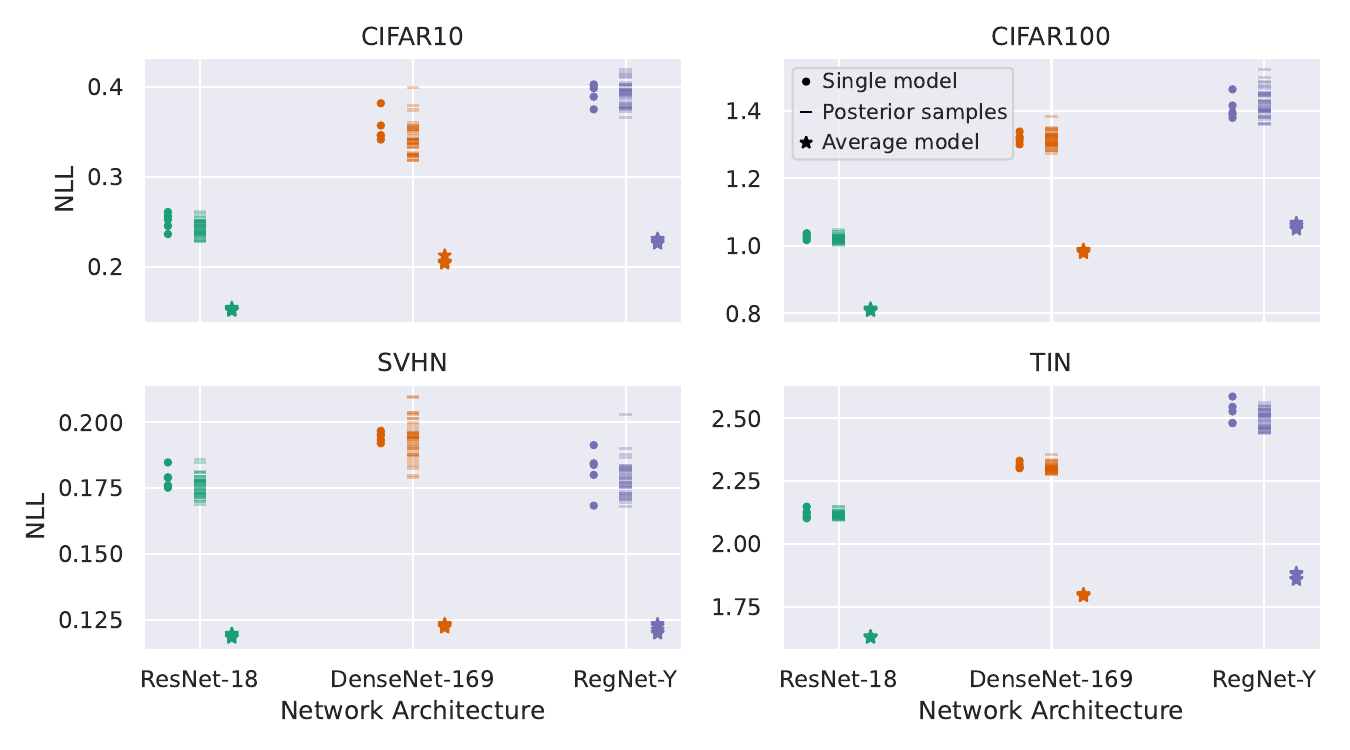}
        \subcaption{NLLs}
    \end{subfigure}
    \begin{subfigure}[b]{0.495\textwidth}
        \includegraphics[width=\textwidth, trim = 0.3cm 0.3cm 0.3cm 0.3cm, clip]{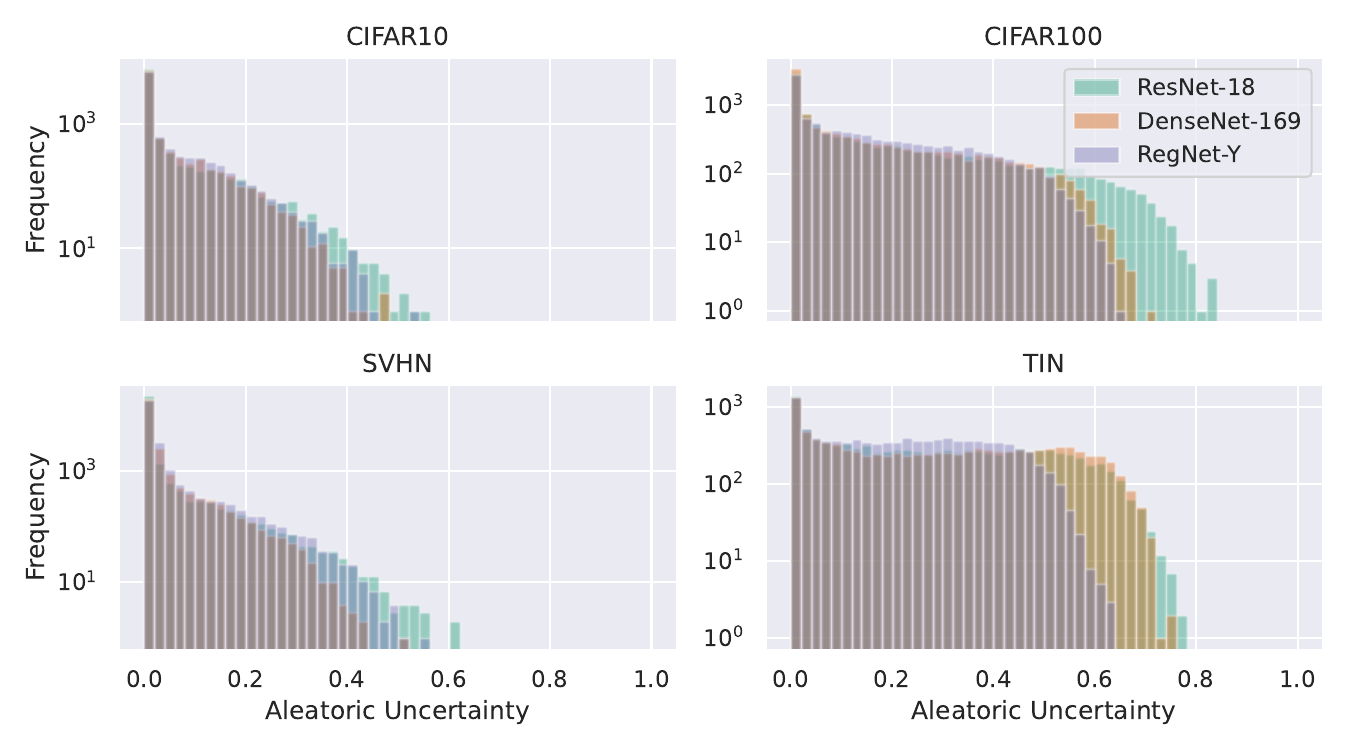}
        \subcaption{AU \texttt{(C)}$/ \log(|\mathcal{Y}|)$}
    \end{subfigure}
    \begin{subfigure}[b]{0.495\textwidth}
        \includegraphics[width=\textwidth, trim = 0.3cm 0.3cm 0.3cm 0.3cm, clip]{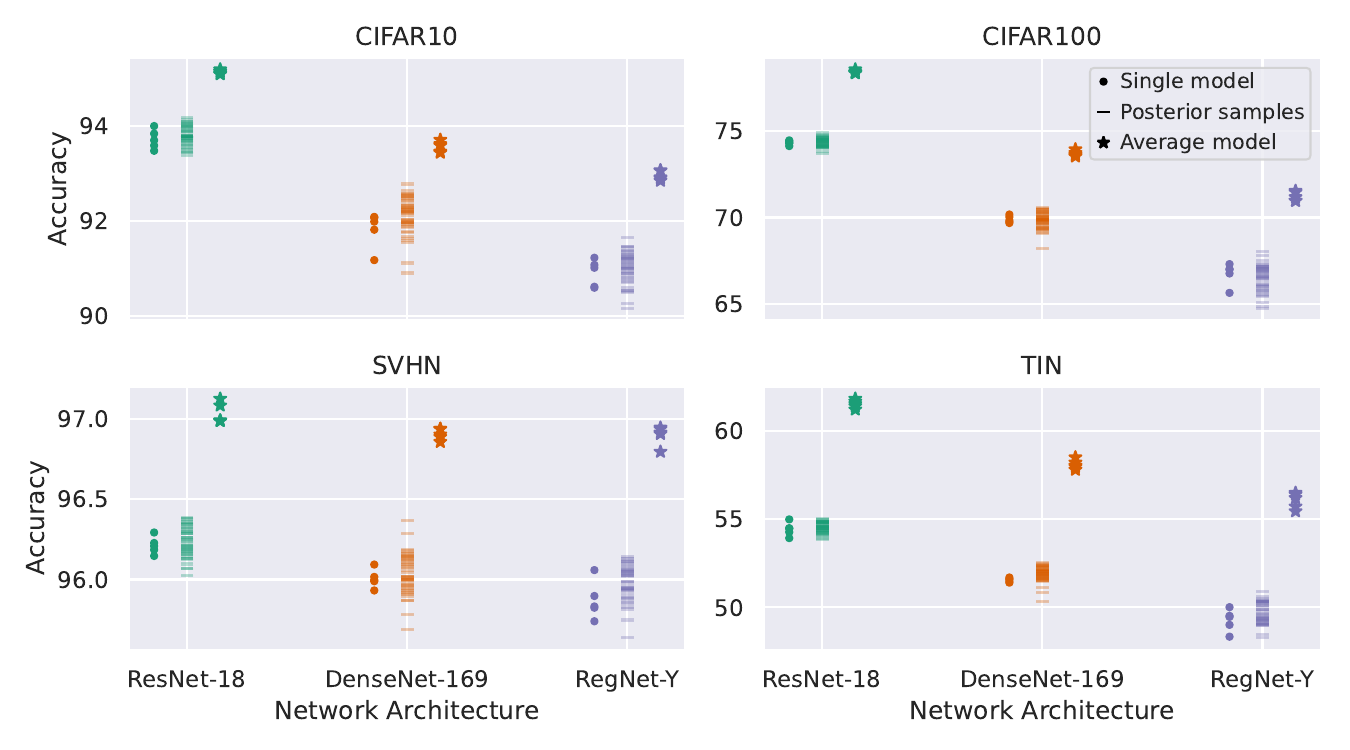}
        \subcaption{Accuracies}
    \end{subfigure}
    \begin{subfigure}[b]{0.495\textwidth}
        \includegraphics[width=\textwidth, trim = 0.3cm 0.3cm 0.3cm 0.3cm, clip]{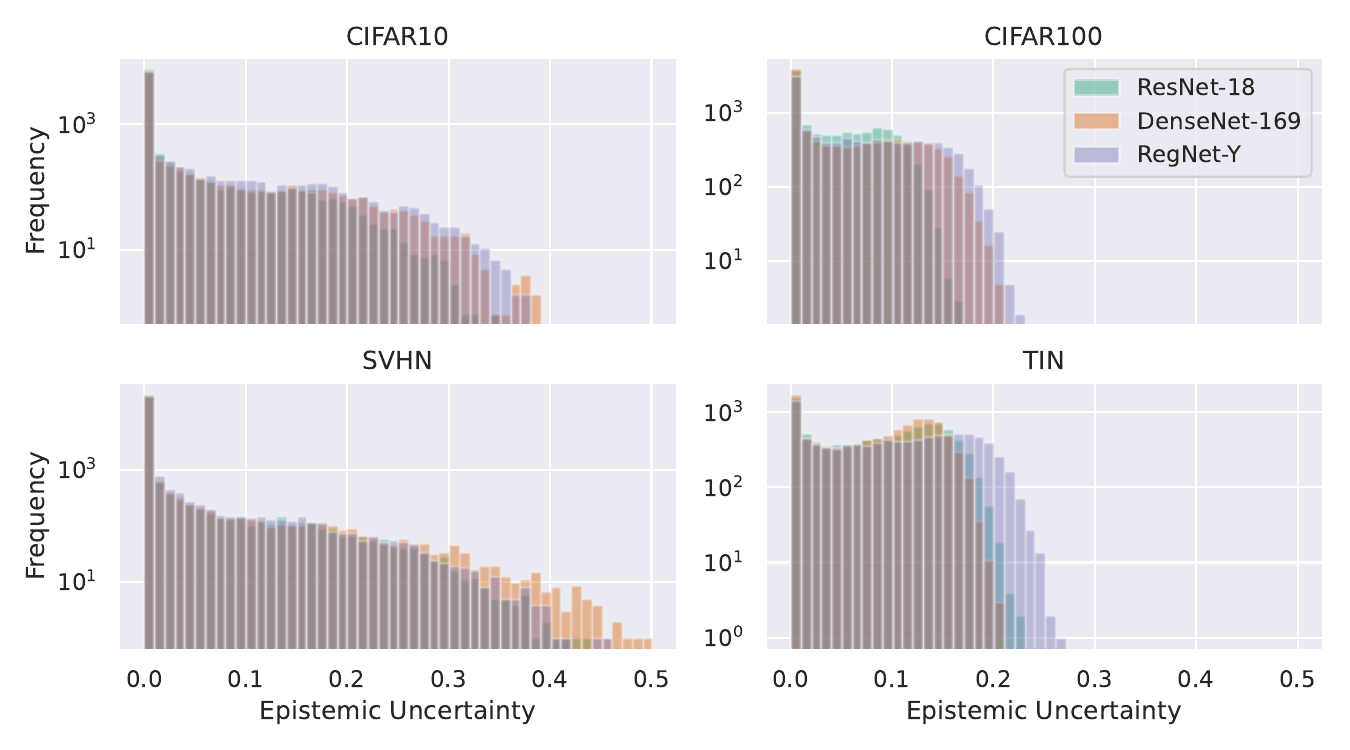}
        \subcaption{EU \texttt{(C2)}$/ \log(|\mathcal{Y}|)$}
    \end{subfigure}
    \caption{\textbf{Comparison of network architectures.} Results are obtained on the test split of the respective dataset. We compare the NLLs and accuracies for different models obtained through DEs on ResNet-18, DenseNet-169 and RegNet-Y 800MF. 
    The single model is randomly selected among all sampled models. We depict all models sampled in five runs. Furthermore, the normalized AU \texttt{(C)} and the normalized EU \texttt{(C2)} are given per sampling method. All three network architectures lead to similar results on all considered datasets.}
    \label{fig:posterior_network_comp}
\end{figure}

We want to assess the influence of the network architecture on the ranking of the results.
To that end, we also trained DEs of DenseNet169 and RegNet-Y 800MF, using the same training recipe as for ResNet-18 described in Sec.~\ref{sec:experiments}.
A comparison of the sampled models is given in Fig.~\ref{fig:posterior_network_comp}.
We observe that ResNet-18 performs a bit better than the two other models, with RegNet-Y 800MF being the worst models in terms of negative log-likelihood (NLL) and accuracies.
In terms of AU and EU, we observe only minor differences in the upper tails of the distributions for CIFAR100 and TIN. 
For CIFAR10 and SVHN, we do not observe differences.
Next, we analyze the influence of the network architecture on the misclassification and OOD detection tasks.

\begin{figure}
    \centering
    \captionsetup[subfigure]{labelformat=empty}
    \captionsetup[subfigure]{aboveskip=2pt, belowskip=3pt}
    \captionsetup{aboveskip=2pt, belowskip=3pt}
    \begin{subfigure}[b]{0.33\textwidth}
        \includegraphics[width=\textwidth, trim = 0.3cm 0.3cm 0.3cm 1cm, clip]{figures/misclassification/misc_deep_ensemble_resnet18_auroc_single_overall.pdf}
        \subcaption{\textit{(i) single model} - ResNet}
    \end{subfigure}
    \hfill
    \begin{subfigure}[b]{0.33\textwidth}
        \includegraphics[width=\textwidth, trim = 0.3cm 0.3cm 0.3cm 1cm, clip]{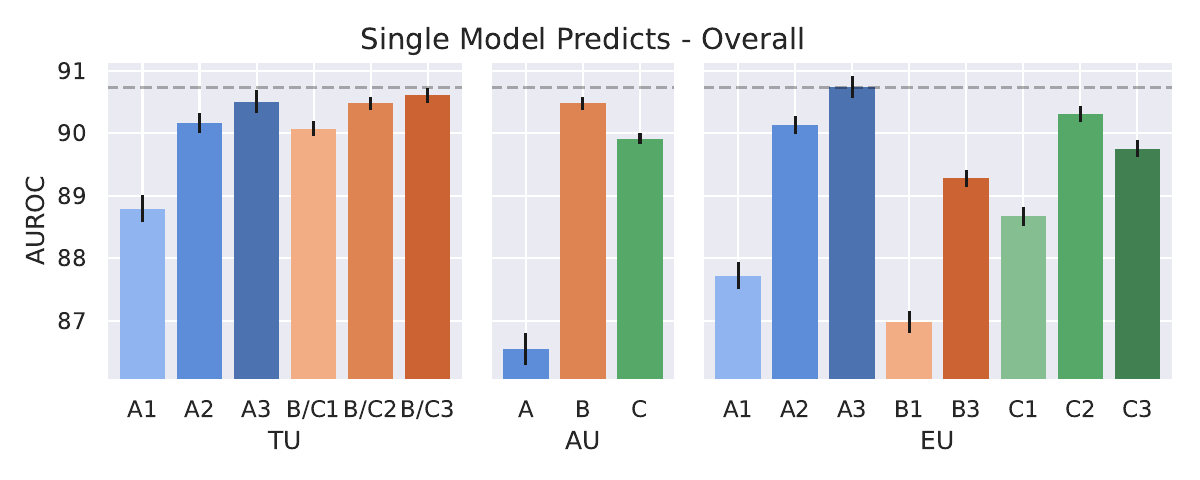}
        \subcaption{\textit{(i) single model} - DenseNet}
    \end{subfigure}
    \hfill
    \begin{subfigure}[b]{0.33\textwidth}
        \includegraphics[width=\textwidth, trim = 0.3cm 0.3cm 0.3cm 1cm, clip]{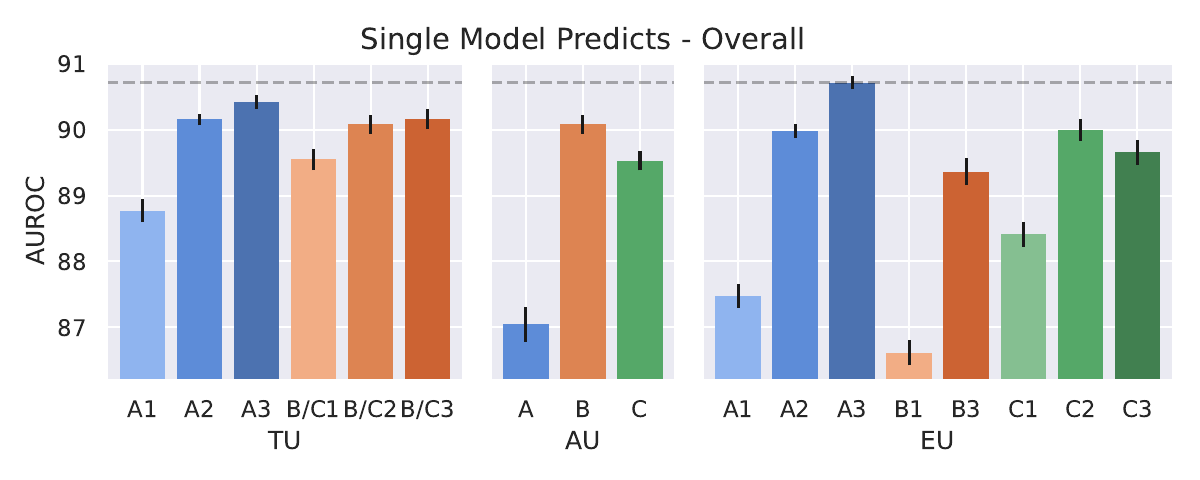}
        \subcaption{\textit{(i) single model} - RegNet}
    \end{subfigure}
    \hfill
    \begin{subfigure}[b]{0.33\textwidth}
        \includegraphics[width=\textwidth, trim = 0.3cm 0.3cm 0.3cm 1cm, clip]{figures/misclassification/misc_deep_ensemble_resnet18_auroc_average_overall.pdf}
        \subcaption{\textit{(ii) average model} - ResNet}
    \end{subfigure}
    \hfill
    \begin{subfigure}[b]{0.33\textwidth}
        \includegraphics[width=\textwidth, trim = 0.3cm 0.3cm 0.3cm 1cm, clip]{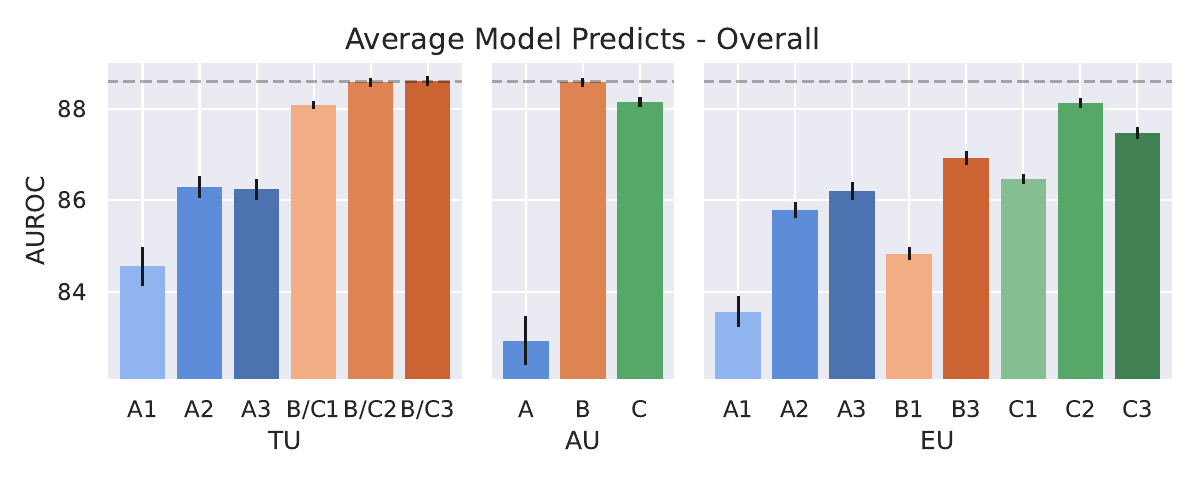}
        \subcaption{\textit{(ii) average model} - DenseNet}
    \end{subfigure}
    \hfill
    \begin{subfigure}[b]{0.33\textwidth}
        \includegraphics[width=\textwidth, trim = 0.3cm 0.3cm 0.3cm 1cm, clip]{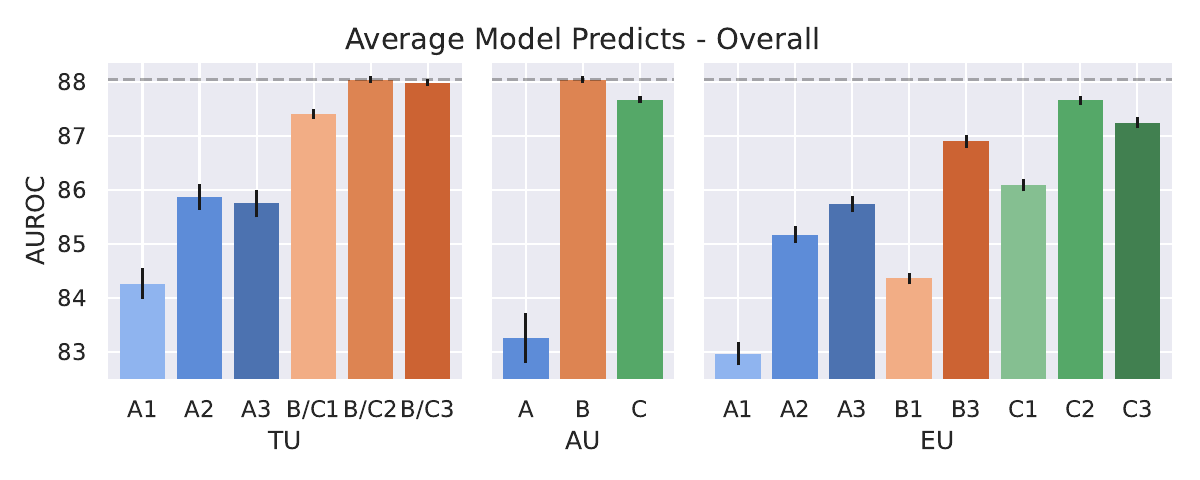}
        \subcaption{\textit{(ii) average model} - RegNet}
    \end{subfigure}
    \hfill
    \begin{subfigure}[b]{0.33\textwidth}
        \includegraphics[width=\textwidth, trim = 0.3cm 0.3cm 0.3cm 1cm, clip]{figures/misclassification/misc_deep_ensemble_resnet18_auroc_posterior_overall.pdf}
        \subcaption{\textit{(iii) according to posterior} - ResNet}
    \end{subfigure}
    \hfill
    \begin{subfigure}[b]{0.33\textwidth}
        \includegraphics[width=\textwidth, trim = 0.3cm 0.3cm 0.3cm 1cm, clip]{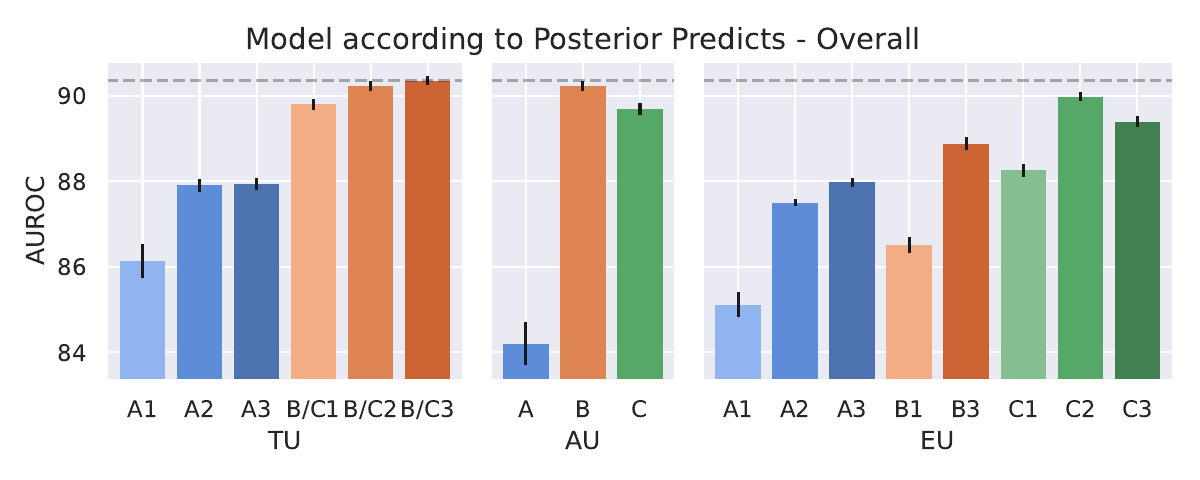}
        \subcaption{\textit{(iii) acc. to posterior} - DenseNet}
    \end{subfigure}
    \hfill
    \begin{subfigure}[b]{0.33\textwidth}
        \includegraphics[width=\textwidth, trim = 0.3cm 0.3cm 0.3cm 1cm, clip]{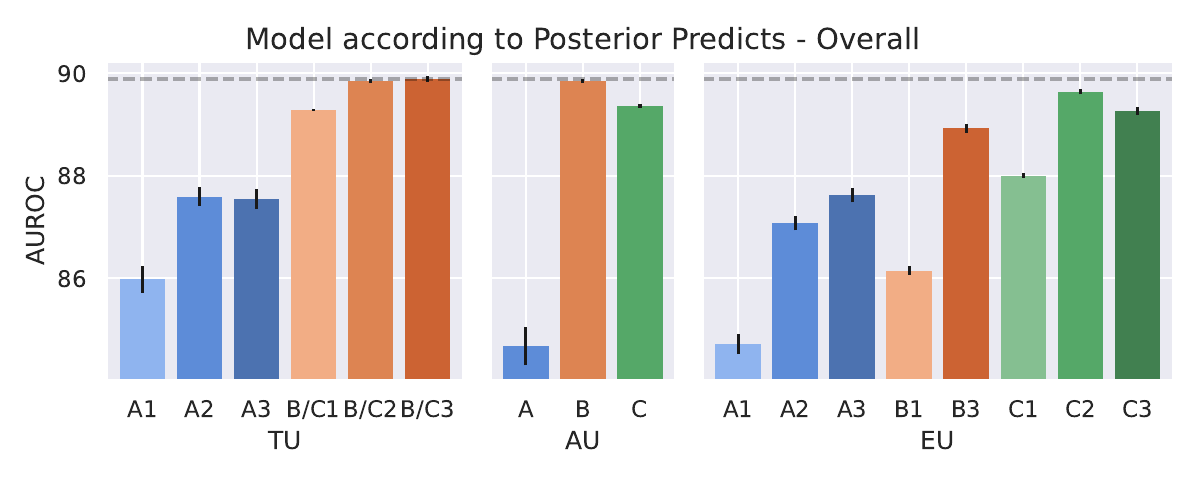}
        \subcaption{\textit{(iii) according to posterior} - RegNet}
    \end{subfigure}
    \caption{\textbf{Misclassification detection results for DE with different model architectures and under different predicting models.} AUROC for distinguishing between correctly and incorrectly predicted samples under different predicting models, using the different proposed measures of uncertainty as score. Means and standard deviations are calculated using five runs.}
    \label{fig:misc:network}
\end{figure}

\paragraph{Misclassification detection.}
The results for misclassification detection using DEs with different model architectures are given in Fig.~\ref{fig:misc:network}.
We observe no major differences for different models (per column) under a given predicting model (per row).

\begin{figure}
    \centering
    \captionsetup[subfigure]{labelformat=empty}
    \captionsetup[subfigure]{aboveskip=2pt, belowskip=3pt}
    \captionsetup{aboveskip=2pt, belowskip=3pt}
    \begin{subfigure}[b]{0.33\textwidth}
        \includegraphics[width=\textwidth, trim = 0.3cm 0.3cm 0.3cm 1cm, clip]{figures/ood/ood_deep_ensemble_resnet18_auroc_overall.pdf}
        \subcaption{ResNet-18}
    \end{subfigure}
    \hfill
    \begin{subfigure}[b]{0.33\textwidth}
        \includegraphics[width=\textwidth, trim = 0.3cm 0.3cm 0.3cm 1cm, clip]{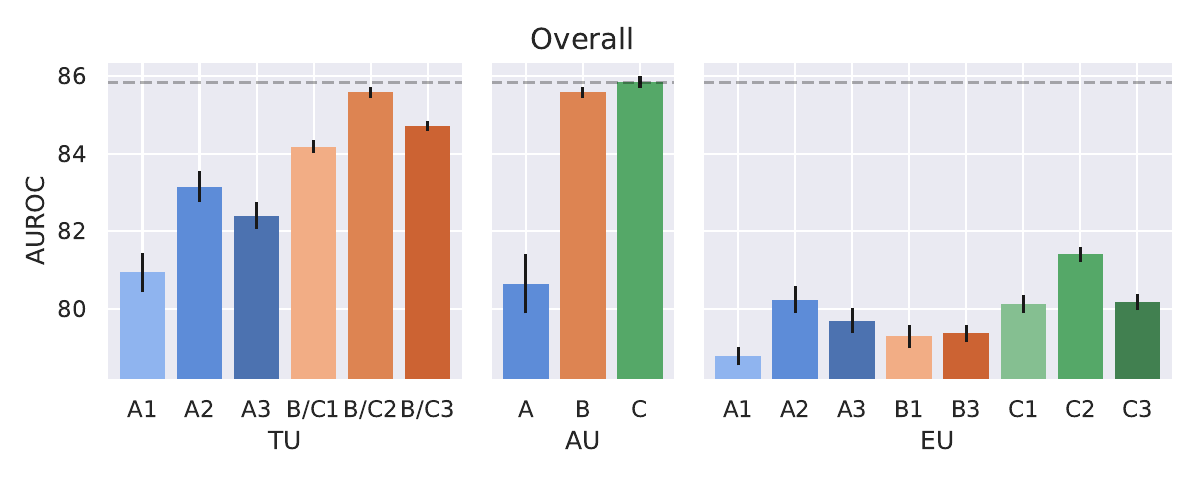}
        \subcaption{DenseNet-169}
    \end{subfigure}
    \hfill
    \begin{subfigure}[b]{0.33\textwidth}
        \includegraphics[width=\textwidth, trim = 0.3cm 0.3cm 0.3cm 1cm, clip]{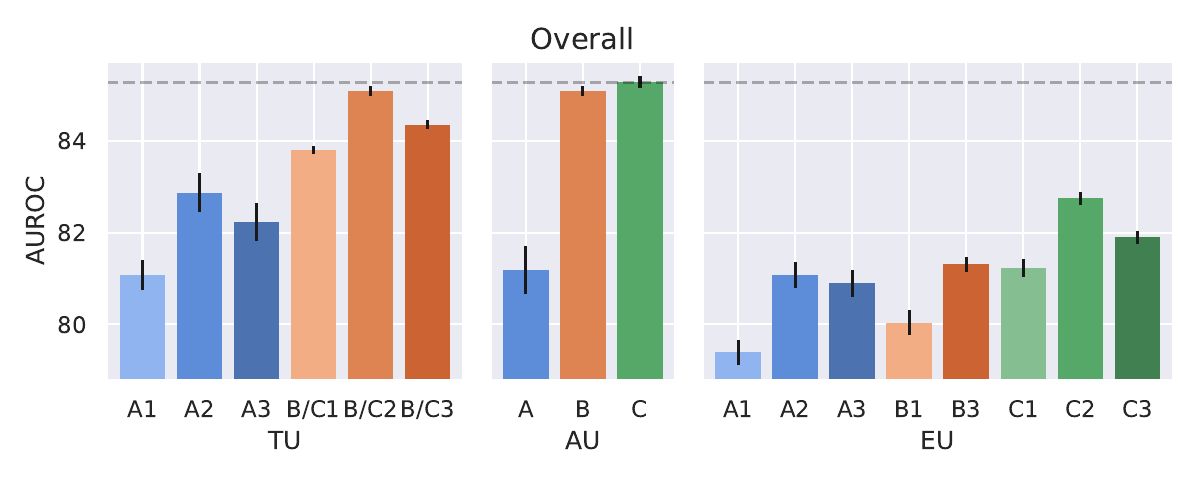}
        \subcaption{RegNet-Y 800MF}
    \end{subfigure}
    \caption{\textbf{OOD detection results for DE with different model architectures.} AUROC for distinguishing between ID and OOD datapoints using the different proposed measures of uncertainty as score. AUROCs are averaged over all ID / OOD combinations. Means and standard deviations are calculated using five runs.}
    \label{fig:ood:network}
\end{figure}

\paragraph{OOD detection.}
The results for OOD detection using DEs with different model architectures are given in Fig.~\ref{fig:ood:network}.
We observe that the AU \texttt{(C)} is the best measure for DenseNet-169 and RegNet-Y 800MF, while it is AU \texttt{(B)} which is equivalent to TU \texttt{(B/C2)} for ResNet-18.
However, the general trends are the same across all architectures.

\clearpage

\subsection{Adversarial Example Detection} \label{sec:apx:adversarial}

We want to investigate the effect of adversarially created inputs on the uncertainty estimates.
Throughout this experiments, we consider adversarial attacks on the single network.
However, it would also be possible to attack the average model, albeit more computationally expensive.
As adversarial examples are known to transfer well between models of similar architecture \citep{Goodfellow:15}, results for attacking the average model are expected to be relatively similar to those presented here.

We consider two different adversarial attacks, FGSM \citep{Goodfellow:15} and PGD under infinity norm perturbation \citep{Madry:18}.
For our experiments, we only consider the subset of the test datasets that are predicted correctly.
This we refer to as the \textit{original} dataset.
Then we apply the adversarial attacks the datapoints in the original dataset and select those datapoints where the model was successfully fooled to predict incorrectly.
This we refer to as the \textit{adversarial} dataset.
We utilize the different uncertainty scores to calculate the AUROC of distinguishing between the original and the adversarial dataset, akin to the OOD detection experiments reported in the main paper.
We also investigated the AUPR and FPR@TPR95 as alternative metrics, which lead to equivalent conclusions.

\begin{figure}
    \centering
    \captionsetup[subfigure]{labelformat=empty}
    \captionsetup[subfigure]{aboveskip=2pt, belowskip=3pt}
    \captionsetup{aboveskip=2pt, belowskip=3pt}
    \begin{subfigure}[b]{0.49\textwidth}
        \includegraphics[width=\textwidth, trim = 0.3cm 0.3cm 0.3cm 0.3cm, clip]{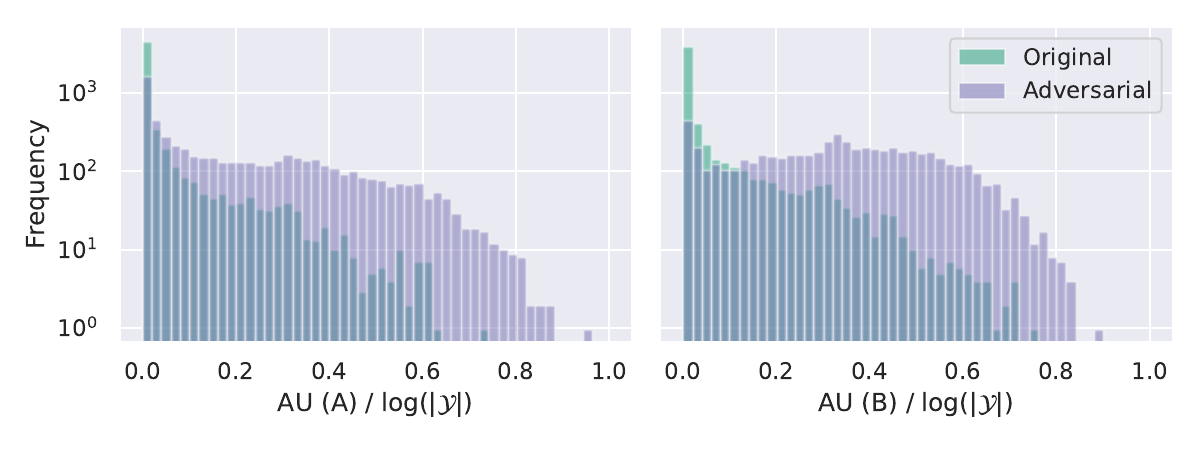}
        \subcaption{CIFAR10}
    \end{subfigure}
    \hfill
    \begin{subfigure}[b]{0.49\textwidth}
        \includegraphics[width=\textwidth, trim = 0.3cm 0.3cm 0.3cm 0.3cm, clip]{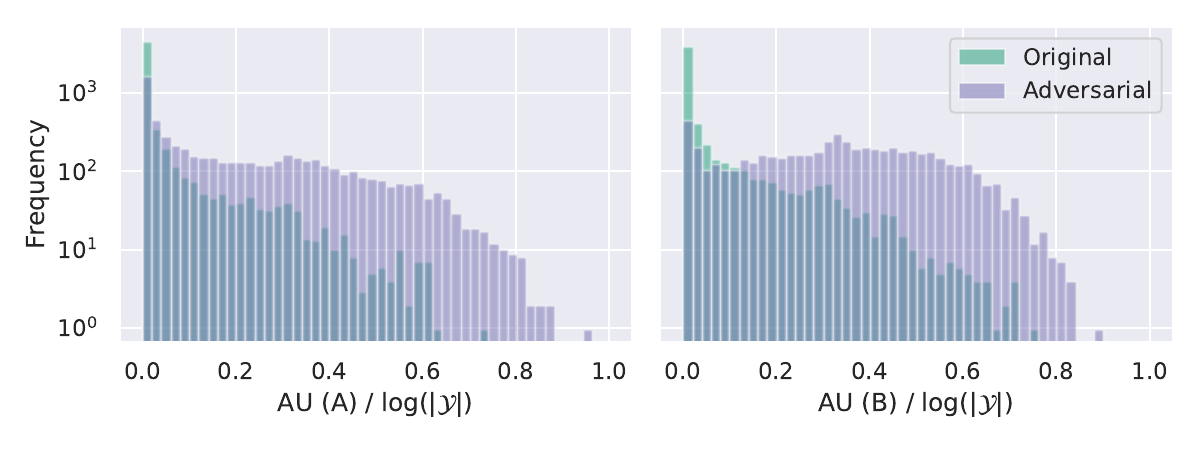}
        \subcaption{CIFAR100}
    \end{subfigure}
    \hfill
    \begin{subfigure}[b]{0.49\textwidth}
        \includegraphics[width=\textwidth, trim = 0.3cm 0.3cm 0.3cm 0.3cm, clip]{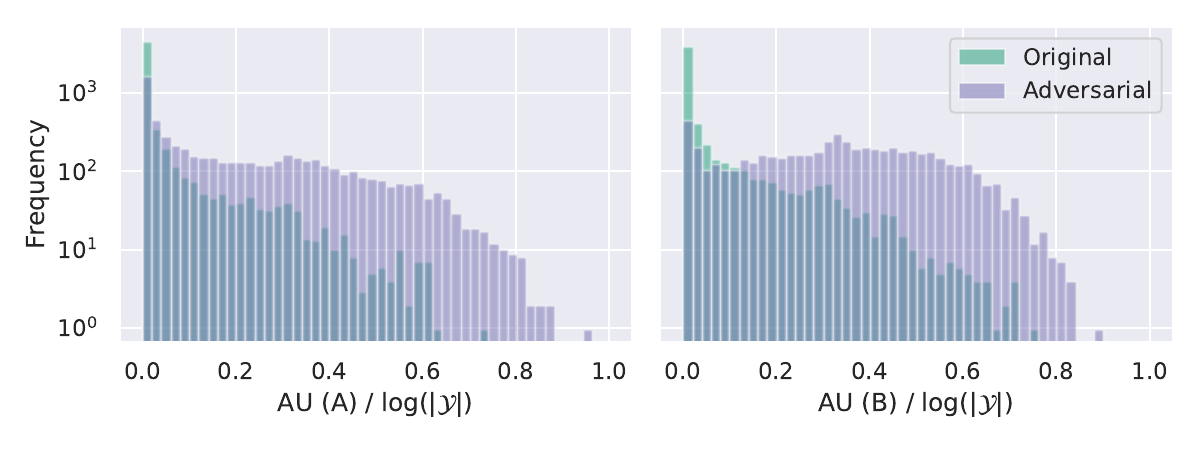}
        \subcaption{SVHN}
    \end{subfigure}
    \hfill
    \begin{subfigure}[b]{0.49\textwidth}
        \includegraphics[width=\textwidth, trim = 0.3cm 0.3cm 0.3cm 0.3cm, clip]{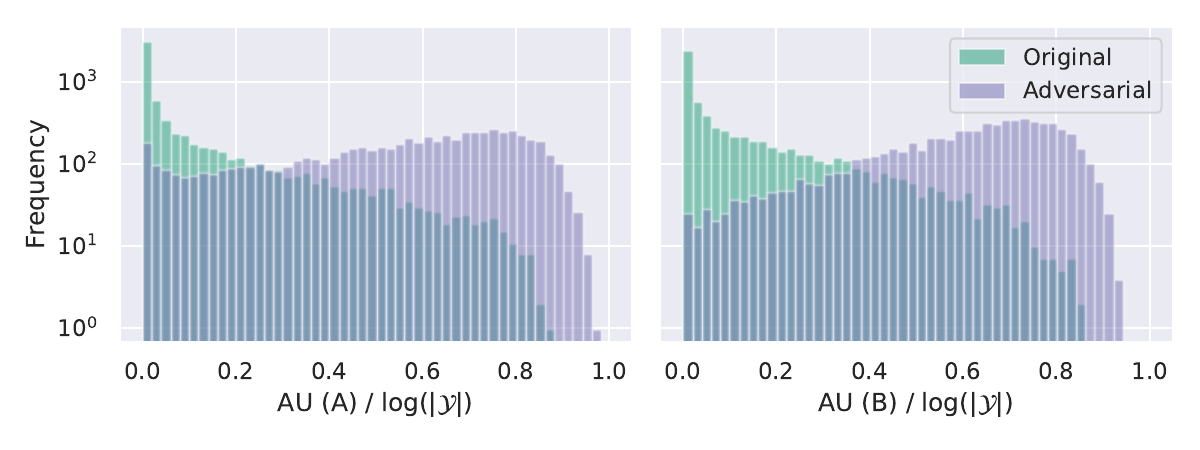}
        \subcaption{TIN}
    \end{subfigure}
    \caption{\textbf{Histogram of AU \texttt{(A)} and AU \texttt{(B)} for original and adversarial datapoints obtained through applying FGSM.} AUs are normalized with $\log (|\mathcal{Y}|)$ to be more comparable across datasets with different number of classes.}
    \label{fig:ae:fgsm_hist_de}
\end{figure}

\paragraph{FGSM.}
We start with the results obtained through the FGSM attack with $\epsilon = 8/255$.
Histograms of the AU \texttt{(A)}, the entropy of the predictive distribution of the single attacked model and the AU \texttt{(B)}, the entropy of the predictive distribution under the average model, are shown in Fig.~\ref{fig:ae:fgsm_hist_de}.
We observe a shift towards higher AUs for the adversarial datapoints compared to the original datapoints.
This shift appears more pronounced for AU \texttt{(B)}, which makes sense as the adversarial examples have been obtained with the single model.

\begin{figure}
    \centering
    \captionsetup[subfigure]{labelformat=empty}
    \includegraphics[width=0.5\textwidth, trim = 0.3cm 0.3cm 0.3cm 1cm, clip]{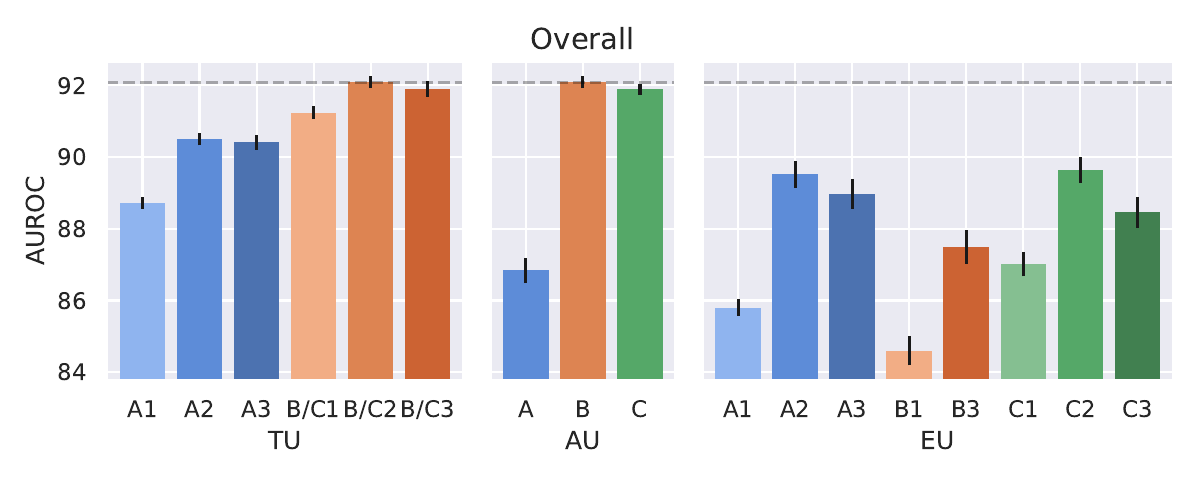}
    \caption{\textbf{Adversarial example detection (FGSM).} Means and standard deviations are calculated on five runs.}
    \label{fig:ae:fgsm}
\end{figure}

The main results are shown in Fig.~\ref{fig:ae:fgsm}, denoting the AUROC of distinguishing between the original and the adversarial datapoints using the different measures of uncertainty as score.
We observe qualitatively very similar results to the OOD detection experiments, in that TU and AU measures for cases \texttt{(B)} and \texttt{(C)} are the most effective.
Surprisingly, EU measures underperform for adversarial example detection, compared to TU and AU measures.

\begin{figure}
    \centering
    \captionsetup[subfigure]{labelformat=empty}
    \captionsetup[subfigure]{aboveskip=2pt, belowskip=3pt}
    \captionsetup{aboveskip=2pt, belowskip=3pt}
    \begin{subfigure}[b]{0.49\textwidth}
        \includegraphics[width=\textwidth, trim = 0.3cm 0.3cm 0.3cm 0.3cm, clip]{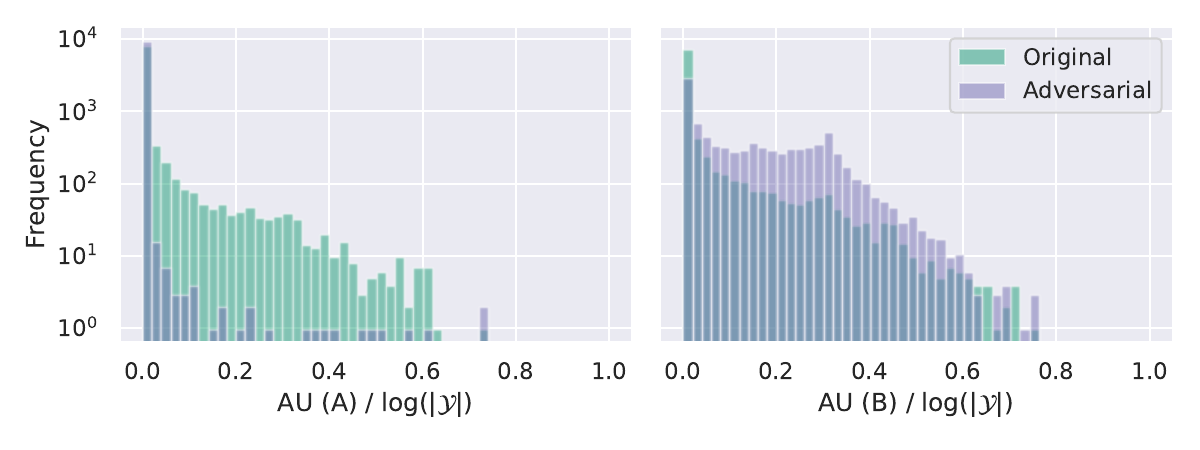}
        \subcaption{CIFAR10}
    \end{subfigure}
    \hfill
    \begin{subfigure}[b]{0.49\textwidth}
        \includegraphics[width=\textwidth, trim = 0.3cm 0.3cm 0.3cm 0.3cm, clip]{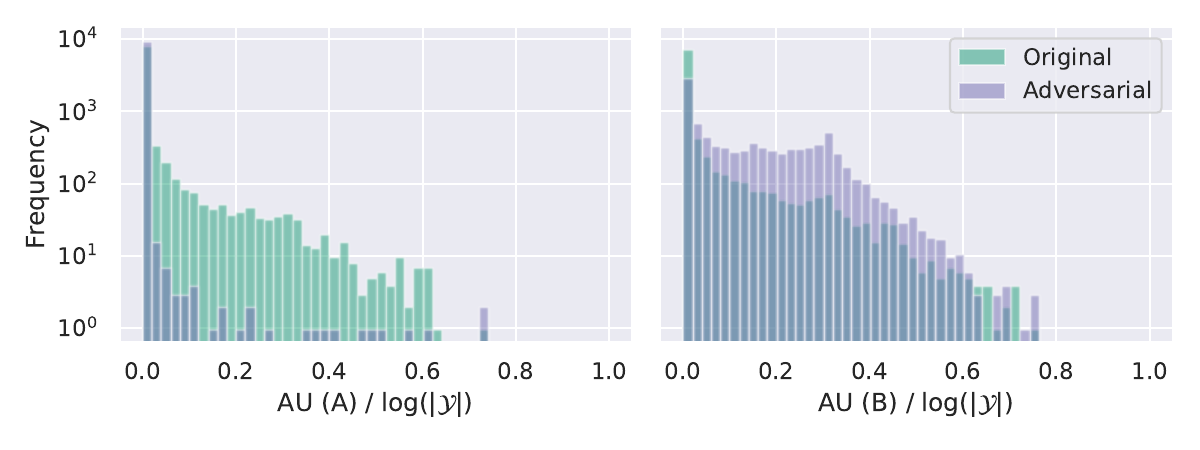}
        \subcaption{CIFAR100}
    \end{subfigure}
    \hfill
    \begin{subfigure}[b]{0.49\textwidth}
        \includegraphics[width=\textwidth, trim = 0.3cm 0.3cm 0.3cm 0.3cm, clip]{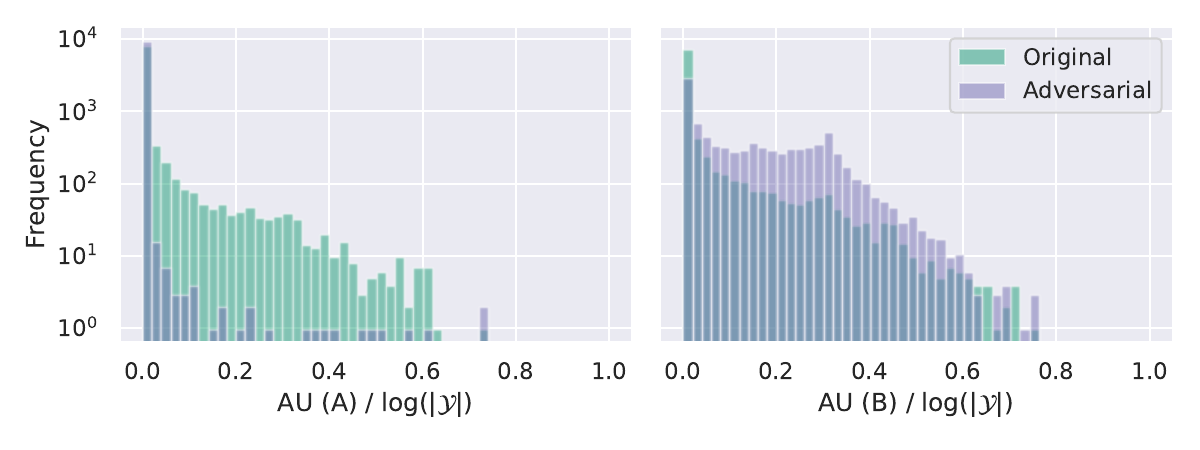}
        \subcaption{SVHN}
    \end{subfigure}
    \hfill
    \begin{subfigure}[b]{0.49\textwidth}
        \includegraphics[width=\textwidth, trim = 0.3cm 0.3cm 0.3cm 0.3cm, clip]{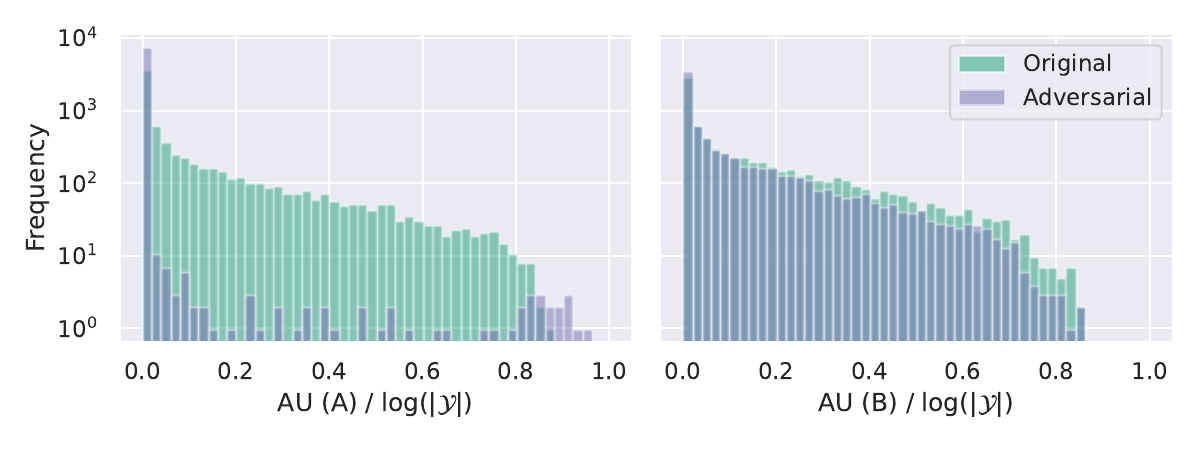}
        \subcaption{TIN}
    \end{subfigure}
    \caption{\textbf{Histogram of AU \texttt{(A)} and AU \texttt{(B)} for original and adversarial datapoints obtained through applying $L_\infty$-PGD.} AUs are normalized with $\log (|\mathcal{Y}|)$ to be more comparable across datasets with different number of classes.}
    \label{fig:ae:pgd_hist_de}
\end{figure}

\paragraph{PGD.}
Next, we conduct the same investigation using the $L_\infty$-PGD attack with $\epsilon = 8/255$.
Histograms of the AU \texttt{(A)}, the entropy of the predictive distribution of the single attacked model and the AU \texttt{(B)}, the entropy of the predictive distribution under the average model, are shown in Fig.~\ref{fig:ae:pgd_hist_de}.
We observe a shift towards \textbf{lower} AU \texttt{(A)} for the adversarial datapoints compared to the original datapoints.
However for for AU \texttt{(B)}, adversarial datapoints exhibit slightly higher values than the original datapoints.

\begin{figure}
    \centering
    \captionsetup[subfigure]{labelformat=empty}
    \begin{subfigure}[b]{0.49\textwidth}
        \includegraphics[width=\textwidth, trim = 0.3cm 0.3cm 0.3cm 1cm, clip]{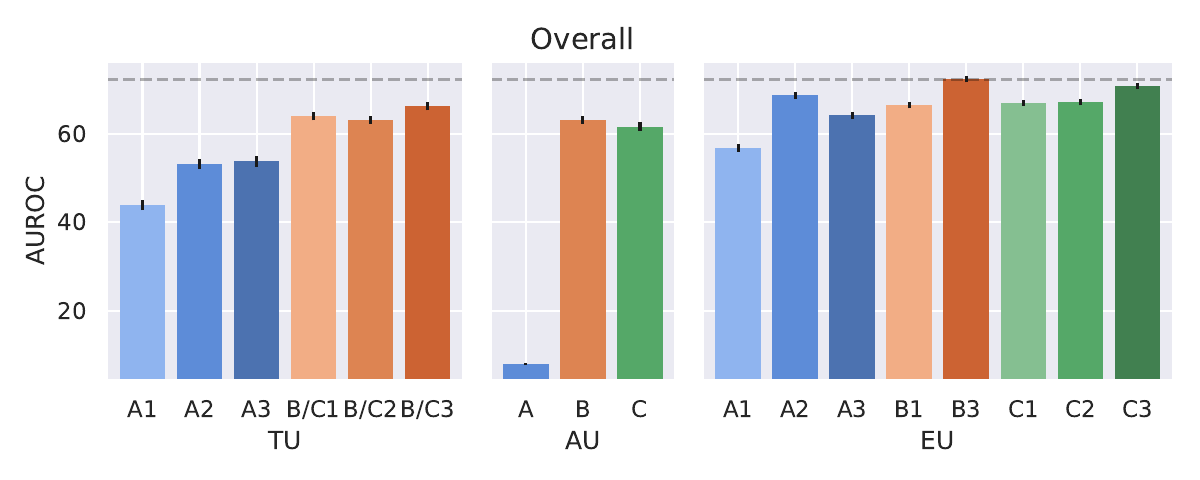}
        \subcaption{(a)}
    \end{subfigure}
    \begin{subfigure}[b]{0.49\textwidth}
        \includegraphics[width=\textwidth, trim = 0.3cm 0.3cm 0.3cm 1cm, clip]{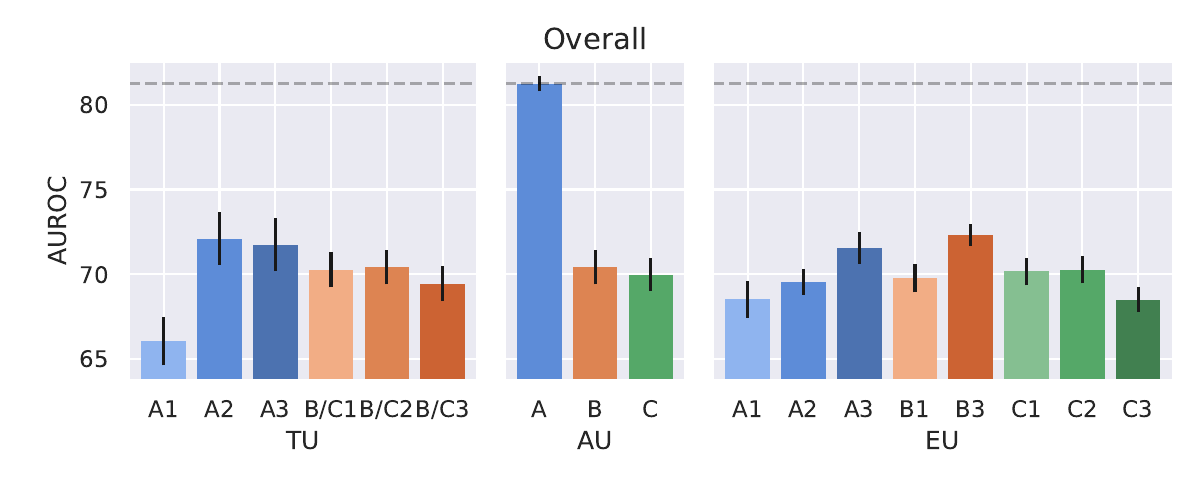}
        \subcaption{(b)}
    \end{subfigure}
    \caption{\textbf{Adversarial example detection ($\text{L}_\infty$-PGD)}. (a) Assuming higher uncertainty for adversarial examples. (b) Assuming lower uncertainty for adversarial examples. Means and standard deviations are calculated on five runs.}
    \label{fig:ae:pgd}
\end{figure}

The results are given in Fig.~\ref{fig:ae:pgd}(a), denoting the AUROC of distinguishing between the original and the adversarial datapoints using the different measures of uncertainty as score.
We observe that all measures except TU \texttt{(A1)} and AU \texttt{(A)} perform better than random.
The very bad performance of AU \texttt{(A)} stems from the fact that adversarial datapoints exhibit lower uncertainties than the original datapoints (see Fig.~\ref{fig:ae:pgd_hist_de}).
However, contrary to the experiments with FGSM, measures of EU perform best for PGD.

The two experiments for adversarial example detection were conducted under the assumption that adversarial datapoints should exhibit higher uncertainty than the original datapoints.
Finally, we investigate a special variant of our experiments with $L_\infty$-PGD adversarial examples, where we assume that adversarial datapoints exhibit \textbf{lower} uncertainty than the original datapoints.
The results are shown in Fig.~\ref{fig:ae:pgd}(b).
We observe that using AU \texttt{(A)} leads to the best results in this setting.
However, these results do not help to attain a mechanism for adversarial robustness, as we leverage additional side information that the single model was fooled into being very confident about the adversarial examples.
Attackers could add constraints on the deviation between the AU \texttt{(A)} under the original and the adversarial datapoint in an improved version of the $L_\infty$-PGD attack, rendering this detection mechanism useless.

\subsection{Active Learning} \label{apx:active_learning}

In this section, we provide additional details on the active learning experiments described in the main paper.

The network architective of the utilized small CNN are: 5x5 conv [1 to 6 channels], 2x2 max-pool, 5x5 conv [6 to 12 channels], 2x2 max-pool, two linear layers with hidden size 32 and a final output linear layer; ReLU activations after each max-pool and linear layer except the last as well as dropout with dropout rate 0.2 between linear layers.
The training of the models utilized the Adam optimizer \citep{Kingma:15} for 50 epochs with a learning rate of 1e-3, a batch size of 32 and l2 weight decay of 1e-4.
Early stopping was performed on the official validation split of the respective datasets, the evaluation of the performance per step was conducted on the official test splits.
Although the size of the training dataset increases each step, the effective size, thus the number of gradient steps per epoch, was kept constant at 1000 for the MNIST experiments and 1600 for the FMNIST experiments.

\subsection{Alternative Posterior Sampling Methods}

\begin{wrapfigure}{r}{0.34\textwidth}
    \centering
    \includegraphics[width=\linewidth]{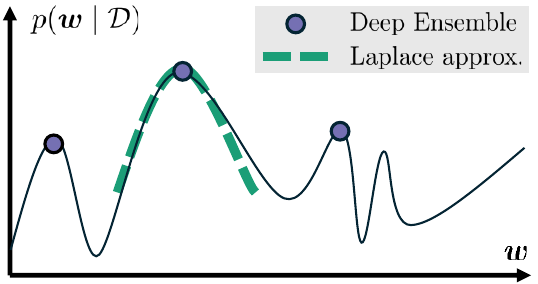}
    \caption{Posterior sampling methods.}
    \label{fig:posterior}
\end{wrapfigure}
In addition to Deep Ensembles (DE) used as posterior sampling method in the main paper, we investigate alternative posterior sampling methods.
Specifically, we consider the Laplace Approximation (LA) \citep{MacKay:92} on the last layer with Kronecker-factored approximate curvature \citep{Ritter:18} using the implementation of \citet{Daxberger:21}.
Furthermore, we consider MC Dropout (MCD) \citep{Gal:16}.
As for DE, we sample 10 models with the alternative posterior sampling methods for the MC approximations of posterior expectations.
Measures based on a single model (combinations with \texttt{(A)}) use the maximum a posteriori (MAP) model for LA, and the model without dropout activated for MCD as the predicting models.

There is a distinction between multi- and single-basin posterior sampling techniques \citep{Wilson:20}, sometimes also referred to as multi- and single-mode approaches \citep{Hoffmann:21}.
We refer to them as global and local posterior sampling techniques for simplicity.
In this categorization, DE is a global method, while LA and MCD are local methods \citep{Fort:19}, see Fig.~\ref{fig:posterior} for an illustration.
We hypothesize that different methods for posterior sampling have a strong impact on which uncertainty measure performs well empirically, especially given whether they are global or local methods.

In the following, we first analyze the characteristics of posterior samples drawn with those different sampling methods.
Then we investigate the importance of aligning the uncertainty measure with the predicting model, detecting distributional mismatch and the disentanglement of measures on these additional posterior sampling methods.

\subsubsection{Characteristics of Posterior Samples}

\begin{figure}[t]
    \centering    
    \captionsetup[subfigure]{labelformat=empty}
    \captionsetup[subfigure]{aboveskip=2pt, belowskip=3pt}
    \captionsetup{aboveskip=2pt, belowskip=3pt}
    \begin{subfigure}[b]{0.495\textwidth}
        \includegraphics[width=\textwidth, trim = 0.3cm 0.3cm 0.3cm 0.3cm, clip]{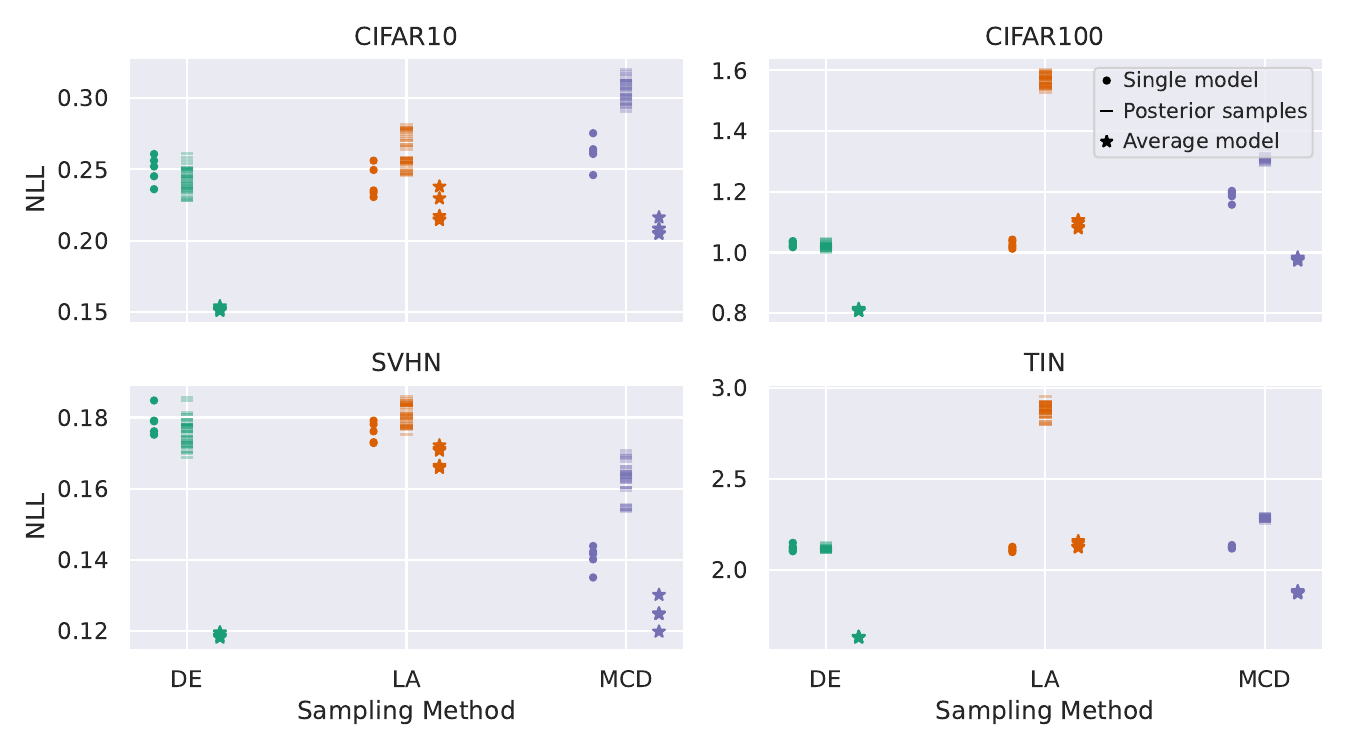}
        \subcaption{Negative Log-Likelihoods (NLLs)}
    \end{subfigure}
    \begin{subfigure}[b]{0.495\textwidth}
        \includegraphics[width=\textwidth, trim = 0.3cm 0.3cm 0.3cm 0.3cm, clip]{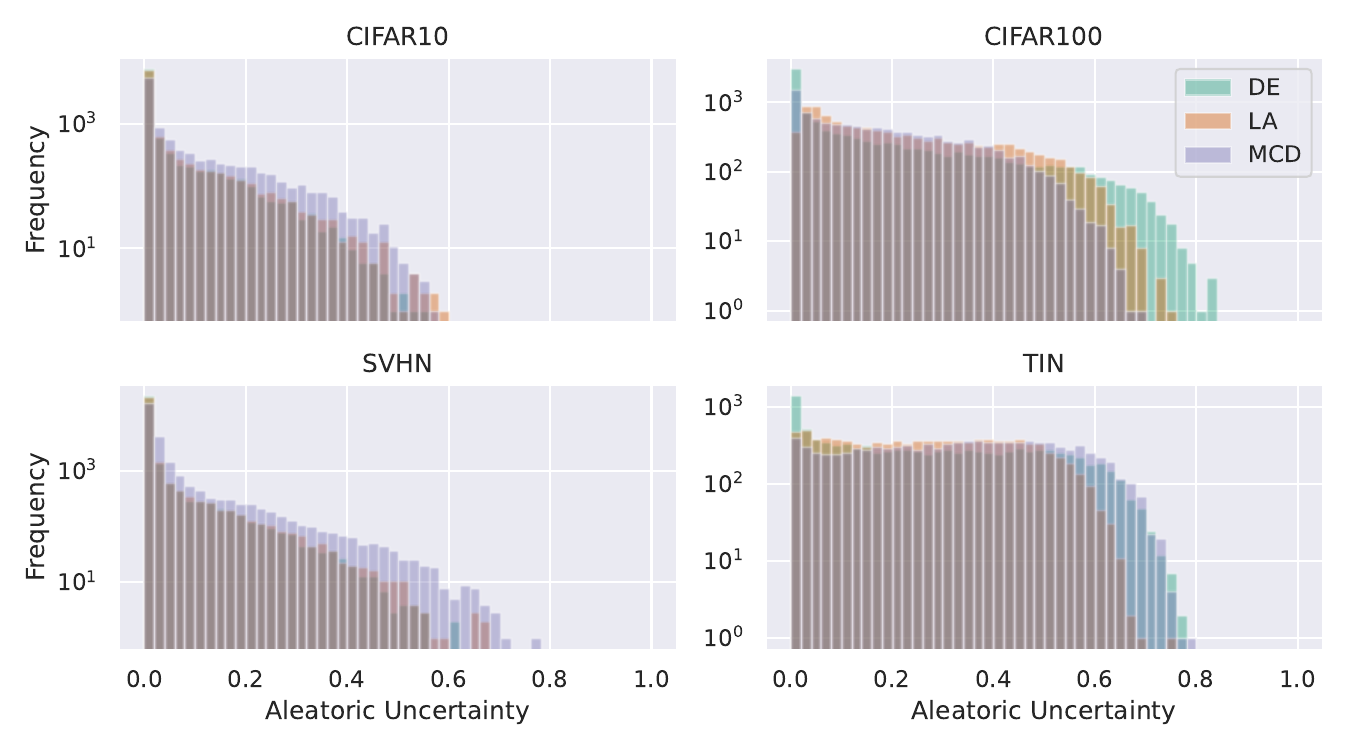}
        \subcaption{AU \texttt{(C)}$/ \log(|\mathcal{Y}|)$}
    \end{subfigure}
    \begin{subfigure}[b]{0.495\textwidth}
        \includegraphics[width=\textwidth, trim = 0.3cm 0.3cm 0.3cm 0.3cm, clip]{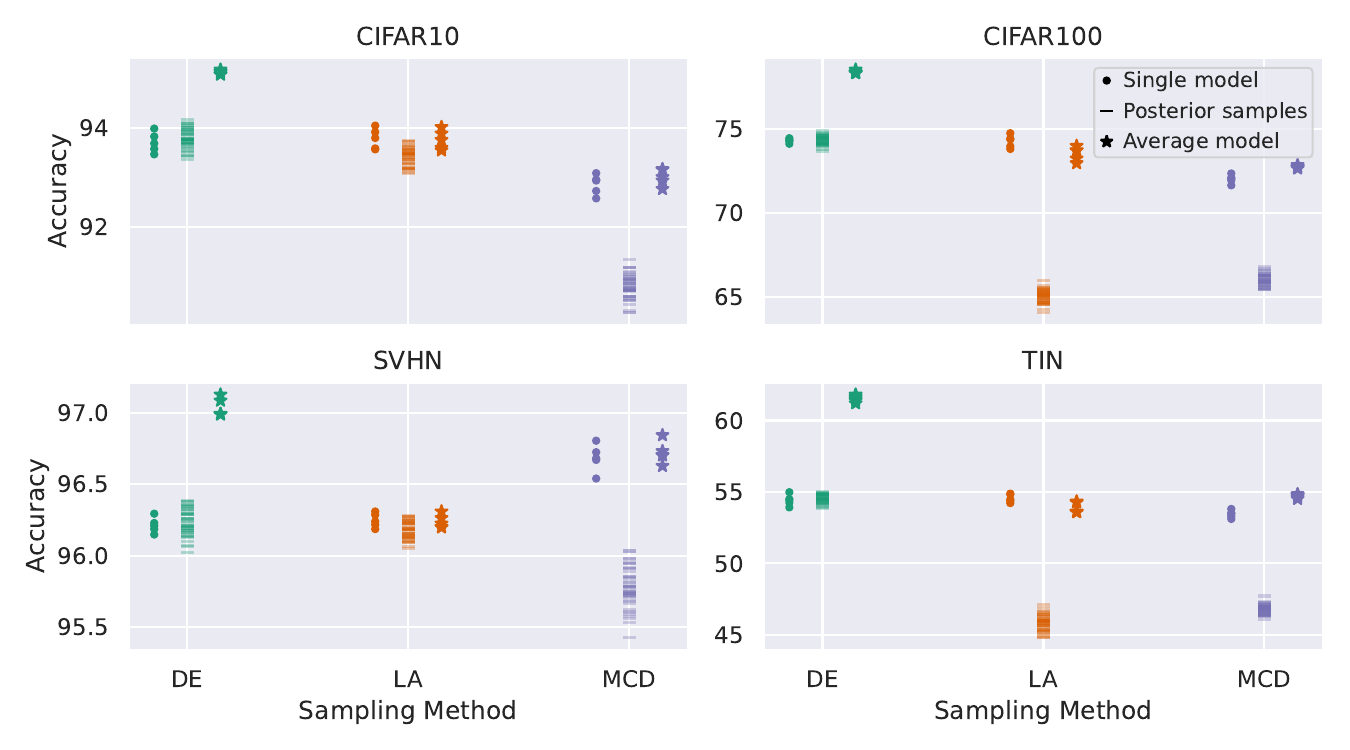}
        \subcaption{Accuracies}
    \end{subfigure}
    \begin{subfigure}[b]{0.495\textwidth}
        \includegraphics[width=\textwidth, trim = 0.3cm 0.3cm 0.3cm 0.3cm, clip]{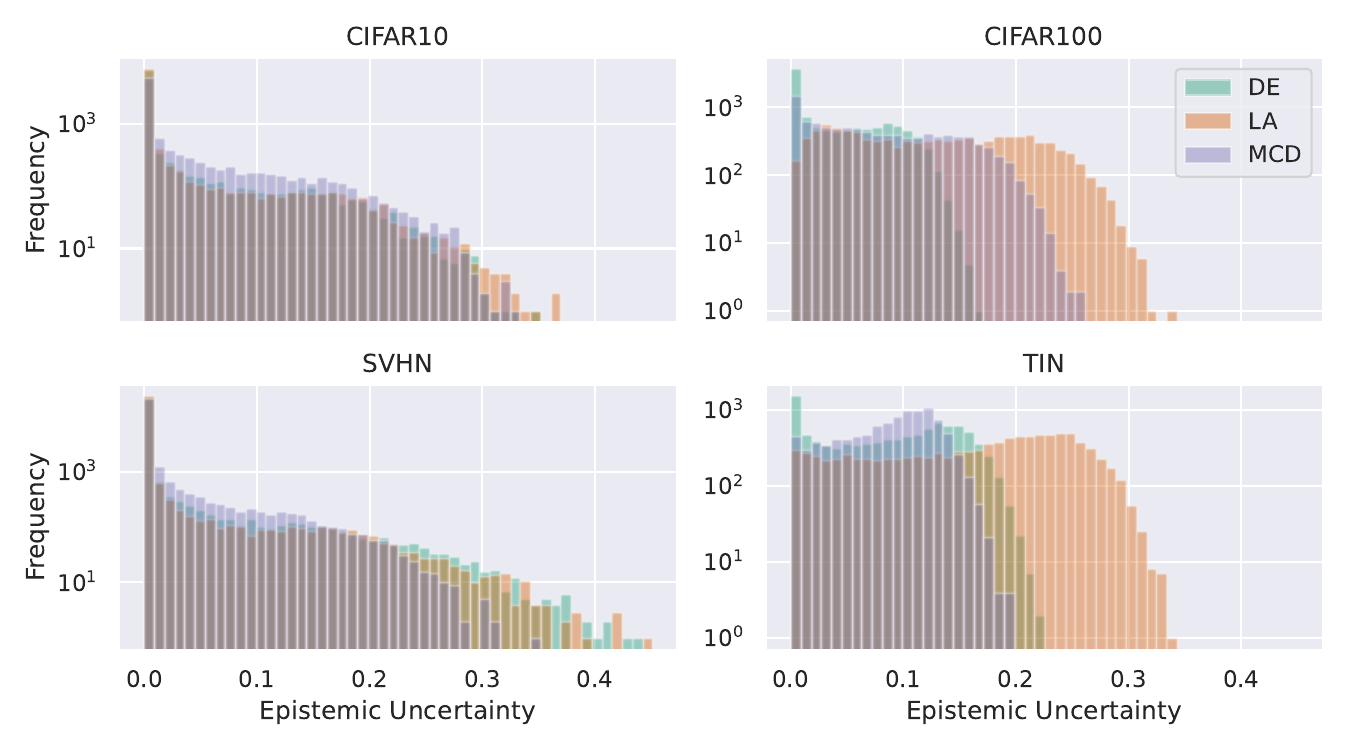}
        \subcaption{EU \texttt{(C2)}$/ \log(|\mathcal{Y}|)$}
    \end{subfigure}
    \caption{\textbf{Comparison of posterior sampling methods.} 
    Results are obtained on the test split of the respective dataset. 
    We compare the NLLs and accuracies for different models obtained through DE, LA and MCD. 
    Similarly, the normalized AU \texttt{(C)} and the normalized EU \texttt{(C2)} are shown per sampling method. 
    All three methods yield similar results for CIFAR10 and SVHN, but differ greatly on CIFAR100 and TIN. 
    Models sampled using LA have higher NLL and lower accuracy. 
    Furthermore, they lead to higher EU and lack predictions with very low AU. 
    Additionally, the average model does not improve over the single model in terms of NLL and accuracy for those two datasets. 
    The results in the two left plots show single models, posterior samples and average models of five runs, those in the two right plots show uncertainties for a single run.}
    \label{fig:posterior_sampling_comp}
\end{figure}

To better understand the performance of different posterior sampling methods, we examine the characteristics of their sampled models.
The results in Fig.~\ref{fig:posterior_sampling_comp} show that these methods perform differently across datasets.
For the global sampling method DE, the average model consistently outperforms individual sampled models with a lower negative log-likelihood (NLL) and higher accuracy across all datasets.
In contrast, for local sampling methods LA and MCD, individual sampled models exhibit higher NLL than both the single model and the average model.
Additionally, the accuracy of individual sampled models is lower than that of the single model.
Specifically, for MCD, the single model's accuracy is comparable to the average model, while for LA, the single model's accuracy exceeds that of the average model.

We further analyze the predictive uncertainties estimated by different posterior sampling methods using measure \texttt{(C2)}, which incorporates posterior samples and is upper-bounded.
To ensure comparability between datasets, we normalize the uncertainties by the maximal predictive uncertainty TU \texttt{(C2)}, equal to the entropy of the uniform distribution $\log(|\mathcal{Y}|)$.
The results in the two right plots in Fig.~\ref{fig:posterior_sampling_comp} show that these methods yield similar distributions of uncertainties for CIFAR10 and SVHN.
However, for CIFAR100 and TIN, DE exhibits many more datapoints with very low EU and AU.

\subsubsection{Aligning the Uncertainty Measure with the Predicting Model}

\paragraph{Selective Prediction.}

The results for LA are provided in Tab.~\ref{tab:selective_prediction_la}, for MCD in Tab.~\ref{tab:selective_prediction_mcd}.
In contrast to the global posterior sampling method DE, we do not find that it is not directly important to align the uncertainty measure to the predicting model for LA and MCD.
We find that no matter the predicting model, it is best to use measures for a single predicting model \texttt{(A)}.
This is likely due to the strong dependence of the sampling space on the original model, i.e. the MAP for LA and the model without dropout for MCD.
An indicator for this is, that AU \texttt{(A)} is better than AU \texttt{(B)} or \texttt{(C)} for all settings, which was never the case for DE (see Tab.~\ref{tab:selective_prediction}).

\setlength{\tabcolsep}{1.2pt}
\renewcommand{\arraystretch}{1.2}
\begin{table*} \smaller
\centering
\caption{\textbf{Selective prediction under different predicting models for LA.}  AUARC for different predicting models, i.e., the single model, the average model and a model according to the posterior, using different predictive uncertainty measures as score. We highlight the two best measures for each setting. For the single predicting model, and the average model, TU \texttt{(A3)} performs best and both TU \texttt{(A1)} and AU \texttt{(A)} second best. For models sampled according to the posterior, TU \texttt{(A3)} performs best, followed by TU \texttt{(A2)}. Results are averaged over all datasets with statistics over five runs.}
\label{tab:selective_prediction_la}
\begin{tabular}{lcccccccccccccccccc}
\hline
Measures: &  \multicolumn{6}{c}{TU}  & \multicolumn{3}{c}{AU} & \multicolumn{8}{c}{EU} & Random \\
\cdashline{2-7} \cdashline{11-18} \cline{1-1}
\multicolumn{1}{l}{{Prediction}}  & \texttt{A1} & \texttt{A2} & \texttt{A3} & \texttt{B/C1} & \textbf{\texttt{B/C2}} & \texttt{B/C3} & \texttt{A} & \textbf{\texttt{B}} & \texttt{C} & \texttt{A1} & \texttt{A2} & \texttt{A3} & \texttt{B1} & \texttt{B3} & \texttt{C1} & \texttt{C2} & \texttt{C3} & Baseline \\
\hline
\textit{Single} & \cellcolor{verylightgray}$87.53$ & $87.51$ & \cellcolor{verylightgray}$87.59$ & $86.87$ & $87.15$ & $87.18$ & \cellcolor{verylightgray}$87.53$ & $87.15$ & $87.03$ & $79.81$ & $83.65$ & $86.72$ & $80.67$ & $85.47$ & $84.22$ & $86.54$ & $85.94$ & $79.72$ \\
\textit{Model} & \cellcolor{verylightgray}$\pm0.14$ & $\pm0.14$ & \cellcolor{verylightgray}$\pm0.15$ & $\pm0.15$ & $\pm0.15$ & $\pm0.16$ & \cellcolor{verylightgray}$\pm0.14$ & $\pm0.15$ & $\pm0.14$ & $\pm0.16$ & $\pm0.20$ & $\pm0.18$ & $\pm0.22$ & $\pm0.20$ & $\pm0.21$ & $\pm0.17$ & $\pm0.18$ & $\pm0.13$ \\
\cdashline{1-19}[1pt/2pt]
\textit{Average}  & \cellcolor{verylightgray}$87.35$ & $87.34$ & \cellcolor{verylightgray}$87.44$ & $86.68$ & $86.96$ & $87.00$ & \cellcolor{verylightgray}$87.35$ & $86.96$ & $86.83$ & $79.44$ & $83.47$ & $86.60$ & $80.37$ & $85.28$ & $84.01$ & $86.37$ & $85.76$ & $79.36$ \\
\textit{Model} & \cellcolor{verylightgray}$\pm0.17$ & $\pm0.17$ & \cellcolor{verylightgray}$\pm0.18$ & $\pm0.18$ & $\pm0.18$ & $\pm0.19$ & \cellcolor{verylightgray}$\pm0.17$ & $\pm0.18$ & $\pm0.17$ & $\pm0.18$ & $\pm0.20$ & $\pm0.19$ & $\pm0.22$ & $\pm0.22$ & $\pm0.22$ & $\pm0.19$ & $\pm0.21$ & $\pm0.16$ \\
\cdashline{1-19}[1pt/2pt]
\textit{Acc. to} & $85.10$ & \cellcolor{verylightgray}$85.15$ & \cellcolor{verylightgray}$85.26$ & $84.67$ & $84.86$ & $84.99$ & $85.10$ & $84.86$ & $84.66$ & $76.63$ & $81.63$ & $84.69$ & $78.17$ & $83.39$ & $82.16$ & $84.53$ & $83.91$ & $76.56$ \\
\textit{Posterior} & $\pm0.12$ & \cellcolor{verylightgray}$\pm0.11$ & \cellcolor{verylightgray}$\pm0.13$ & $\pm0.12$ & $\pm0.11$ & $\pm0.13$ & $\pm0.12$ & $\pm0.11$ & $\pm0.11$ & $\pm0.14$ & $\pm0.12$ & $\pm0.14$ & $\pm0.17$ & $\pm0.16$ & $\pm0.17$ & $\pm0.14$ & $\pm0.15$ & $\pm0.08$ \\
\hline
\end{tabular}
\end{table*}
\renewcommand{\arraystretch}{1}

\setlength{\tabcolsep}{1.2pt}
\renewcommand{\arraystretch}{1.2}
\begin{table*} \smaller
\centering
\caption{\textbf{Selective prediction under different predicting models for MCD.}  AUARC for different predicting models, i.e., the single model, the average model and a model according to the posterior, using different predictive uncertainty measures as score. We highlight the two best measures for each setting. Regardless of the predicting model, TU \texttt{(A2)} performs best, followed by TU \texttt{(A3)}. Results are averaged over all datasets with statistics over five runs.}
\label{tab:selective_prediction_mcd}
\begin{tabular}{lcccccccccccccccccc}
\hline
Measures: &  \multicolumn{6}{c}{TU}  & \multicolumn{3}{c}{AU} & \multicolumn{8}{c}{EU} & Random \\
\cdashline{2-7} \cdashline{11-18} \cline{1-1}
\multicolumn{1}{l}{{Prediction}}  & \texttt{A1} & \texttt{A2} & \texttt{A3} & \texttt{B/C1} & \textbf{\texttt{B/C2}} & \texttt{B/C3} & \texttt{A} & \textbf{\texttt{B}} & \texttt{C} & \texttt{A1} & \texttt{A2} & \texttt{A3} & \texttt{B1} & \texttt{B3} & \texttt{C1} & \texttt{C2} & \texttt{C3} & Baseline \\
\hline
\textit{Single} & $86.97$ & \cellcolor{verylightgray}$87.15$ & \cellcolor{verylightgray}$87.11$ & $85.59$ & $86.56$ & $86.42$ & $86.97$ & $86.56$ & $86.53$ & $78.81$ & $84.57$ & $85.86$ & $81.53$ & $84.30$ & $82.83$ & $85.40$ & $84.76$ & $78.75$ \\
\textit{Model} & $\pm0.07$ & \cellcolor{verylightgray}$\pm0.07$ & \cellcolor{verylightgray}$\pm0.07$ & $\pm0.05$ & $\pm0.06$ & $\pm0.06$ & $\pm0.07$ & $\pm0.06$ & $\pm0.06$ & $\pm0.08$ & $\pm0.08$ & $\pm0.09$ & $\pm0.05$ & $\pm0.09$ & $\pm0.06$ & $\pm0.07$ & $\pm0.08$ & $\pm0.09$ \\
\cdashline{1-19}[1pt/2pt]
\textit{Average}  & $87.24$ & \cellcolor{verylightgray}$87.42$ & \cellcolor{verylightgray}$87.35$ & $86.04$ & $86.97$ & $86.81$ & $87.24$ & $86.97$ & $86.94$ & $79.38$ & $84.90$ & $86.11$ & $82.04$ & $84.68$ & $83.30$ & $85.77$ & $85.13$ & $79.30$ \\
\textit{Model} & $\pm0.06$ & \cellcolor{verylightgray}$\pm0.05$ & \cellcolor{verylightgray}$\pm0.05$ & $\pm0.06$ & $\pm0.04$ & $\pm0.04$ & $\pm0.06$ & $\pm0.04$ & $\pm0.05$ & $\pm0.07$ & $\pm0.06$ & $\pm0.05$ & $\pm0.08$ & $\pm0.04$ & $\pm0.07$ & $\pm0.05$ & $\pm0.04$ & $\pm0.07$ \\
\cdashline{1-19}[1pt/2pt]
\textit{Acc. to} & $84.93$ & \cellcolor{verylightgray}$85.22$ & \cellcolor{verylightgray}$85.21$ & $84.06$ & $84.88$ & $84.81$ & $84.93$ & $84.88$ & $84.78$ & $76.47$ & $83.03$ & $84.29$ & $80.01$ & $82.94$ & $81.42$ & $84.00$ & $83.39$ & $76.36$ \\
\textit{Posterior} & $\pm0.07$ & \cellcolor{verylightgray}$\pm0.07$ & \cellcolor{verylightgray}$\pm0.08$ & $\pm0.08$ & $\pm0.05$ & $\pm0.05$ & $\pm0.07$ & $\pm0.05$ & $\pm0.05$ & $\pm0.14$ & $\pm0.11$ & $\pm0.09$ & $\pm0.09$ & $\pm0.06$ & $\pm0.09$ & $\pm0.07$ & $\pm0.07$ & $\pm0.12$ \\
\hline
\end{tabular}
\end{table*}
\renewcommand{\arraystretch}{1}

\paragraph{Misclassification Detection.}

\begin{figure}
  \centering
    \captionsetup{aboveskip=2pt, belowskip=1pt}
  \begin{minipage}[t]{0.49\textwidth}
    \centering
    {\small Prediction: \textit{Single Model}}
    \includegraphics[width=\linewidth, trim = 0.3cm 1.5cm 0.3cm 1cm, clip]{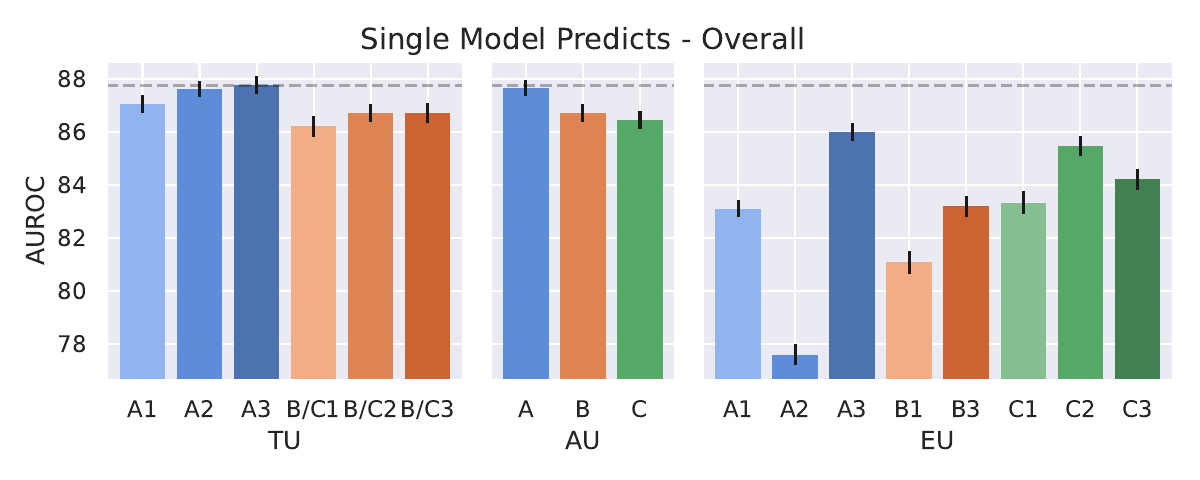}
    {\small Prediction: \textit{Average Model}}
    \includegraphics[width=\linewidth, trim = 0.3cm 1.5cm 0.3cm 1cm, clip]{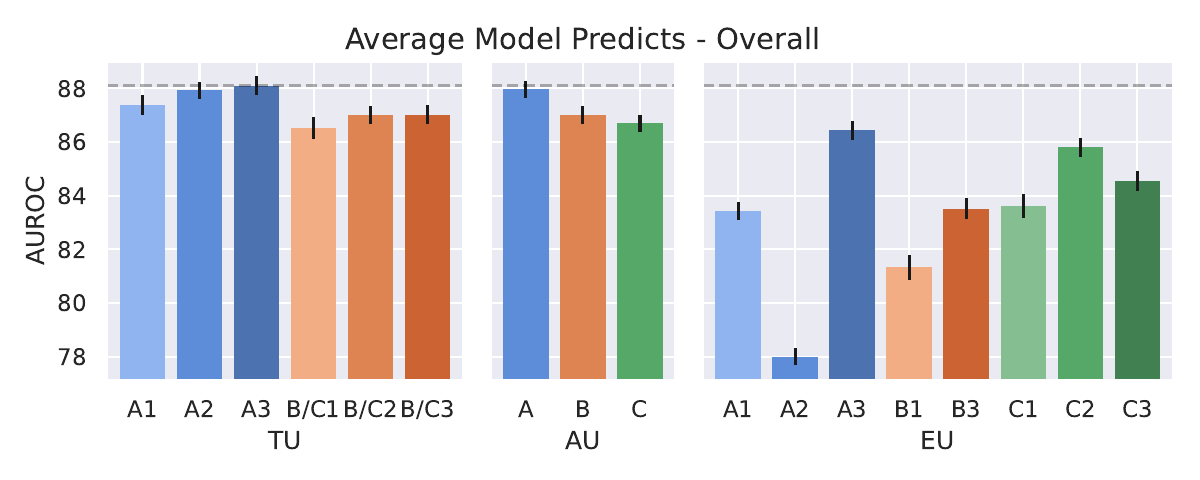}
    {\small Prediction: \textit{According to Posterior}}
    \includegraphics[width=\linewidth, trim = 0.3cm 0.3cm 0.3cm 1cm, clip]{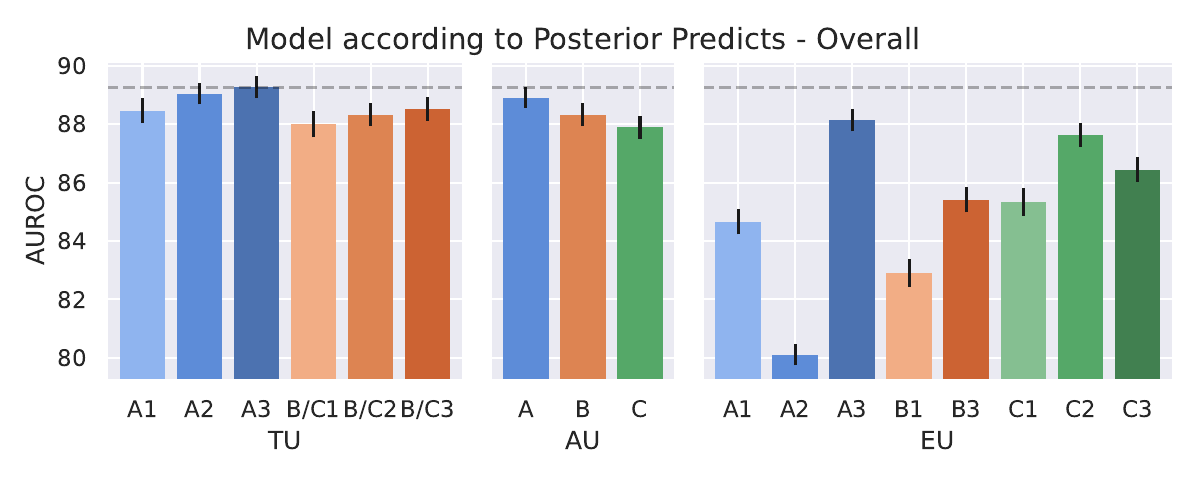} 
    \caption{\textbf{Misclassification detection under different predicting models for LA.} AUROC for distinguishing correct from incorrect predictions under different predicting models, using the different proposed measures of uncertainty as score. For all settings, TU \texttt{(A3)} performs best. AUROCs are averages over all datasets. Statistics over five runs.}
    \label{fig:misc_la}
  \end{minipage}
  \hfill
  \begin{minipage}[t]{0.49\textwidth}
    \centering
    {\small Prediction: \textit{Single Model}}
    \includegraphics[width=\linewidth, trim = 0.3cm 1.5cm 0.3cm 1cm, clip]{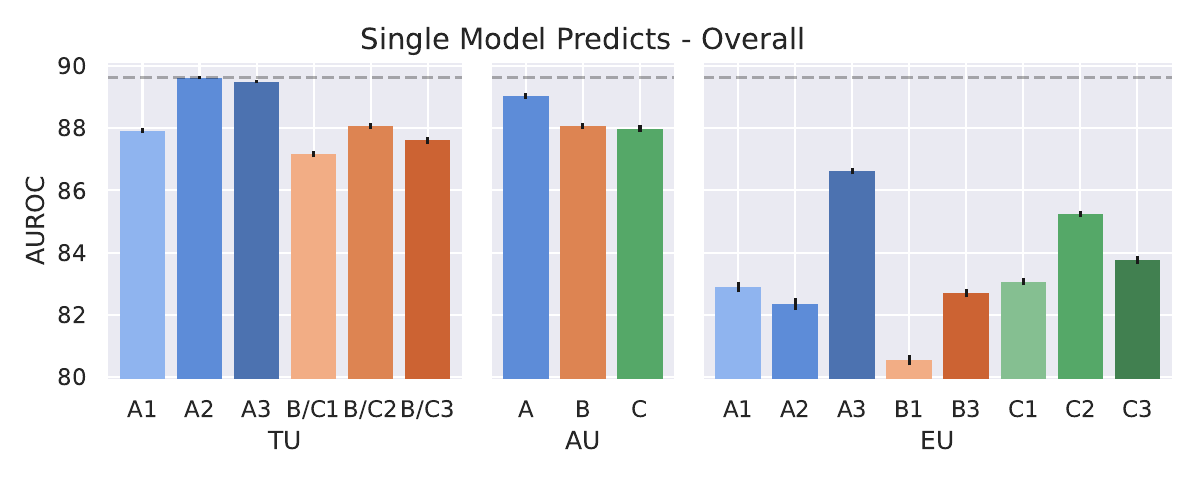}
    {\small Prediction: \textit{Average Model}}
    \includegraphics[width=\linewidth, trim = 0.3cm 1.5cm 0.3cm 1cm, clip]{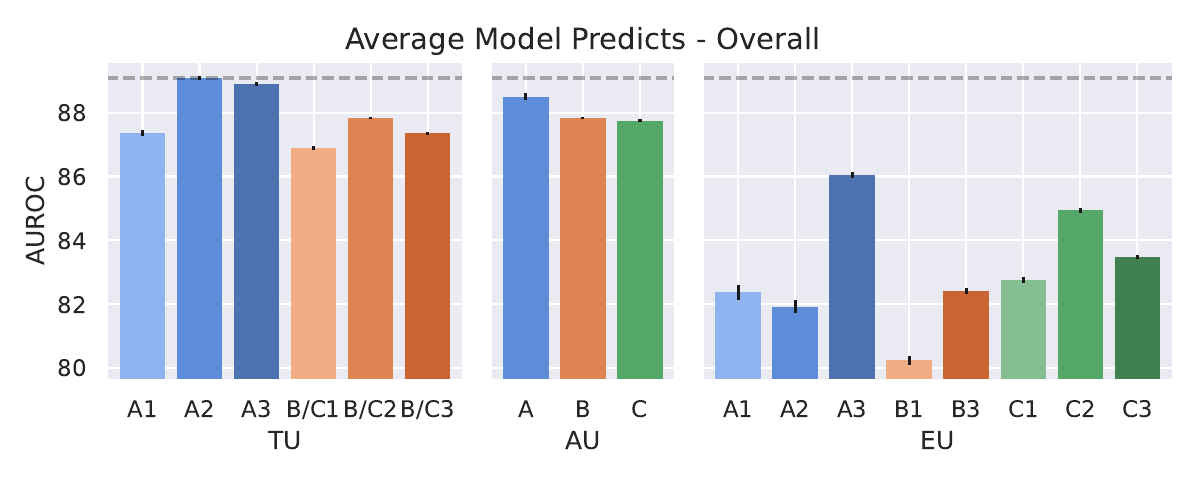}
    {\small Prediction: \textit{According to Posterior}}
    \includegraphics[width=\linewidth, trim = 0.3cm 0.3cm 0.3cm 1cm, clip]{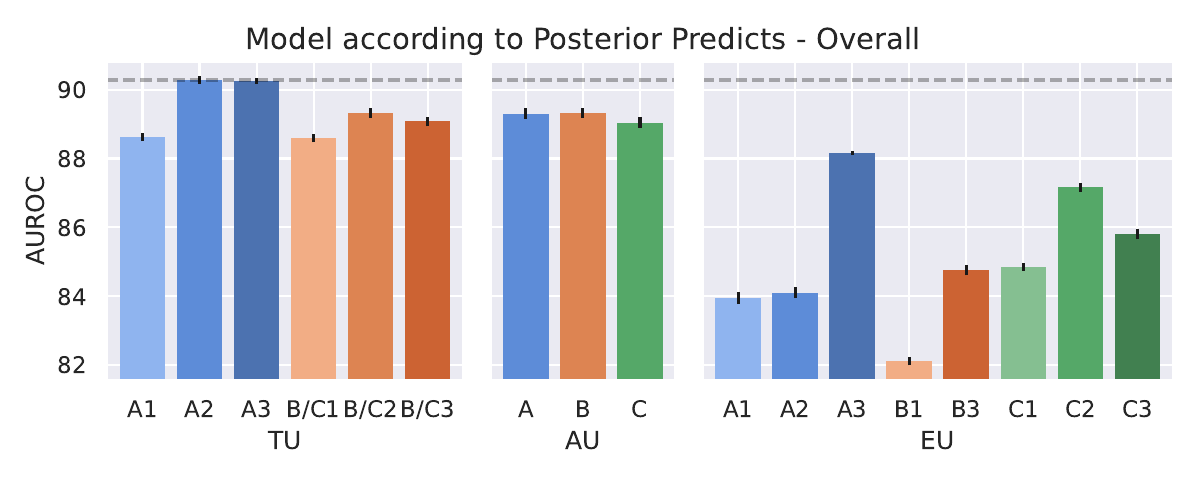} 
    \caption{\textbf{Misclassification detection under different predicting models for MCD.} AUROC for distinguishing correct from incorrect predictions under different predicting models, using the different proposed measures of uncertainty as score. For all settings, TU \texttt{(A2)} performs best. AUROCs are averages over all datasets. Statistics over five runs.}
    \label{fig:misc_mcd}
    
  \end{minipage}
\end{figure}

For misclassification detection, we have similar findings as for selective prediction.
The results for LA are shown in Fig.~\ref{fig:misc_la}, those for MCD in Fig.~\ref{fig:misc_mcd}.
Again, the measures for \texttt{(A)} perform best, regardless of the predicting model.
The reason for this behavior is likely again the same as for misclassification, that the original model, i.e. the MAP for LA and the model without dropout for MCD, has a strong dependence on the sampling space for those posterior sampling methods.
In accordance with this perspective, we find that AU \texttt{(A)} is better than AU \texttt{(B)} or \texttt{(C)} for all settings, which was never the case for DE (see Tab.~\ref{tab:selective_prediction}).

\subsubsection{Detecting Distributional Mismatch}

\begin{figure}
  \centering
    \captionsetup{aboveskip=2pt, belowskip=1pt}
  \begin{minipage}[t]{0.49\textwidth}
    \centering
    \includegraphics[width=\linewidth, trim = 0.3cm 0.3cm 0.3cm 1cm, clip]{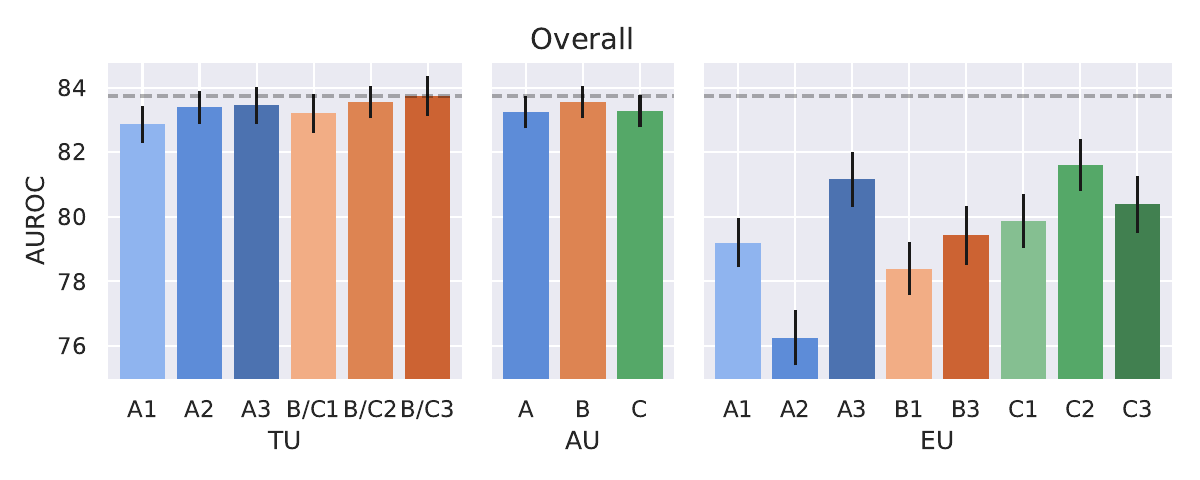}
    \caption{\textbf{OOD detection for LA.} AUROC for distinguishing between ID and OOD datapoints using the different proposed measures of uncertainty as score. TU \texttt{(B/C2)} and TU \texttt{(B/C3)} perform best. AUROCs are averaged over all ID / OOD combinations. Statistics over five runs.}
    \label{fig:ood_la}
  \end{minipage}
  \hfill
  \begin{minipage}[t]{0.49\textwidth}
    \centering
    \includegraphics[width=\linewidth, trim = 0.3cm 0.3cm 0.3cm 1cm, clip]{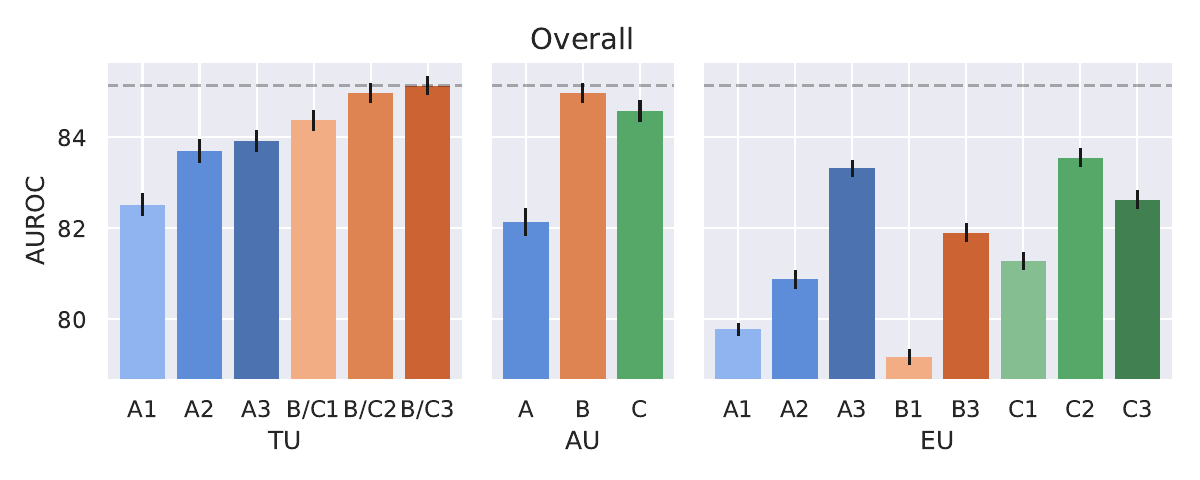}
    \caption{\textbf{OOD detection for MCD.} AUROC for distinguishing between ID and OOD datapoints using the different proposed measures of uncertainty as score. TU \texttt{(B/C2)} and TU \texttt{(B/C3)} perform best. AUROCs are averaged over all ID / OOD combinations. Statistics over five runs.}
    \label{fig:ood_mcd}
  \end{minipage}
\end{figure}

\paragraph{OOD Detection.}
The results for LA are provided in Fig.~\ref{fig:ood_la}, the results for MCD in Fig.~\ref{fig:ood_mcd}.
For MCD, the results are largely comparable to the results for DE in the main paper (see Fig.~\ref{fig:ood}), where TU \texttt{(B/C2)} which is equivalent to AU \texttt{(B)} and TU \texttt{(B/C3)} perform best.
The EU measures perform very poor in general, worse than most TU and AU measures.
For LA, we observe that all measures for TU and AU perform nearly identical and much better than EU meaures.
In sum, we surprisingly find that EU measures never outperform the best TU and AU measures for all posterior sampling methods.

\paragraph{Distribution Shift Detection.}

\begin{figure*}[t]
    \centering
    \captionsetup{aboveskip=2pt, belowskip=2pt}
    \begin{subfigure}[b]{\textwidth}
        \includegraphics[width=\textwidth, trim = 0.3cm 0.3cm 0.3cm 1cm, clip]{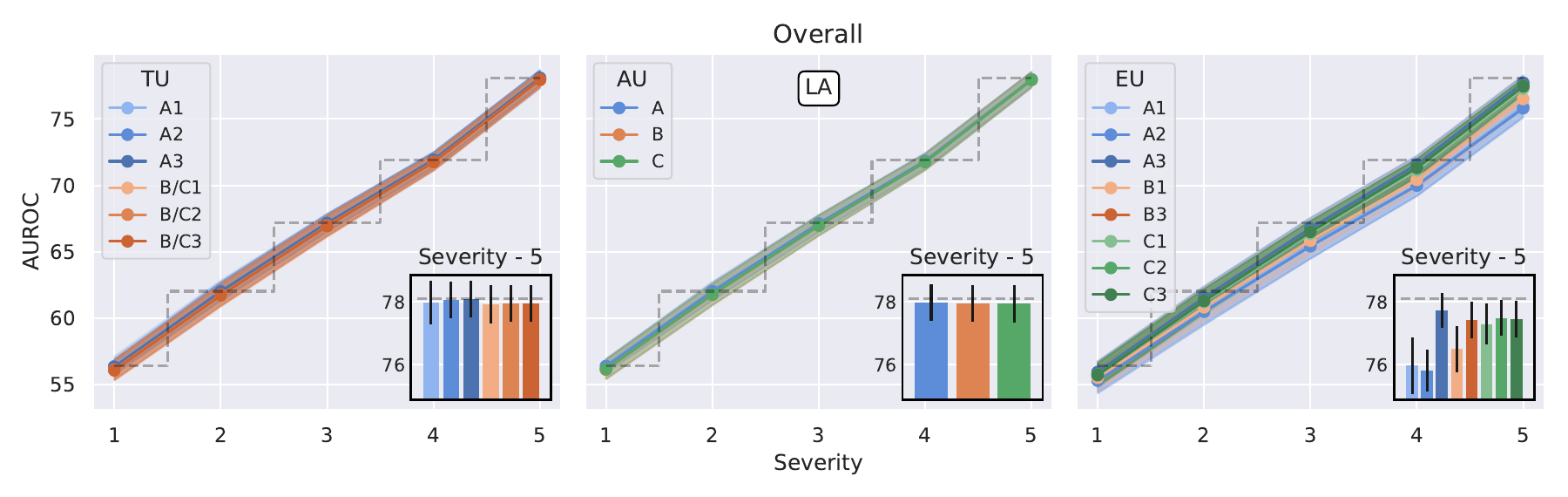}
    \end{subfigure}
    \caption{\textbf{Distribution shift detection on CIFAR10-C for LA.} AUROC for distinguishing between clean and corrupted test datapoints, using the different proposed measures of uncertainty as score. Black dashed line shows the maximum AUROC over all measures per severity. Insets shows detailed results for the highest severity. Statistics over five runs.}
    \label{fig:cifar10c_la}
\end{figure*}

\begin{figure*}[t]
    \centering
    \captionsetup{aboveskip=2pt, belowskip=2pt}
    \begin{subfigure}[b]{\textwidth}
        \includegraphics[width=\textwidth, trim = 0.3cm 0.3cm 0.3cm 1cm, clip]{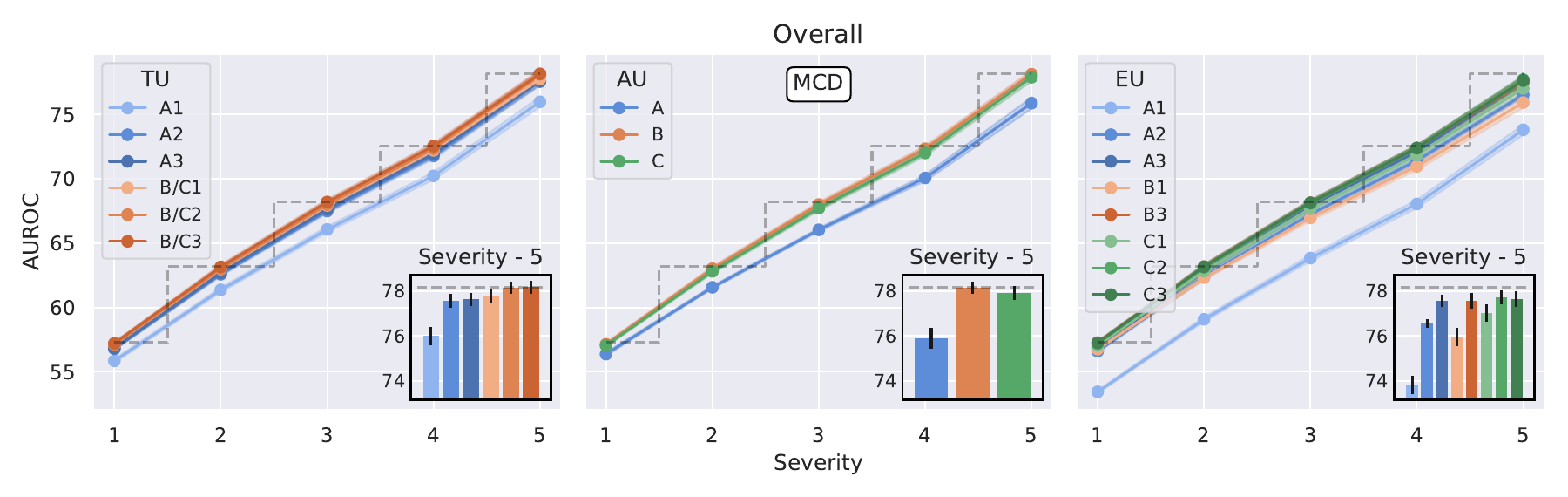}
    \end{subfigure}
    \caption{\textbf{Distribution shift detection on CIFAR10-C for MCD.} AUROC for distinguishing between clean and corrupted test datapoints, using the different proposed measures of uncertainty as score. Black dashed line shows the maximum AUROC over all measures per severity. Insets shows detailed results for the highest severity. Statistics over five runs.}
    \label{fig:cifar10c_mcd}
\end{figure*}

The results for distribution shift detection for LA are given in Fig.~\ref{fig:cifar10c_la}, the results for MCD in Fig.~\ref{fig:cifar10c_mcd}.
For LA, all TU and AU measures perform nearly identically, outperforming all EU measures in all severities.
For MCD, we find that except for the highest severity, the best TU, AU, and EU measures perform roughly on par.

\subsubsection{Disentanglement of Measures}

\begin{figure}
  \centering
    \captionsetup{aboveskip=2pt, belowskip=1pt}
  \begin{minipage}[t]{0.49\textwidth}
    \centering
    \includegraphics[width=\linewidth, trim = 0.4cm 0.5cm 1.5cm 1cm, clip]{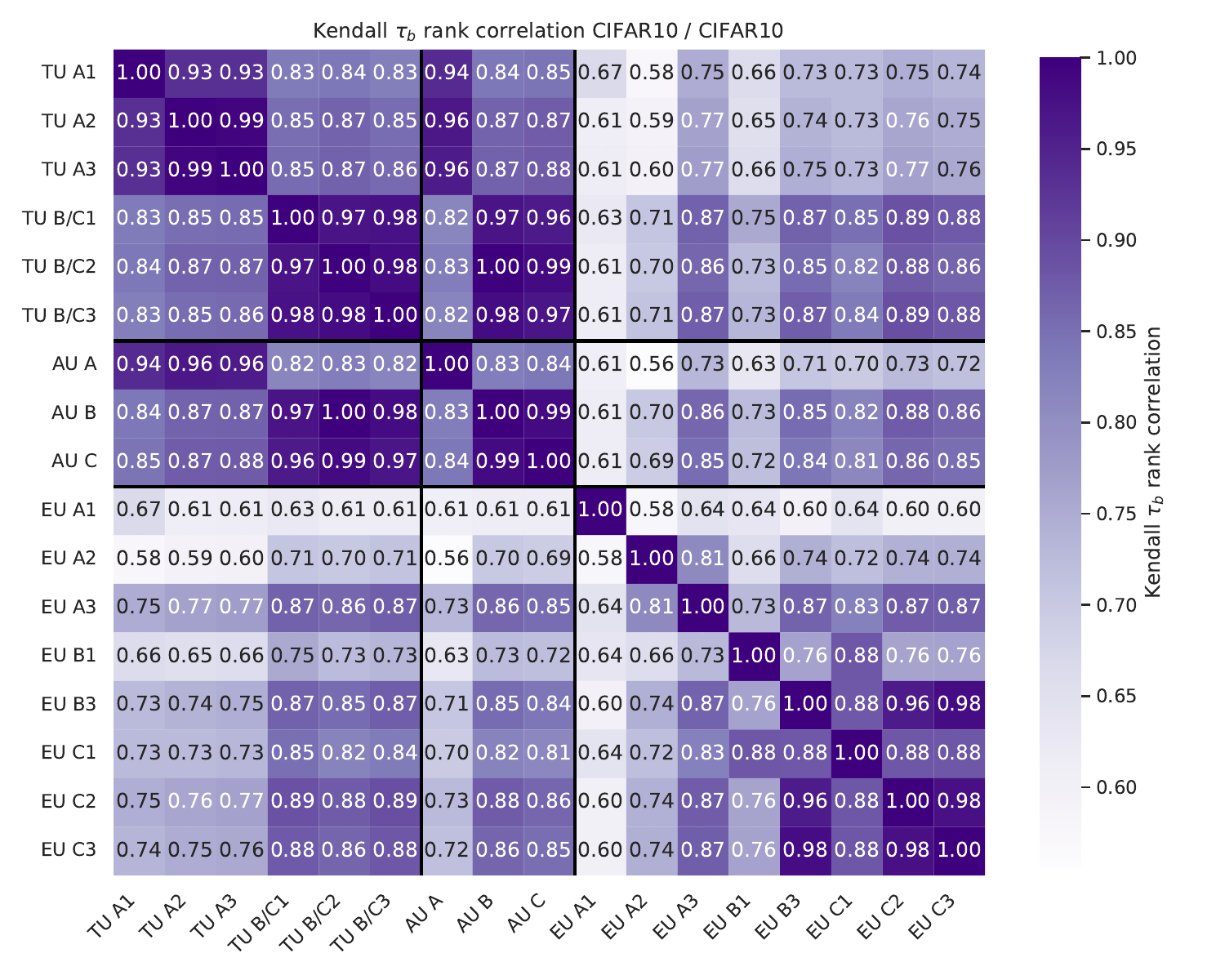}
    \caption{\textbf{Correlation of uncertainty measures on ID dataset (CIFAR10) for LA.} High rank correlation blocks exist.}
    \label{fig:correlation_id_la}
  \end{minipage}
  \hfill
  \begin{minipage}[t]{0.49\textwidth}
    \centering
    \includegraphics[width=\linewidth, trim = 0.4cm 0.5cm 1.5cm 1cm, clip]{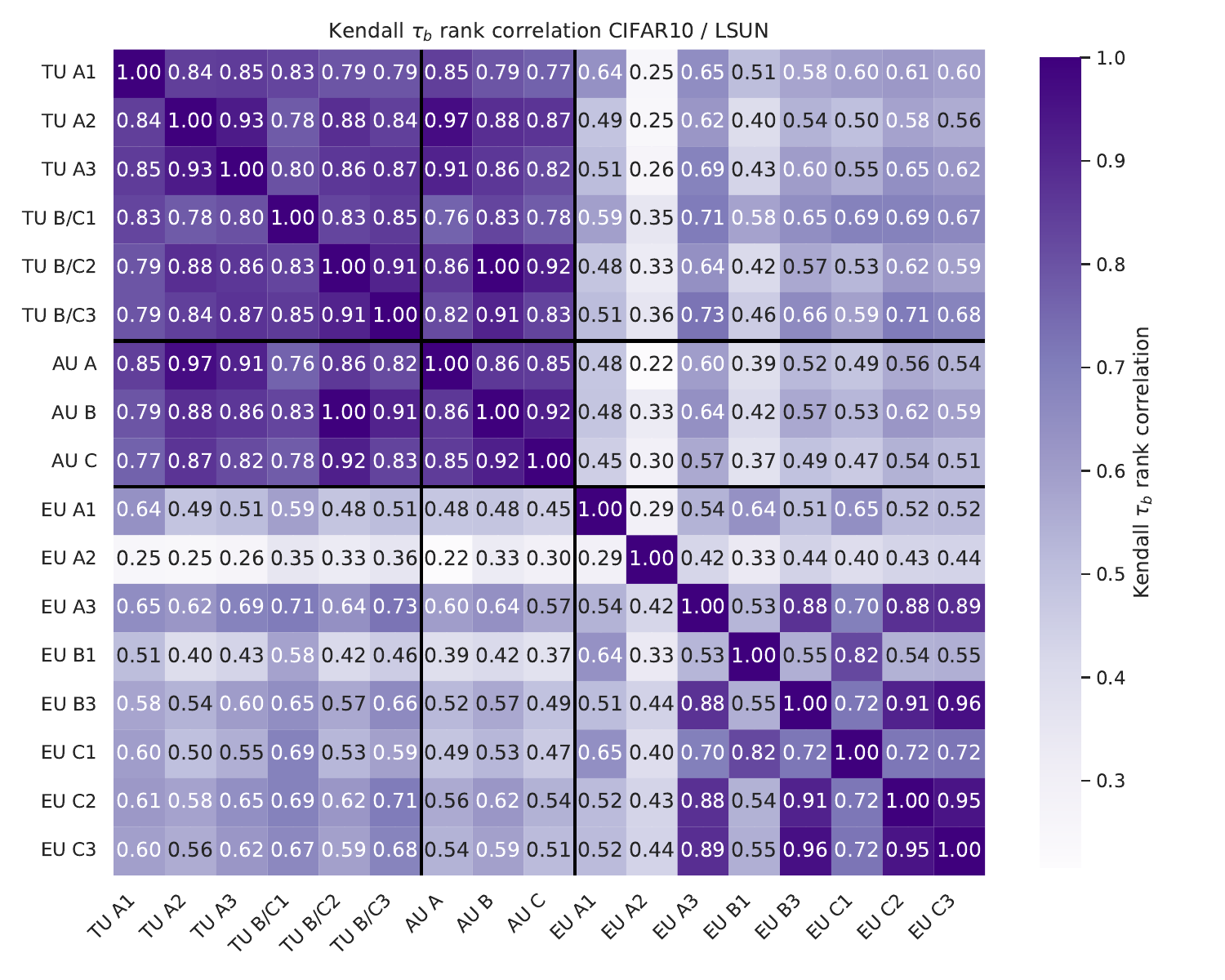}
    \caption{\textbf{Correlation of uncertainty measures on OOD dataset (LSUN) for LA.} All TU and AU measures are highly correlated.}
    \label{fig:correlation_ood_la}
  \end{minipage}
\end{figure}

\begin{figure}
  \centering
    \captionsetup{aboveskip=2pt, belowskip=1pt}
  \begin{minipage}[t]{0.49\textwidth}
    \centering
    \includegraphics[width=\linewidth, trim = 0.4cm 0.5cm 1.5cm 1cm, clip]{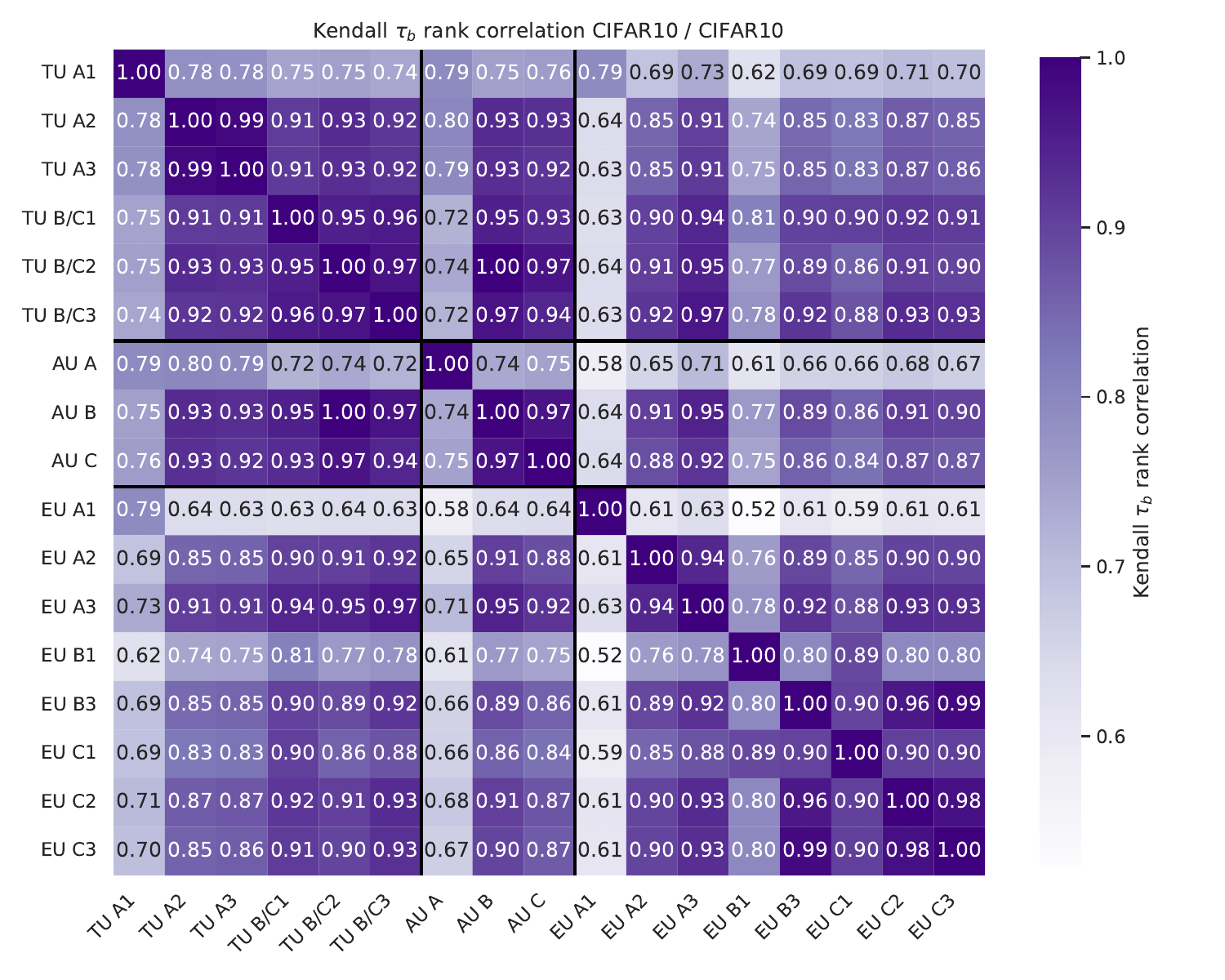}
    \caption{\textbf{Correlation of uncertainty measures on ID dataset (CIFAR10) for MCD.} High rank correlation blocks exist.}
    \label{fig:correlation_id_mcd}
  \end{minipage}
  \hfill
  \begin{minipage}[t]{0.49\textwidth}
    \centering
    \includegraphics[width=\linewidth, trim = 0.4cm 0.5cm 1.5cm 1cm, clip]{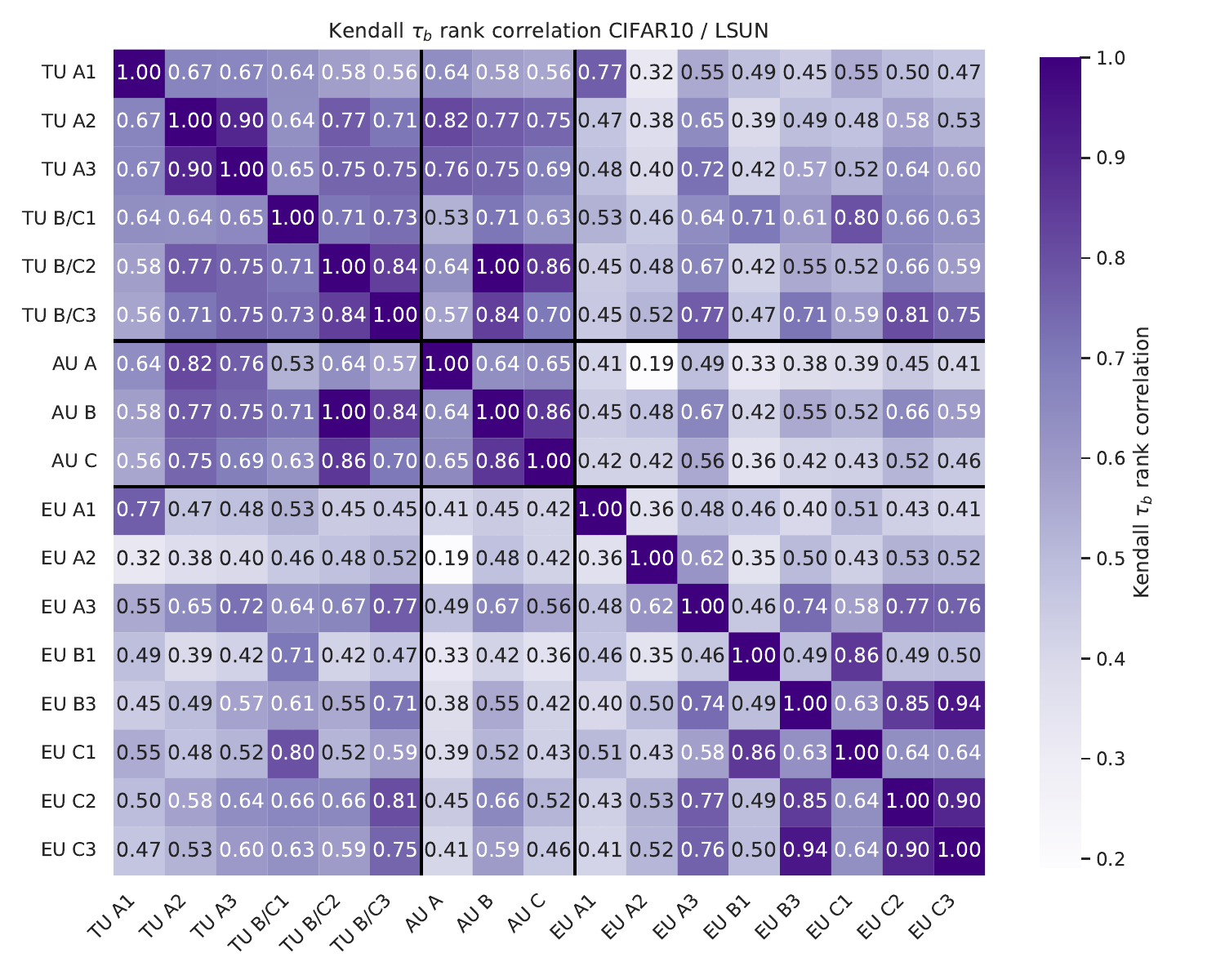}
    \caption{\textbf{Correlation of uncertainty measures on OOD dataset (LSUN) for MCD.} Rank correlations are very low overall.}
    \label{fig:correlation_ood_mcd}
  \end{minipage}
\end{figure}

The results for LA are provided in Fig.~\ref{fig:correlation_id_la} and Fig.~\ref{fig:correlation_ood_la}, the results for MCD in Fig.~\ref{fig:correlation_id_mcd} and Fig.~\ref{fig:correlation_ood_mcd}.
For MCD, the results are largely consistent with the findings for DE in the main paper for the OOD data.
For ID data, we find even more correlation between measures, not only within TU, AU, and EU measures, but also between those.
For LA, we find that the TU and AU measures are all highly correlated with the OOD data compared to other posterior sampling methods.
This explains the very similar performance of all TU and AU measures on OOD detection with LA as a posterior sampling method.
For the ID data, the results for LA are very similar to the results with DE in the main paper.

\subsubsection{Conclusions}

Overall, the behavior of the global posterior sampling method DE discussed in the main paper and the local posterior sampling methods LA and MCD discussed here differ in quite some dimensions.

First, we find that for LA the sampled models often perform poorly, especially for SVHN and TIN with more classes.
Here, the average model does not actually improve over the single models, as is the case for DE and MCD.
This might be a limitation of the particular LA used in this work.

Second, there does not appear to be a benefit in aligning the uncertainty measure to the predicting model for neither LA nor MCD.
We hypothesize that this is due to the fact that local methods probe local sensitivity around the original model, i.e. the MAP for LA and the model without dropout for MCD, leading to uncertainty measures based on this original model performing very well.
For the global method DE, there is no such original model that dominates the sampling space.

Third, for detecting distributional mismatch, MCD largely leads to the same results as DE.
However, for LA we find that all TU and AU measures perform on very similar levels.
This is also evident when we look at the rank correlations of measures.
Here we find a strong correlation between all the TU and AU measures on the OOD data for LA.
This is probably because the models sampled with LA do not strongly disagree with the original model on the OOD data.
Other than that, we find that the results match those presented for DE in the main paper.

\end{document}